\documentclass[twoside,11pt]{article}

\usepackage{jmlr2e}

\usepackage{natbib}
\usepackage{epigraph}
\usepackage{hyperref}
\usepackage{booktabs}
\usepackage[textsize=tiny]{todonotes}
\usepackage{namedtensor}
\usepackage{multirow}
\usepackage{floatflt}
\usepackage{makecell}
\usepackage{fancyhdr}
\usepackage{pifont}
\usepackage{comment}
\usepackage{tcolorbox}

\newcommand{\xmark}{\ding{55}}

\usepackage{listings}
\usepackage{amsfonts}
\ndef{\ax}{ax}
\ndef{\dd}{d}
\ndef{\layer}{layer}
\ndef{\seq}{seq}
\ndef{\subseq}{subseq}
\ndef{\key}{key}
\ndef{\val}{val}
\ndef{\heads}{heads}
\ndef{\batch}{batch}
\ndef{\inp}{input} \ndef{\hidden}{hidden} \ndef{\out}{out}
\ndef{\height}{height} \ndef{\width}{width} \ndef{\chans}{chans}
\ndef{\kernel}{kernel} \ndef{\kh}{kh} \ndef{\kw}{kw}
\ndef{\vocab}{vocab}
\ndef{\classes}{classes}
\ndef{\state}{state}
\ndef{\emb}{emb}

\usepackage{ulem}
\usepackage{hhline}
\usepackage{colortbl}

\usepackage{caption}
\usepackage{tabularx}
\usepackage{longtable}
\usepackage{adjustbox}
\usepackage{wrapfig}

\ndef{\R}{\mathbf{R}}

\newcommand{\orange}[1]{\textcolor{orange}{#1}}

\definecolor{bluegrad}{RGB}{200, 220, 240}
\definecolor{redgrad}{RGB}{255, 210, 210}

\definecolor{tablegreen}{RGB}{34, 139, 34}
\definecolor{tablered}{RGB}{220, 20, 60}

\definecolor{a_color}{HTML}{4285F4}
\definecolor{r_color}{HTML}{ED8E55}
\definecolor{s_color}{HTML}{E6C800} 

\newcommand{\ad}{\textcolor{a_color}{A}}
\newcommand{\rp}{\textcolor{r_color}{R}}
\newcommand{\se}{\textcolor{s_color}{S}}

\newcommand{\tred}[1]{\textcolor{tablered}{#1}}
\newcommand{\gren}[1]{\textcolor{tablegreen}{#1}}

\newcommand{\yes}[0]{\gren{\checkmark}}
\newcommand{\no}[0]{\tred{\xmark}}

\newcommand{\peft}{PEFT}

\renewcommand{\cite}[1]{\citep{#1}}

\usepackage[utf8]{inputenc}
\usepackage[T1]{fontenc}

\ShortHeadings{Scaling Down to Scale Up: A Guide to Parameter-Efficient Fine-Tuning}{Lialin, Deshpande, Yao, and Rumshisky}
\firstpageno{1}

\begin{document}

\title{Scaling Down to Scale Up: \\A Guide to Parameter-Efficient Fine-Tuning}

\author{\name Vladislav\ Lialin \email vlialin@cs.uml.edu \thanks{Correspondence to: \texttt{vlad.lialin@gmail.com}} \\
    \addr University of Massachusetts Lowell
    \AND
    \name Vijeta\ Deshpande \email vijeta\_deshpande@student.uml.edu \\
    \addr University of Massachusetts Lowell
    \AND
    \name Xiaowei\ Yao \email xiaowei\_yao@student.uml.edu \\
    \addr University of Massachusetts Lowell
    \AND
    \name Anna\ Rumshisky \email arum@cs.uml.edu \\
    \addr University of Massachusetts Lowell \\
    \addr Amazon Alexa AI
}

\editor{}

\maketitle

\begin{abstract}
This paper presents a systematic overview of parameter-efficient fine-tuning methods, covering over 50 papers published between early 2019 and mid-2024. These methods aim to address the challenges of fine-tuning large language models by training only a small subset of parameters. We provide a taxonomy that covers a broad range of methods and present a detailed method comparison with a specific focus on real-life efficiency in fine-tuning multibillion-scale language models.
We also conduct an extensive head-to-head experimental comparison of 15 diverse PEFT methods, evaluating their performance and eﬀiciency on models up to 11B parameters. Our findings reveal that methods previously shown to surpass a strong LoRA baseline face diﬀiculties in resource-constrained settings, where hyperparameter optimization is limited and the network is fine-tuned only for a few epochs. Finally, we provide a set of practical recommendations for using PEFT methods and outline potential future research directions.
\end{abstract}


\section{Introduction}
\begin{floatingfigure}[r]{0.4\textwidth}
\epigraph{One thing that should be learned from the bitter lesson is the great power of general purpose methods, of methods that continue to scale with increased computation...}{Rich Sutton, The Bitter Lesson}
\vspace{-1em}
\end{floatingfigure}

In October 2018, BERT Large \cite{devlin2018bert} with 350 million parameters was the biggest Transformer model \cite{vaswani2017attention} ever trained. At the time, contemporary hardware struggled to fine-tune this model. The section ``Out-of-memory issues'' on BERT's GitHub\footnotemark{} specifies the maximum batch size for BERT Large given 12Gb of GPU RAM and 512 tokens as \textbf{zero}.
Five years in, publicly available models grew to 176 billion parameters \cite{bloom,zhang2022opt,zeng2022glm130b}, i.e. by a factor of 500. Published literature includes models up to 1 trillion parameters \cite{palm,megatron,switch}. However, single-GPU RAM increased less than 10 times due to the high cost of HBM memory.
Model size scales almost \textbf{two orders of magnitude quicker} than computational resources making fine-tuning the largest models to downstream tasks infeasible for most and impractical for everyone.

\footnotetext{\href{https://github.com/google-research/bert}{github.com/google-research/bert}}

In-context learning \cite{radford2019language} thus became the new normal, the standard way to pass downstream task training data to billion-scale language models. However, the limited context length imposed by the transformer architecture \cite{vaswani2017attention, huang2018music}, the absence of ICL abilities in moderately large language models \cite{lu2023emergent}, the quadratic increase in computational cost with an increase in context length (or demonstrations in ICL) \cite{keles2023computational}, and the sensitivity of ICL performance \cite{bertsch2024context} present challenges in the utility, reliability, and efficiency of ICL. 
In cases where the model performs at par or better in the ICL setting compared to the fine-tuned model, fine-tuning is still a lucrative strategy due to the impractical inference cost of ICL \cite{bertsch2024context}.
Thus, we, as a community of researchers and engineers, need efficient ways to train on downstream task data.

Parameter-efficient fine-tuning (PEFT) aims to resolve this problem by only training a small set of parameters, which might be a subset of the existing model parameters or a set of newly added parameters. These methods differ in parameter and memory efficiency, training speed, final model quality, and additional inference costs (if any).

In the last few years, more than a hundred PEFT papers have been published, with several studies \cite{delta_tuning} providing a good overview of the most popular methods, such as Adapters \cite{adapters}, BitFit \cite{bitfit}, LoRA \cite{lora}, Compacter \cite{compacter}, and Soft Prompts \cite{p_tuning,prefix_tuning}.

\citet{modular_deep_learning} presented a survey on modular deep learning, providing an overview of several similar methods from the perspective of modularity and multi-task inference. Our focus differs by concentrating on PEFT methods, specifically for fine-tuning large language models, where minimizing RAM consumption and training time without sacrificing performance is crucial.

This survey presents a systematic overview, comparison, and taxonomy of 30 parameter-efficient fine-tuning methods. Over the last year, research efforts have also focused on replicating the success of PEFT in the pre-training regime. Hence, we also discuss a few prominent methods that aim to achieve efficiency gains during pre-training. We discuss 30 methods in-depth, covering over 50 papers published from 
early 2019 to mid-2024. We highlight the current unresolved challenges in PEFT, including the limited theoretical understanding, the performance gap between PEFT and traditional fine-tuning, and reporting issues.

We conduct the most 
extensive experimental comparison of PEFT methods 
, evaluating 14 methods and their variations across five datasets and three model sizes (0.7B, 3B, and 11B). The study includes a detailed comparison of these methods' efficiency in terms of GPU memory consumption and throughput. Our findings reveal that methods previously shown to outperform LoRA struggle to do so in resource-constrained settings and exhibit high hyperparameter sensitivity in hybrid PEFT methods. 

We found that Kronecker-based reparametrizations, while not enhancing memory efficiency compared to matrix-product counterparts, can improve training and inference speeds with efficient implementation. Surprisingly, Layer Norm tuning performs exceptionally well compared to most PEFT methods in our study. We also note a significant discrepancy between reported and actual trainable parameter counts in PEFT methods. This leads to unforeseen outcomes, such as the high computational costs of Prompt Tuning and Hybrid methods.
Our code is available on Github\footnote{\href{http://github.com/guitaricet/peft_comparison}{github.com/guitaricet/peft\_comparison}}.

In conclusion, we suggest several avenues for improvement, such as developing standardized PEFT benchmarks,
conducting in-depth studies on hyperparameters and interpretability,
exploring the difference in training dynamics of reparametrized neural networks,
further improving training and inference efficiency of PEFT methods,
and utility of PEFT methods with quantized backbone models.


\section{Background: Transformer}
Many of the parameter-efficient fine-tuning techniques discussed in this survey can be applied to all neural networks, though some are specifically designed for the Transformer architecture \cite{vaswani2017attention}. Given that Transformers are the largest neural networks ever trained, these methods are particularly valuable. Thus, we present a brief overview of the Transformer to provide context for these techniques.

\begin{figure}[h]
    \begin{minipage}{0.5\textwidth} 
        \begin{equation}
            \label{eq:def_att}
            \begin{aligned}
                & \operatorname{Att}
                \colon
                \mathbb{R}^{\key}
                \times
                \mathbb{R}^{\seq \times\key}
                \times \mathbb{R}^{\seq \times \val}
                \rightarrow \mathbb{R}^{\val} \\
                & \operatorname{Att}(Q,K,V) = \left( \nfun{\seq}{softmax} \frac{Q \ndot{\key} K}{\sqrt{|\key|}} \right) \ndot{\seq} V \\
            \end{aligned}
        \end{equation}
        \begin{equation*}
            \label{eq:kqv}
            \begin{aligned}
                Q(x) &= x \cdot W_Q + b_k, \\
                K(x) &= x \cdot W_K + b_q, \\
                V(x) &= x \cdot W_V + b_v, \\
            \end{aligned}
        \end{equation*}

        \begin{equation*}
            \begin{aligned}
                W_Q, \text{ } W_K &\in \R^{\inp \times \key}, W_V \in \R^{\inp \times \val} \\
                b_Q, \text{ } b_K &\in \R^{\key}, \text{ } b_V \in \R^{\val}
            \end{aligned}
        \end{equation*}
    \end{minipage}%
    \begin{minipage}{0.5\textwidth} 
        \centering
        \includegraphics[width=0.6\linewidth]{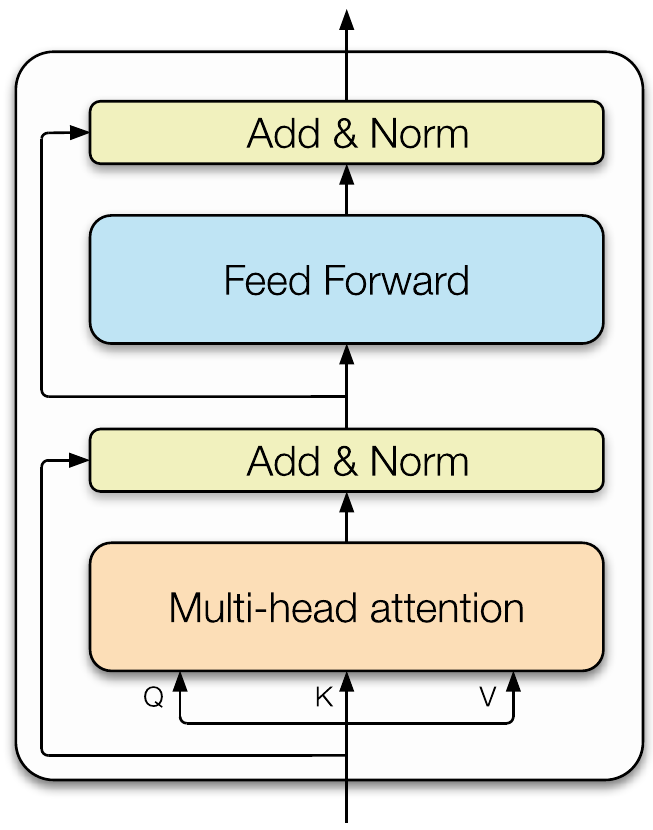}
        \caption{Basic Transformer block}
        \label{fig:transformer}
    \end{minipage}
\end{figure}

The core building block of the Transformer architecture consists of multi-head attention (MHA) followed by a fully-connected layer (FFN), as illustrated in Figure~\ref{fig:transformer}. Both attention and fully-connected layers incorporate residual connections \cite{resnet} and Layer Normalization \cite{layer_norm} to improve trainability.

The heart of the Transformer is the attention operation \cite{bahdanau2014neural}. It computes the softmax-normalized weighted average of the input tokens. Attention weights are computed according to the pairwise dot-product between each token key and query. Keys and queries are typically computed as simple linear projections of the input tokens.
Equation \ref{eq:def_att} describes it in the NamedTensor notation \cite{named_tensor}.

A number of methods act specifically on the matrices $W_K$, $W_Q$, $W_V$; they provide the main mechanism to pass information from one token to another and control what information (value) is being passed.

Although specific implementations of the Transformer may vary, such as incorporating a cross-attention layer in seq2seq networks or using LayerNorm before sublayers (Pre-LN), most parameter-efficient fine-tuning methods for Transformers only rely on the basic MHA + FFN structure. These methods can be readily adapted to architectural variations.

\section{Taxonomy of PEFT: a birds-eye view}
\label{sec:taxonomy}

\begin{figure*}
    \centering
    \includegraphics[width=1.0 \textwidth]{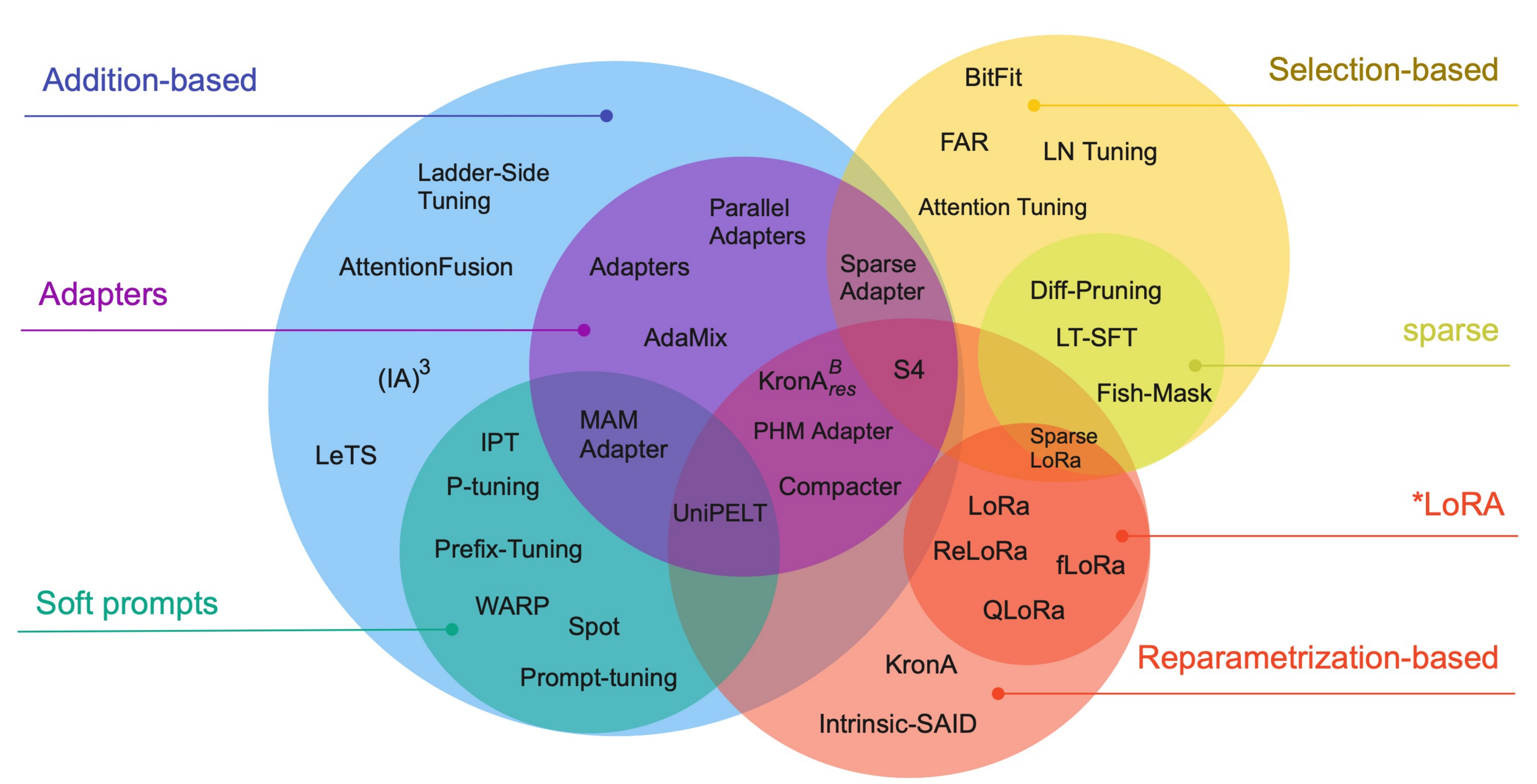}
    \caption{Parameter-efficient fine-tuning methods taxonomy. We identify three main classes of methods: \textbf{Addition}-based, \textbf{Selection}-based, and \textbf{Reparametrization}-based. Within additive methods, we distinguish two large included groups: \textbf{Adapter-like} methods and \textbf{Soft prompts}.}
    \label{fig:taxonomy}
\end{figure*}

PEFT methods can be categorized by their approach or objective. In terms of approach, they can introduce new parameters, fine-tune existing ones, or reparameterize them. In terms of objectives, they aim to minimize memory footprint, improve storage efficiency, or add modularity.
In this section, we begin by presenting a taxonomy based on the former. Figure \ref{fig:taxonomy} and Sections \ref{sec:additive}-\ref{sec:hybrid} present an overview of the taxonomy and illustrate over 30 PEFT methods. In the following sections, we describe these PEFT methods in detail, \textbf{accompanied by easy-to-understand pseudo-code}, in Sections \ref{sec:additive_section_adapters} - \ref{sec:hybrid_section}.

\subsection{\ad dditiion-based methods}
\label{sec:additive}

Addition-based methods augment the pre-trained model with additional parameters or layers and train only the newly introduced elements. This is the largest and most widely explored category of PEFT methods. Within this category, two large subcategories have emerged: Adapter-like methods and Soft Prompts.

%
\paragraph{Adapters}
Adapters \cite{adapters} are a type of additive PEFT method that introduces small fully-connected networks after Transformer sub-layers. The idea has been widely adopted \cite{adapterhub} \footnote{\href{https://github.com/adapter-hub/adapter-transformers}{github.com/adapter-hub/adapter-transformers}}, and multiple variations of Adapters have been proposed. These variations include modifying the placement of adapters \cite{parallel_adapter,parallel_adapter2}, pruning \cite{sparse_adapter}, and using reparametrization to reduce the number of trainable parameters \cite{compacter}.
Section \ref{sec:additive_section_adapters} discusses Adapter-based methods in detail.

\paragraph{Soft Prompts}

Language model prompting \cite{radford2019language} aims to control the behavior of a language model by modifying the input text, which typically consists of a task description accompanied by a few in-context examples. However, these methods are difficult to optimize and are inherently limited by the maximum model input length. To address these drawbacks, the concept of ``soft'' prompts was introduced \cite{p_tuning,prompt_tuning,prefix_tuning}, where a part of the model's input embeddings is fine-tuned via gradient descent. This pivots the problem of finding prompts in a discrete space to a continuous optimization problem. Soft prompts can be trained for the input layer only \cite{p_tuning,prompt_tuning} or for all layers \cite{prefix_tuning}. Recent advancements explore how soft prompts could be pre-trained or partially reused to reduce the fine-tuning costs \cite{spot,Hambardzumyan2021WARPWA,prompt_mapping,ipt}. We discuss Soft Prompts in Section \ref{sec:additive_section_soft_prompts}.

\paragraph{Other additive approaches}
Additive methods are a diverse category of parameter-efficient fine-tuning techniques that extend beyond adapters and soft prompts. We discuss these in Section \ref{sec:additive_section_others}.

\paragraph{Why add parameters?}
Although these methods introduce additional parameters to the network, they achieve significant speed and memory improvements. This is achieved by reducing the size of the gradients and optimizer states. In the case of Adam \cite{adam}, for every byte of trainable parameter, one extra byte is needed for its gradient, and two more bytes are needed to store the optimizer state: the first and second moments of the gradient. In practice, training a model requires 12-20 times more GPU memory than the model weights. By saving memory on optimizer states, gradients, and quantizing frozen model parameters \cite{dettmers2023qlora}, additive PEFT methods enable the fine-tuning of much larger networks or the use of larger microbatch sizes\footnote{Batch size = microbatch size $\times$ gradient accumulation $\times$ num. devices}, which improves training throughput on GPUs.
Moreover, the optimizer update step takes less time for PEFT methods than for full fine-tuning due to the smaller number of parameters to be updated. This effect is very noticeable in practice. Finally, optimizing fewer parameters in distributed setups drastically reduces communication volume.

\footnotetext[5]{Depends on sparse operations hardware support.}

\subsection{\se election-based methods}

Arguably the earliest example of selective PEFT is fine-tuning only a few top layers of a network \cite{last_layer_tuning}. Modern approaches are usually based on the type of layer \cite{cross_attention_tuning,layer_norm_tuning} or the internal structure, such as tuning only model biases \cite{bitfit} or particular rows \cite{far_edge}.

Sparse training is another version of selective PEFT that often ignores the structure and selects parameters to tune based on carefully designed selection criteria \cite{fish_mask,lottery_ticket_tuning,diff_pruning}.
However, sparse parameter updates present multiple engineering and efficiency challenges. Some of them have been tackled in recent research on parameter reconfiguration \cite{far_edge} (Section \ref{sec:far}) and NxM sparsity \cite{nxm_transformer}. Nevertheless, unrestricted unstructured sparsity is still impractical on contemporary hardware.

We discuss selective methods in detail in Section \ref{sec:selective_methods}.

\subsection{\rp eparametrization-based methods}
Reparametrization-based parameter-efficient fine-tuning methods leverage low-rank representations to minimize the number of trainable parameters. The notion that neural networks have low-dimensional representations has been widely explored in both empirical and theoretical analyses of deep learning \cite{parametercounting,measuring_the_intrinsic_dimension,Arora2018StrongerGB,malladi2022kernel}.

\citet{intrinsic_said} have demonstrated that fine-tuning can be performed effectively in low-rank subspaces. Furthermore, they showed that the size of the subspace that needs adaptation is smaller for larger models or models pre-trained for longer periods. Their approach, referred to as Intrinsic SAID (Section \ref{sec:intrinsic_said}), employs the Fastfood transform \cite{Le2013FastfoodAK} to reparametrize the update to neural network parameters. This inspired multiple methods, such as IntrinsicSAID, LoRA, and KronA (Sections \ref{sec:intrinsic_said}, \ref{sec:lora}, \ref{sec:krona}).

However, perhaps the most well-known reparametrization-based method is Low-Rank Adaptation or LoRA \cite{lora}, which employs a simple low-rank matrix decomposition to parametrize the weight update $\delta W = W^{\text{down}}W^{\text{up}}$. This approach is straightforward to implement and has been evaluated on models with up to 175 billion parameters. We provide a detailed discussion of this method in Section \ref{sec:lora}. More recent works \cite{compacter,krona} have also explored the use of Kronecker product reparametrization ($\delta W = A \otimes B$), which yields a more favorable tradeoff between rank and parameter count. Reparametrization-based methods have recently become widely popular, demonstrating their effectiveness on models up to 175B parameters \cite{lora}. Section \ref{sec:reparametrization_based_methods} discusses these methods in detail.

\subsection{Hybrid methods}
\label{sec:hybrid}
A number of methods combine ideas from multiple categories of PEFT \cite{parallel_adapter,sparse_adapter,unipelt,compacter}. This allows the use of different algorithmic tradeoffs to optimize for a specific goal, such as the number of trainable parameters. For instance, the MAM Adapter (Section \ref{sec:mam_adapter}) incorporates both Adapters and Prompt tuning. UniPELT (Section \ref{sec:unipelt}) adds LoRA to the mixture. Compacter and KronA$^B_{res}$ reparametrize the adapters using the Kronecker product to reduce their parameter count (Sections \ref{sec:compacter} and \ref{sec:krona}).

\begin{table*}[t]
\centering
\begin{small}
\begin{tabular}{l|c|ccc|c}
\toprule
\multirow{2}{*}{\textbf{Method}} & \multirow{2}{*}{\textbf{Type}} & \multicolumn{3}{c|}{\textbf{Efficiency}} & \multirow{2}{*}{\textbf{Inference overhead}} \\
& & \textbf{Disk} & \textbf{RAM} & \textbf{BP} & \\
\midrule
Adapters          \cite{adapters}                & \ad     & \yes & \yes & \no  & + FFN                    \\
AdaMix            \cite{adamix}                  & \ad     & \yes & \yes & \no  & + FFN                    \\
SparseAdapter     \cite{sparse_adapter}          & \ad \se & \yes & \yes & \no  & + FFN                    \\
Cross-Attn tuning \cite{cross_attention_tuning}  & \se     & \yes & \yes & \no  & \gren{No overhead}           \\
BitFit            \cite{bitfit}                  & \se     & \yes & \yes & \no  & \gren{No overhead}           \\
DiffPruning       \cite{diff_pruning}            & \se     & \yes & \no  & \no  & \gren{No overhead}           \\
Fish-Mask         \cite{fish_mask}               & \se     & \yes & \orange{\xmark}\footnotemark[5] & \no  & \gren{No overhead} \\
LT-SFT            \cite{lottery_ticket_tuning}   & \se     & \yes & \orange{\xmark}\footnotemark[5] & \no  & \gren{No overhead} \\
Prompt Tuning     \cite{prompt_tuning}           & \ad     & \yes & \yes & \no  & + input                  \\
Prefix-Tuning     \cite{prefix_tuning}           & \ad     & \yes & \yes & \no  & + input                  \\
Spot              \cite{spot}                    & \ad     & \yes & \yes & \no  & + input                \\
IPT               \cite{ipt}                     & \ad     & \yes & \yes & \no  & + FFN and input \\
MAM Adapter       \cite{parallel_adapter}        & \ad     & \yes & \yes & \no  & + FFN and input          \\
Parallel Adapter  \cite{parallel_adapter}        & \ad     & \yes & \yes & \no  & + FFN                    \\
\footnotesize{Intrinsinc SAID} \cite{intrinsic_said} & \rp & \yes  & \no & \no  & \gren{No overhead}            \\
LoRa              \cite{lora}                    & \rp     & \yes & \yes & \no  & \gren{No overhead}           \\
DoRA              \cite{liu2024dora}            & \rp      & \yes & \yes & \no  & \gren{No overhead}        \\
UniPELT           \cite{unipelt}                 & \ad \rp & \yes & \yes & \no  & + FFN and input          \\
\footnotesize{Compacter}         \cite{compacter}               & \ad \rp & \yes & \yes & \no  & + FFN                    \\
\footnotesize{PHM Adapter} \cite{compacter}      & \ad \rp & \yes & \yes & \no  & + FFN                    \\
KronA             \cite{krona}                   &     \rp & \yes & \yes & \no  & \gren{No overhead}           \\
KronA$^B_{res}$   \cite{krona}                   & \ad \rp & \yes & \yes & \no  & +  linear layer          \\
(IA)$^3$          \cite{t_few}                   & \ad     & \yes & \yes & \no  & + gating                 \\
Attention Fusion  \cite{attention_fusion}        & \ad     & \yes & \yes & \yes & + decoder         \\
LeTS              \cite{lets}                    & \ad     & \yes & \yes & \yes & + FFN                    \\
Ladder Side-Tuning \cite{ladder_side_tuning}     & \ad     & \yes & \yes & \yes & + decoder                \\
FAR               \cite{far_edge}                & \se     & \yes & \yes & \no  & \gren{No overhead}           \\
S4-model          \cite{design_spaces}           &\ad \rp \se& \yes & \yes & \no  & + FFN and input          \\
\bottomrule
\end{tabular}
\end{small}
\vspace{2pt}
\caption{Comparing \peft{} methods across storage (\textbf{disk}), memory (\textbf{RAM}), and computational efficiency in terms of reducing backpropagation costs (\textbf{BP}) and \textbf{inference overhead}. The \yes{} means that the method is more effective than full fine-tuning, the \no{} means that it is less efficient than full fine-tuning.\\Types: \textbf{\ad} -- additive, \textbf{\se}~--~selective, \textbf{\rp} -- reparametrization-based.
}
\label{tab:efficiency}
\end{table*}



\newcommand{\unsim}{\mathord{\sim}}


\begin{table*}[ht]
\centering
\begin{small}
\begin{tabular}{l|cc|ccc}
\toprule
\multirow{2}{*}{\textbf{Method}} & \multirow{2}{2cm}{\centering\textbf{\%Trainable parameters}} & \multirow{2}{2cm}{\centering\textbf{\%Changed parameters}} & \multicolumn{3}{c}{\textbf{Evaluated on}} \\
& & & \textbf{\textless1B} & \textbf{\textless20B} & \textbf{\textgreater20B} \\
\midrule
Adapters            & 0.1 - 6    & 0.1 - 6      & \yes & \yes & \yes \\
AdaMix              & 0.1 - 0.2  &0.1 - 0.2     & \yes & \no  & \no  \\
SparseAdapter       & 2.0        & 2.0          & \yes & \no  & \no  \\
BitFit              & 0.05 - 0.1 &0.05 - 0.1    & \yes & \yes & \yes \\
DiffPruning         & 200        & \textbf{0.5} & \yes & \no  & \no  \\
Fish-Mask           & 0.01 - 0.5 &0.01 - 0.5    & \yes & \yes & \no  \\
Prompt Tuning       & 0.1        & 0.1          & \yes & \yes & \yes \\
Prefix-Tuning       & 0.1 - 4.0  & 0.1 - 4.0    & \yes & \yes & \yes \\
IPT                 & 56.0       & 56.0     & \yes & \no  & \no \\
MAM Adapter         & 0.5        & 0.5          & \yes & \no  & \no  \\
Parallel Adapter    & 0.5        & 0.5          & \yes & \no  & \no  \\
Intrinsinc SAID     & 0.001 - 0.1& \textbf{100}          & \yes & \yes & \no  \\
LoRa                & 0.01 - 0.5 & \textbf{$\unsim$30}   & \yes & \yes & \yes \\
DoRA                & 0.01 - 0.5 & \textbf{$\unsim$30}   & \no  & \yes & \no \\
UniPELT             & 1.0        & 1.0      & \yes & \no  & \no  \\ 
Compacter           & 0.05-0.07  & \textbf{$\unsim$0.1} & \yes & \yes & \no  \\
PHM Adapter         & 0.2        & \textbf{$\unsim$1.0} & \yes & \no  & \no  \\
KronA               & 0.07       & \textbf{$\unsim$30.0}& \yes& \no  & \no  \\
KronA$^B_{res}$     & 0.07       & \textbf{$\unsim$1.0} & \yes & \no  & \no  \\
(IA)$^3$            & 0.02       & 0.02     & \no  & \yes & \no  \\
Ladder Side-Tuning  & 7.5        & 7.5      & \yes & \yes & \no  \\
FAR                 & 6.6-26.4   & 6.6-26.4 & \yes & \no  & \no  \\
S4-model            & 0.5        &\textbf{\textgreater 0.5} & \yes & \yes & \no  \\

\bottomrule
\end{tabular}
\vspace{2pt}
\end{small}
\caption{What model sizes PEFT methods have been evaluated on and their typical amount of trainable parameters as reported in the literature. For updated number of trainable parameters lookup Table \ref{tab:peft_comparison_dataset_average_tall}.}
\label{tab:scale_table}
\end{table*}


\section{A Deep Dive into PEFT}
\label{sec:methods}
In the following sections, we dive into the details of various parameter-efficient fine-tuning approaches. We describe the distinctions and tradeoffs between them in terms of the dimensions outlined in Section \ref{sec:peft_comparison_theory}. We \textbf{bold} a one-sentence summary of each method to simplify skimming.

In the method description, we also indicate whether it has been applied to models with fewer than 1 billion, 20 billion, or more than 20 billion parameters. We stick to indicating parameter counts where possible because the words ``small'' and ``large'' change their meaning too quickly. For a summary, refer to Table \ref{tab:scale_table}. Finally, we provide a brief pseudo-PyTorch implementation of the most important part of the algorithm where feasible.

\section{Additive methods: Adapters}
\label{sec:additive_section_adapters}
We begin our exploration of PEFT methods with one of the largest sub-families: methods that add fully-connected networks between model layers, known as adapters.

\subsection{Adapters}
\label{sec:adapters}

\citet{adapters} introduced the idea of adapters for NLP. Rebuffi et al. \cite{Rebuffi2017Adapters,Rebuffi2018EfficientPO} initially proposed similar concepts for image classification tasks, and Houlsby et al. extended this concept to NLP by proposing \textbf{the addition of fully-connected networks after attention and FFN layers} in Transformer. Unlike the transformer FFN block, Adapters usually have a smaller hidden dimension than the input. Adapters have demonstrated impressive parameter efficiency, showing that it is possible to achieve full fine-tuning performance by tuning less than 4\% of the total model parameters.

\begin{lstlisting}
def transformer_block_with_adapter(x):
    residual = x
    x = SelfAttention(x)
    x = FFN(x)  
    x = Adapter(x)  # Adapter after FFN
    x = LN(x + residual)
\end{lstlisting}

\citet{adapter_fusion} found that inserting the adapter only after the self-attention layer (after normalization) achieves similar performance as using two adapters per transformer block. 

\subsection{AdaMix}

AdaMix \cite{adamix} improves the performance of adapters by \textbf{utilizing multiple adapters in a mixture-of-experts (MoE) fashion} \cite{moe}. This means each adapter layer consists of a set of layers (experts), and for each forward pass, only a small set of experts is activated. In contrast to a regular MoE, which selects and weights multiple experts using a routing network, AdaMix randomly selects a single expert for each forward pass to minimize computational costs.

To stabilize training, the authors propose consistency regularization, which minimizes the symmetrized KL divergence between two models' forward passes with different sets of experts selected. Another difference from a regular MoE layer is that up and down projections of the adapter are selected independently. After training, the adapter weights are averaged across the experts to reduce inference costs.

\begin{lstlisting}
def transformer_block_with_adamix(x):
    residual = x
    x = SelfAttention(x)
    x = LN(x + residual)
    residual = x
    x = FFN(x)
    # adamix starts here
    x = random_choice(experts_up)(x)
    x = nonlinearity(x)
    x = random_choice(experts_down)(x)
    x = LN(x + residual)
    return x

def consistency_regularization(x):
    logits1 = transformer_adamix(x)
    # second pass uses different experts
    logits2 = transformer_adamix(x)
    r = symmetrized_KL(logits1, logits2)
    return r
\end{lstlisting}

Although AdaMix achieves better performance than regular adapters with the same inference cost, it can use more memory during training. \citet{adamix} show that AdaMix can use much smaller adapter hidden states than regular adapters, which amortizes trainable parameter overhead over the number of experts ($\sim$4-8). However, consistency regularization increases computational memory requirements, as it needs to keep two versions of the hidden states and gradients over two forward passes with different experts.


\section{Additive Methods: Soft Prompts}
\label{sec:additive_section_soft_prompts}

Prompting language models has demonstrated remarkable performance in zero- and few-shot scenarios \cite{brown2020language_gpt3,pet}. However, optimizing discrete natural language prompts or using in-context learning becomes impractical with many training examples. To overcome this challenge, the concept of ``soft'' or ``continuous'' prompts was proposed \cite{prefix_tuning,prompt_tuning,p_tuning}, converting the discrete optimization problem of finding the best "hard" prompt into a continuous one.

\subsection{Prompt Tuning}
\label{sec:prompt_tuning}

Prompt tuning \cite{prompt_tuning} proposes to \textbf{prepend the input embeddings}. These tensors are commonly referred to as ``soft prompts,'' and they are optimized directly through gradient descent.


\begin{lstlisting}
def prompt_tuning_attention(input_ids):
    q = x @ W_q
    k = cat([s_k, x]) @ W_k  # prepend a
    v = cat([s_v, x]) @ W_v  # soft prompt
    return softmax(q @ k.T) @ V
\end{lstlisting}

Ablation studies by \citet{promspt_mapping} on prompt length (1 to 150 tokens) and model size (10M to 11B parameters) reveal that prompt tuning becomes more parameter efficient as the model size increases. For instance, prompt tuning of T5-11B achieves the same SuperGLUE \cite{wang2019superglue} performance with either 5 or 150 soft prompt tokens.

Furthermore, efficiency increases more rapidly with model size. T5-large performance saturates at prompt length 20 or 20K trainable parameters ($0.002\%$), and T5-XL performance saturates at prompt length 5, also 20K trainable parameters ($0.0002\%$). However, prompt tuning only becomes comparable with full fine-tuning at the 10B model scale. Additionally, increasing sequence length by 20-100 tokens can significantly increase computation, given the quadratic complexity of the transformer. Overall, soft prompts are incredibly parameter-efficient but come with inference overhead and are more applicable to larger models.

\subsection{Prefix Tuning}
\label{sec:prefix_tuning}

\citet{prefix_tuning} independently develop the idea of soft prompts with a distinctive flavor: \textbf{shared trainable parameters are prepended to the hidden states of all layers}. The same prefix $\mathbf{P}_\theta \in \R^{l \times h}$ is prepended to all of the hidden states. They observe that directly optimizing the soft prompt leads to instabilities during training. Instead, soft prompts are parameterized through a feed-forward network $\mathbf{P}_\theta = \operatorname{FFN}(\hat{\mathbf{P}}_\theta)$. During training, $\hat{\mathbf{P}}_\theta$ and the parameters of the FFN are optimized. After training, only $\mathbf{P}_\theta$ is needed for inference, and the FFN can be discarded.

Pseudocode for a single layer:
\begin{lstlisting}
def transformer_block_for_prefix_tuning(x):
    soft_prompt = FFN(soft_prompt)
    x = concat([soft_prompt, x], dim=seq)
    return transformer_block(x)
\end{lstlisting}

Note that the approach is very similar to Prompt Tuning (Section \ref{sec:prompt_tuning}), but the soft prompts are added in each layer.

In their experiments, \citet{prefix_tuning} apply prefix tuning to the BART \cite{lewis2019bart} model (less than 1B parameters) for different generation tasks and show a performance close to full fine-tuning by training only $0.1\%$ of the parameters. Soft prompt lengths used in the study vary from 10 to 200 tokens.

\subsection{Adaptive Prefix Tuning}

\citet{zhang2023towards} extends prefix tuning and proposes adaptive changes to the prompt length. The authors achieve the \textbf{adaptive adjustment via gating and scaling of the input to each layer}. Specifically, the input to each transformer layer is first passed through an FFN ($W$) with sigmoid activation. The sigmoid-activated values act as gates for each pseudo token added in the soft prompt, providing a structure to adaptively select or deselect certain soft tokens in certain layers. Furthermore, the sigmoid-activated values are scaled with separate $\lambda$ parameters. $W$ and $\lambda$ are both learnable parameters and are separately defined for each layer in the language model. Hence, $W$ and $\lambda$ cause significant computation, memory, and storage overhead on top of the prefix-tuning approach \cite{prefix_tuning}.

\begin{lstlisting}
def adaptive_adjustment(x, layer_index):
    alpha = sigmoid(W[layer_index, ...] @ x)
    adaptive_factor = lambda_[layer_index] * alpha
    return adaptive_factor

def prefix_tuning(x, layer_index):
    # adaptive gating and scaling
    adaptive_factor = adaptive_adjustment(x, layer_index)
    
    # soft prompt
    soft_prompt = FFN(soft_prompt)
    soft_prompt *= adaptive_factor

    # regular prefix-tuning
    x = concat([soft_prompt, x], dim=seq)
    return transformer_block(x)
\end{lstlisting}

However, the overhead is shown to be beneficial in the BERT and DeBERTa series models. Adaptive prefix tuning consistently outperforms the standard prefix tuning approach and even surpasses the full fine-tuning performance in most cases.

\subsection{Intrinsic Prompt Tuning (IPT)}
Prompt tuning is slow to converge.
A few studies \cite{prompt_mapping,spot} have proposed pre-training soft prompts to improve performance and convergence speed. However, these methods do not provide solutions to reduce the number of parameters per task.


\citet{ipt} hypothesize that the high-dimensional space used to define a soft prompt contains a low-dimensional ``intrinsic task subspace'' and \textbf{learn it using an autoencoder in a multi-task fashion}.

The IPT method works in three steps. First, given a set of training tasks, their soft prompts are learned in the standard way (Section \ref{sec:prompt_tuning}). Then, these prompts are used to train an autoencoder that compresses their dimensionality. After this, the encoder part is discarded, and only the input to the autoencoder decoder is trained on new tasks.

\begin{lstlisting}
def autoencoder(soft_prompt):
    soft_prompt = soft_prompt.flatten()
    P = FFN_A(soft_prompt)  # encoder
    P = FFN_B(P)            # decoder
    P = P.reshape(prompt_len, hidden)
    return P

def ipt_model(x):
    P = FFN_B(intrinsic_subspace_prompt)
    P = P.reshape(prompt_len, hidden)
    x = concat([P, x], dim=seq)
    return model(x)
\end{lstlisting}

Even though the IPT framework reduces the number of parameters for the unseen tasks, this reduction comes at the price of training the autoencoder. The authors conduct experiments with the BART-base model and a prompt length of 100. The resulting autoencoder, which is implemented\footnote{\href{https://github.com/thunlp/Intrinsic-Prompt-Tuning/blob/master/bartPrompt.py}{github.com/thunlp/Intrinsic-Prompt-Tuning}} as a fully-connected network that accepts a one-dimensional tensor of size $76800$, reaches 78 million parameters. This constitutes over 56\% of the total parameters in the BART-base model. Therefore, significantly more efficient methods of prompt autoencoding are required to make IPT practically applicable.

\section{Additive Methods: Other Approaches}
\label{sec:additive_section_others}


\subsection{Ladder-Side Tuning (LST)}
\label{sec:lst}

Ladder-Side Tuning \cite{ladder_side_tuning} \textbf{trains a small transformer network on the side of the pre-trained network}. This side network combines the hidden states of the pre-trained backbone network with its own hidden states.

This way, the side network only uses the pre-trained model as a feature extractor, and backpropagation must only be computed through the side network, saving on both memory and compute during training.
To improve the performance and parameter efficiency of LST, the side network is initialized from the structurally pruned pre-trained model parameters and uses half as many layers.

Here \texttt{h\_pt} is the output of the corresponding layer of the pre-trained network, and \texttt{alpha} is an input-independent trainable scalar gate:

\begin{lstlisting}
def ladder_side_layer(x, h_pt):
    h_pt = h_pt @ W_down  # to x.shape
    gate = sigmoid(alpha)
    x = gate * x + (1 - gate) * h_pt
    return transformer_block(x)

def ladder_side_network(x):
    with no_grad():
        H_pt = pretrained_network(x, return_all_hiddens=True)
    for i in range(layers):
        layer = ladder_side_layers[i]
        x = layer(x, H_pt[i])
    return x
\end{lstlisting}

LST demonstrated a three-fold RAM reduction in fine-tuning T5-Base compared to full fine-tuning and a two-fold RAM usage reduction compared to LoRA (Section \ref{sec:lora}) with a small degradation in accuracy. Moreover, LST outperforms these methods when controlling for RAM usage.

\subsection{(IA)$^3$}
\label{sec:ia3}

\citet{t_few} proposes a new parameter-efficient method to multi-task fine-tune T0 \cite{t_few}. Their proposed fine-tuning method, (IA)$^3$, learns new parameters $l_v$, $l_k$, and $l_{ff}$, which \textbf{rescale key, value, and hidden FFN activations}. Specifically:

\begin{lstlisting}
def transformer_block_with_ia3(x):
    residual = x
    x = ia3_self_attention(x)
    x = LN(x + residual)
    residual = x
    x = x @ W_1         # FFN input
    x = l_ff * gelu(x)  # (IA)3 scaling
    x = x @ W_2         # FFN output
    x = LN(x + residual)
    return x

def ia3_self_attention(x):
    k, q, v = x @ W_k, x @ W_q, x @ W_v
    k = l_k * k
    v = l_v * v
    return softmax(q @ k.T) @ V
\end{lstlisting}

Training only these three vectors, $l_v$, $l_k$, and $l_{ff}$, for each transformer block leads to high parameter efficiency. For T0-3B, it only updates about $0.02\%$ of model parameters and outperforms other methods, including Compacter (Section \ref{sec:compacter}), which has a similar parameter count, and LoRA (Section \ref{sec:lora}), which has 16 times more trainable parameters. Unlike adapter-tuned models, (IA)$^3$-tuned models exhibit minimal overhead. Vectors $l_v$ and $l_k$ can be integrated into the corresponding linear layers, and the only overhead comes from $l_{ff}$.
\section{Selective Methods}
\label{sec:selective_methods}
Selective methods fine-tune a subset of the existing parameters of the model. It could be a layer depth-based selection, layer type-based selection, or even individual parameter selection.

\subsection{BitFit}
\label{sec:bitfit}

\citet{bitfit} proposes to \textbf{only fine-tune the biases of the network}. That is, for every layer, $W$ is unchanged and only $b$ is trained.

BitFit only updates about 0.05\% of the model parameters. The original paper demonstrated that the method achieves similar or better performance than full fine-tuning in low- and medium-data scenarios for BERT models (less than 1B parameters). Further research showed that for models larger than 1B parameters, BitFit significantly underperforms compared to full fine-tuning and other PEFT methods \cite{t_zero,t_few,lora}.

\begin{lstlisting}
  params = (p for n, p in model.named_parameters()
            if "bias" in n)
  optimizer = Optimizer(params)
\end{lstlisting}

\paragraph{Bias-less architectures} In our experimental comparison, we noticed that several popular architectures do not use bias terms in the network. For example, T5 only uses bias terms in the relative attention weights, and LLaMA does not use bias terms throughout the network.


\subsection{DiffPruning}
\label{sec:diff_pruning}

DiffPruning \cite{diff_pruning} aims to achieve parameter efficiency by learning a sparse update of a neural network's weights. The method introduces a learnable binary mask on the weights, denoted by $\delta = z \circ \Delta W$, where $\circ$ represents the Hadamard product. This \textbf{parameter mask is learned during model fine-tuning} as part of the regularization objective, which is a differentiable approximation to the $L_0$ norm of the update vector $\delta$.

DiffPruning achieves comparable performance to full fine-tuning while modifying only 0.5\% of the model parameters in \textless1B scenarios. This makes it a useful method for multi-task deployment in edge (mobile) scenarios where storage is limited. However, this method requires more memory than traditional fine-tuning, as it involves optimizing all parameters during training in addition to the learnable binary mask.

\subsection{Freeze and Reconfigure (FAR)}
\label{sec:far}

FAR \cite{far_edge} \textbf{selects columns of parameter matrices to train and reconfigures linear layers into trainable and frozen components}. The method operates in two stages. In the first stage, the most important rows of parameter matrices are identified for updating. This process is similar to structured pruning and can use any pruning method. In their paper, the authors fine-tune the model on a percentage of the data and select the top-$r$ rows based on the $L_1$ distance between the fine-tuned and original models.

In the second stage, the network is reconfigured by splitting each parameter matrix $W \in \mathbb{R}^{in \times h}$ into a trainable component $W_{t} \in \mathbb{R}^{in \times h'}$ and a frozen component $W_{f} \in \mathbb{R}^{in \times (h - h')}$, where $h'$ is the desired number of trainable parameters. The matrix multiplications with $W_{t}$ and $W_{f}$ are computed independently, and the results are concatenated. A similar operation is performed on biases.
\vspace{1.5em}  

\begin{lstlisting}
def far_layer(x):
    h1 = x @ W_t
    h2 = x @ W_f
    return concat([h1, h2], dim=-1)
\end{lstlisting}

While this approach creates additional compute overhead during training, it provides great flexibility in terms of parameter selection on modern hardware using standard frameworks like PyTorch. After training, the parameters can be reconfigured back, removing any inference overhead.

The original paper focused on edge scenarios and used DistilBERT (66M parameters) for their experiments. It was only applied to feed-forward layers, as these make up the majority of the DistilBERT parameters. FAR achieved similar performance to fine-tuning on GLUE and SQuAD 2.0 \cite{squad2} while updating only 6\% of the parameters.

\subsection{FishMask}

FishMask \cite{fish_mask} is a \textbf{sparse fine-tuning method that selects the top-p parameters of the model based on their Fisher information}. Fisher information is estimated in a common way through a diagonal approximation.
\begin{equation*}
    \hat{F_{\theta}} = \frac{1}{N} \sum_{i = 1}^{N} \mathbb{E}_{y \sim p_{\theta}(y|x_i)}(\nabla_{\theta} \log p_{\theta}(y|x_i))^2
    \label{eqn:fisher_approximation}
\end{equation*}

Or in pseudocode:
\begin{lstlisting}
  sparsity = 0.99
  N = len(data)
  for x, y in data:
      loss = loss_fn(model(x), y)
      loss.backward()
      for n, p in model.named_params():
          fisher[n] += p.grad ** 2 / N
  
  threshold = percentile(fisher, sparsity)
  masks = {n: f > threshold
           for n, f in fisher.items()}
\end{lstlisting}

FishMask requires computing gradients for all parameters on several batches of the data. However, after the highest-Fisher parameters are selected, only they need to be optimized.

The method generally performs on par with adapters but sub-par to LoRA and (IA)$^3$ (Sections \ref{sec:lora} and \ref{sec:ia3}). It has been evaluated on BERT (less than 1B parameters) and T0-3B models. However, FishMask is computationally intensive and inefficient on contemporary deep learning hardware due to the lack of support for sparse operations.

\subsection{DiffFit}

\citet{xie2023difffit} combines various selective approaches with additive style scaling and specifically targets tuning of the diffusion model. In their study, \citet{xie2023difffit} \textbf{tunes the bias, layer normalization, and embedding parameters and adds new scaling parameters for residual mixing}. The scaling parameters are added separately for the self-attention and the FFN layers. The pseudocode for the scaling is as follows:

\begin{lstlisting}
def forward(x, c, t):
    x += gamma_1 * self_attn(x, c, t)
    x += gamma_2 * FFN(x, c, t)
    return x
\end{lstlisting}

The authors conduct experiments focused on diffusion models and show that DiffFit outperforms other PEFT approaches. In addition, due to the minor overhead of the scaling parameters, DiffFit provides faster tuning speed and minimal degradation in inference and storage efficiency.

\section{Reparametrization-based methods}
\label{sec:reparametrization_based_methods}

These methods use the idea of reparametrizing the weights of the network using a low-rank transformation. This decreases the trainable parameter count while still allowing the method to work with high-dimensional matrices, such as the pre-trained parameters of the networks.

\subsection{Intrinsic SAID}
\label{sec:intrinsic_said}

In their work, \citet{intrinsic_said} investigates the intrinsic dimensionality of fine-tuning and demonstrates that this process can be performed in a low-rank subspace. Specifically, they use the \textbf{Fastfood transform to reparametrize the update to the model weights}. Fastfood is a compute-efficient dimensionality expansion transform $F:\R^d \rightarrow \R^D$ that can be done in~$O(D \log d)$ time and $O(D)$ memory.

They show that larger models require changes in a lower-rank subspace compared to smaller models to achieve the same fine-tuning performance. This observation motivates both scaling large models and parameter-efficient fine-tuning. It is important to note that, unlike methods that select a particular subset of parameters for fine-tuning, Intrinsic SAID updates all model parameters in a low-rank manner, i.e., $\theta = \theta_0 + F(\theta^d)$, where $\theta_0 \in \R^D$ denotes the pre-trained model parameters and $\theta^d \in \R^d$ denotes the parameters to be optimized. Therefore, while the number of optimizable parameters is low, the $O(D)$ memory complexity of Fastfood and the update to all of the model's parameters make Intrinsic SAID impractical for fine-tuning large networks. For more details on Fastfood, we refer the reader to the original paper by \citet{Le2013FastfoodAK}.


\subsection{LoRA}
\label{sec:lora}

LoRA \cite{lora} takes inspiration from Intrinsic SAID and proposes a simpler way to perform low-rank fine-tuning. \textbf{Parameter update for a weight matrix in LoRA is decomposed into a product of two low-rank matrices:}
\begin{equation*}
\begin{aligned}
    \delta W = W_A W_B, \ W_A \in \R^{\text{in} \times r}, W_B \in \R^{r \times \text{out}}.\\
\end{aligned}
\end{equation*}

All pre-trained model parameters are kept frozen, and only $W_A$ and $W_B$ matrices are trainable. The scaling factor is constant and typically equals $\frac{1}{r}$. After training, they can be integrated into the original $W$ by just adding the matrix $W_A W_B$ to the original matrix $W$.

Pseudocode is very simple:
\begin{lstlisting}
def lora_linear(x):
    h = x @ W          # regular linear
    dh = x @ W_A @ W_B # low-rank update
    h += scale * dh    # scaling
    return h
\end{lstlisting}

In Transformers, LoRA is typically used for $W_K$ and $W_V$ projection matrices in multi-head attention modules. However, to achieve the best possible performance, it is best to apply LoRA to all weight matrices in the model \cite{dettmers2023qlora}. The method outperforms BitFit and Adapters and has been evaluated on models with up to 175B parameters.


\subsection{KronA}
\label{sec:krona}
KronA \cite{krona} replaces matrix factorization $\delta W \mathrel{\mkern-10mu}=\mathrel{\mkern-10mu}W_A W_B$ in LoRA with a \textbf{matrix factorization through a Kronecker product $\delta W \mathrel{\mkern-5mu} = \mathrel{\mkern-5mu} W_A \otimes W_B$}.

This yields a better rank-to-parameters tradeoff because the Kronecker product preserves the rank of the original matrices being multiplied. Or, in other words, $\operatorname{rank}(A \otimes B) = \operatorname{rank}A \cdot \operatorname{rank}B$. Additionally, \citet{krona} uses an efficient Kronecker product-vector product operation $x (A \otimes B)$, which avoids representing $\delta W$ explicitly and leads to significant speedups. KronA$^B_{res}$, also presented in \citet{krona}, is a parallel adapter that uses Kronecker product parameterization of the weights and includes a residual connection.

Krona pseudocode:
\begin{lstlisting}
def krona_linear(x):
    x = x @ W  # regular linear
    x += kronecker_vector_prod(x, W_A, W_B)
    return scale * x

# same as x @ kronecker_product(A, B)
def kronecker_vector_prod(x, A, B):
    x = x.reshape(A.shape[1], B.shape[1])
    x = A.T @ x @ B
    return x.reshape(-1)
\end{lstlisting}

On GLUE, KronA methods perform on par or better than adapters (Section \ref{sec:adapters}), LoRA (Section \ref{sec:lora}), and Compacter (Section \ref{sec:compacter}) at the same trainable parameter count of $0.07\%$, while being significantly faster than adapters or Compacter at inference time. The method was evaluated only on small models (less than 1B parameters).

\paragraph{Background: Kronecker product}
Kronecker product is a tensor operation defined as
\begin{equation}
\begin{aligned}
\mathbf{A} \otimes \mathbf{B} &: \R^{n \times m} \times \R^{k \times l} \rightarrow \R^{nk \times ml}\\
\mathbf{A} \otimes \mathbf{B} &= 
    \begin{bmatrix}
    a_{1,1} \mathbf{B} & \cdots & a_{1,n} \mathbf{B} \\
    \vdots & \ddots & \vdots \\
    a_{m,1} \mathbf{B} & \cdots & a_{m,n} \mathbf{B}
    \end{bmatrix}
\end{aligned}
\end{equation}

\vspace{1em}

It can be easily implemented \footnote{Source: \url{github.com/rabeehk/compacter}} in PyTorch using the command {\small{ \texttt{torch.einsum} }}
\begin{lstlisting}
def batched_kronecker_product(a, b):
    bs, i, j = a.shape
    bs, k, m = b.shape
    res = einsum("bij,bkm->bikjm", a, b)
    return res.view(bs, i * k, j * m)
\end{lstlisting}

\subsection{DoRA}
In the study conducted by \citet{liu2024dora}, the authors examine the differences between full fine-tuning and LoRA fine-tuning. Particularly, the authors analyze the change in direction ($\Delta D$) and magnitude ($\Delta M$) of the weight matrices, comparing pre- and post-full fine-tuning checkpoints. With values of $\Delta D$ and $\Delta M$ calculated for various checkpoints and multiple layers, the authors show the existence of an inverse relationship between $\Delta D$ and $\Delta M$ for full fine-tuning. However, for LoRA tuning, the $\Delta D$ and $\Delta M$ are directly proportional to each other. Hence, the authors propose DoRA, a method that \textbf{decouples magnitude from the rest of the weight update}, similar to weight normalization \cite{weight_normalization}. The implementation of DoRA is as follows:

\begin{lstlisting}
def initialize_m(W):
    # learnable scaling factor (magnitude component)
    m = Parameter(l2_norm(W), requires_grad=True)
    return m

def dora_linear(x):
    W_cur = W + (W_A @ W_B)
    v = W_cur / l2_norm(W_cur)  # normalize weight
    W_cur = m * v               # mult. by learnable magnitude

    # regular forward pass
    return x @ W_cur + b
\end{lstlisting}

The authors theoretically show that with the proposed scaling, an inverse relationship between $\Delta D$ and $\Delta M$ can be achieved with LoRA-style fine-tuning. In other words, DoRA can achieve either large directional changes with small adjustments to the magnitude, or vice versa. Experiments conducted on various tasks and models show consistent improvements in DoRA over LoRA. Interestingly, DoRA improves sample efficiency on instruction tuning tasks and shows significant performance improvements in the low-rank (4, 8) regime over LoRA. 

\subsection{GLoRA}
In LoRA, the pre-trained weights are varied only through addition. While effective, more difficult tasks might benefit from further scaling and activation. Following the same hypothesis, \citet{chavan2023one} proposes a variation of LoRA called GLoRA (short for Generalized-LoRA). In GLoRA, the authors improve the capability of LoRA at the cost of \textbf{more learnable parameters that scale and shift parameters or activations or both}. However, all learnable parameters of GLoRA are merged back into the frozen model, hence incurring no additional inference cost. The GLoRA update can be written as follows: 

\begin{equation*}
    f(x) = (W_{0} + W_{0}A + B)x + CW_{0} + Db_{0} + E + b_0 
\end{equation*}

Where, $W_0$ and $b_0$ are pre-trained parameters and are kept frozen during the fine-tuning process. The GLoRA parameters, $A, B, C, D$, and $E$ are learnable and are tuned during fine-tuning. In the above expression, the matrix $B$ replicates the LoRA update. All other learnable parameters i.e., $A, C, D$, and $E$ hence, add an overhead for fine-tuning compared to LoRA. However, to reduce this overhead, matrices $A, B$ and $C$ can be implemented with low-rank approximation i.e., $A = A_{in} \cdot A_{out}$ such that $A_{in} \in \mathbb{R}^{d_1 \times r}$ and $A_{out} \in \mathbb{R}^{r \times d_2}$, for $A \in \mathbb{R}^{d_1 \times d_2}$. The authors also specify that all new parameters ($A$ to $E$) can be converted into vectors or scalars to manage the computational budget.  

\begin{lstlisting}
def glora_linear(x):
    # adjustment to the input
    W_cur = W_0 + AW_0 + B 
    h = x @ W_cur + CW_0

    # add bias (mostly not present in recent transformer LLMs)
    if bias_present:
        h += Db_0 + E + b_0

    return h
\end{lstlisting}

In the above GLoRA update, all learnable parameters (A, B, C, D, and E) can merge back into the original model after fine-tuning. 
Across various suites of vision and language tasks, the authors show that GLoRA outperforms LoRA. When the number of trainable parameters is held constant for both LoRA and GLoRA, GLoRA outperforms LoRA on the VTAB-1K benchmark. Additionally, GLoRA shows better sample efficiency than LoRA and other PEFT methods with a comparable number of trainable parameters.

\subsection{AdaLoRA}

In AdaLoRA, Zhang et al. \cite{zhang2023adaptive} introduce an adaptive method to reduce the ranks of the $W_A$ and $W_B$ matrices in LoRA. \textbf{They reformulate the LoRA update in an SVD-like format}: \begin{equation*} W = W + W_A\ \Lambda\ W_B \end{equation*} where $\Lambda$ is a diagonal matrix representing singular values. To adjust the rank, the authors prune $\Lambda$, affecting the dimensions of $W_A$ and $W_B$ based on importance scores that reflect the impact on loss. These scores are smoothed exponentially to guide parameter elimination. The method includes regularization to preserve orthogonality between $W_A$ and $W_B$. Their experiments with DeBERTa-base and BART-large demonstrate that AdaLoRA outperforms LoRA and other parameter-efficient fine-tuning (PEFT) methods under a parameter budget constraint. The study highlights that the initial transformer layers do not need high-rank updates. However, it requires tracking additional variables (namely $I, \Bar{I}, \Bar{U}$, and $S$, refer to \cite{zhang2023adaptive}) to prune $\Lambda$ values. This can induce significant memory overhead as the language model size or the rank is scaled. Nonetheless, AdaLoRA presents insightful findings that motivate the need to develop a selective ranking strategy for LoRA that can function without significant memory overhead. We present a simplified pseudocode of the main operations in AdaLoRA as follows,

\begin{lstlisting}
def prune(W_A, ..., dW_A, ..., k):
    # calculate the importance score
    s = importance_score(W_A, ..., dW_A, ...)

    # find bottom-k values
    indices = find_bottom_k(s, k)

    # prune
    W_A[indices, :] = W_A[indices, :].detach()
    W_B[:, indices] = W_A[:, indices].detach()
    lambda_[indices] = 0
    lambda_[indices] = lambda_[indices].detach()
    return W_A, W_B, lambda_

def ada_lora_linear(x):
    h = x @ W                    # regular linear
    h += x @ W_A @ lambda_ @ W_B # low-rank update
    return h

\end{lstlisting}


\subsection{GaLore}

Usually PEFT methods improve computational efficiency by reducing the number of trainable parameters at the cost of lowering the expressivity. \citet{zhao2024galore} propose to \textbf{train all model's parameters, but using low-rank regularization of gradients}. 
In essence, GaLore makes full-tuning efficient by lowering the memory consumption of optimizer states (first and second momentum).
The authors additionally provide a theoretical justification of the method.


Based on the theoretical analysis, the authors propose GaLore with the following salient properties: 1. No low-rank approximation of parameter matrices; 2. Decomposition of gradients in lower dimensional space; 3. Gradient regularization in lower dimensional space; 4. Projection of the regularized gradient in the original space and updating parameters. GaLore achieves a reduction in memory benefits by reducing storage requirements for the gradient statistics (first- and second-order momentum) and by making the gradient regularization leaner. 

\begin{lstlisting}
def galore_update(w, grad, lr):
    u, s, v = SVD(grad)
    m, n = u.shape[-1], v.shape[0]
    p = u if m < n else v
    grad_ = p.T @ grad
    # update optimizer momentum
    m, v = update_momentum(grad_)
    # regularize grad
    g = m / (sqrt(v) + eps)
    # project back
    grad = alpha * (p @ g)
    return  w - lr * grad  # param update

def update_momentum(grad_):
    # first order
    m = beta_1 * m + (1 - beta_1) * grad_
    m \= (1 - beta_1**t)
    # second order
    v = beta_2 * m + (1 - beta_2) * grad_**2
    v \= (1 - beta_2**t)
    return m, v
\end{lstlisting}

The authors evaluate the proposed method for pre-training as well as fine-tuning. On the pre-training end, GaLore achieves better validation perplexity on C4 data compared to LoRA \cite{lora} and ReLoRA \cite{relora} for a range of model sizes varying from 60M to 1B. Compared to full-rank pre-training GaLore achieves comparable but slightly worse perplexity values while significantly reducing memory usage. For the fine-tuning end, the authors fine-tuned RoBERTa-base on GLUE tasks and highlighted the benefit of GaLore over LoRA.

\subsection{Quantization and LoRA}
In LoRA, the language model is held frozen in full precision. Due to the increasing size of language models, the frozen language model causes a significantly large memory footprint. Hence, recent studies have focused on reducing the memory load of backbone LLMs by reducing the precision of the backbone LLM parameters.

A study conducted by \citet{dettmers2023qlora} addresses this issue with QLoRA. The authors achieve memory benefits in three ways. First, the authors introduce a 4-bit NormalFloat quantization method that reduces the memory required to store the backbone model. Second, the authors reduce the memory footprint of quantization constants by introducing double quantization, i.e., quantization of the quantization constants. Both of the first two steps realize significant memory gains in storing the backbone model on GPU. Lastly, the authors implement a paging mechanism between the CPU and GPU for optimizer states, offloading optimizer states to the CPU in case of memory spikes. With extensive experiments, the authors highlight the performance benefits of their proposed quantization approach.

However, post-training quantization of the final LoRA-tuned model often results in performance discrepancies \cite{xu2023qa}. In another study, \citet{xu2023qa} presents a group-wise quantization strategy that overcomes the need for post-training quantization, hence leading to better performance, even in lower precision regimes. More studies have proposed better quantization-aware fine-tuning methods in recent literature, e.g., IR-QLoRA \cite{qin2024accurate}, LoftQ \cite{li2023loftq}, L4Q \cite{jeon2024l4q}, LQ-LoRA \cite{guo2023lq}.





\section{Hybrid Approaches}
\label{sec:hybrid_section}


Hybrid methods leverage the strengths of various techniques while mitigating their weaknesses to improve the performance and efficiency of PEFT.

\subsection{SparseAdapter}
\label{sec:sparse_adapter}

\citet{sparse_adapter} propose a \textit{Large-Sparse} strategy to train adapter layers. In this strategy, they use a \textbf{large hidden dimension for the added module and prune around 40\% of the values at initialization}. \textit{Large-Sparse} consistently outperforms its non-sparse counterpart with the same trainable parameter count. However, training and inference costs can be higher depending on hardware support for sparse tensors and operations. It is also worth noting that computing the pruning mask for this method may require obtaining gradients for all newly added parameters.

\subsection{MAM Adapters}
\label{sec:mam_adapter}

In their study, \citet{parallel_adapter} thoroughly investigated adapter placement and soft prompts. They concluded that scaled parallel adapters outperform sequentially-placed adapters and that placing an adapter in parallel to FFN outperforms multi-head attention-parallel adapters.
They also noticed that soft prompts can efficiently modify attentions by only changing 0.1\% of the parameters and propose to 'mix-and-match' (MAM) these ideas. Their final model, \textbf{MAM Adapter, is a combination of scaled parallel adapter for the FFN layer and soft prompt}.
\begin{lstlisting}
def transformer_block_mam(x):
    x = concat([x, soft_prompt], dim=seq)
    residual = x
    x = SelfAttention(x)
    x = LN(x + residual)
    x_a = FFN(x) # parallel adapter
    x_a = scale * x_a
    x = LN(x + x_adapter)
    return x
\end{lstlisting}
The MAM method outperforms BitFit and PromptTuning by a large margin and consistently outperforms LoRA (Section \ref{sec:lora}), Adapters (Section \ref{sec:adapters}), and Prefix Tuning (Section \ref{sec:prefix_tuning}) with a 200 soft prompt length and 7\% extra parameters. The experiments were conducted on models with fewer than 1 billion parameters.

It is worth noting that parallel adapters were independently studied by \citet{parallel_adapter2} in the domain of machine translation.




\subsection{UniPELT}
\label{sec:unipelt}
UniPELT \cite{unipelt} is a \textbf{gated combination of LoRA, Prefix-tuning, and Adapters}. LoRA reparametrization is used for $W_Q$ and $W_V$ attention matrices, prefix-tuning is applied to keys and values of each layer, and adapters are added after the feed-forward layer of the transformer block. For each of the modules, gating is implemented as a linear layer projecting the module input into a dimension of size one, sigmoid activation, and averaging the resulting vector over the sequence length. Trainable parameters include LoRA matrices $W_A, W_B$, prompt tuning parameters $P_q, P_k$, adapter parameters, and gating function weights.

Schematic implementation of UniPELT (omitting attention heads for simplicity):
\begin{lstlisting}
def transformer_block_with_unipelt(x):
    residual = x
    x = unipelt_self_attention(x)
    x = LN(x + residual)
    residual = x
    x = FFN(x)
    adapter_gate = gate(x)
    x = adapter_gate * FFN(x)
    x = LN(x + residual)
    return x

def unipelt_self_attention(x):
    k, q, v = x @ W_k, x @ W_q, x @ W_v
    # lora for queries and values
    lora_gate = gate(x)
    q += lora_gate * W_qA @ W_aB
    v += lora_gate * W_vA @ W_vB
    # prefix tuning
    pt_gate = gate(x)
    q_prefix = pt_gate * P_q
    k_prefix = pt_gate * P_k
    return softmax(q @ k.T) @ V

def gate(x):
    x = Linear(x)
    x = sigmoid(x)
    return mean(x, dim=seq)
\end{lstlisting}

UniPELT demonstrates significant improvements over individual LoRA, Adapters, and Prefix Tuning approaches in low-data scenarios with only 100 examples. In higher data scenarios, UniPELT performs on par or better than these approaches. \citet{unipelt} reports that UniPELT uses 1.3\% trainable model parameters on BERT models with fewer than 1 billion parameters.

\subsection{Compacter}
\label{sec:compacter}
Compacter \cite{compacter} \textbf{utilizes the Kronecker product, low-rank matrices, and parameter sharing across layers to produce adapter weights}. Each parameter $W$ in an adapter is equal to the sum of Kronecker products
\begin{equation}
\begin{aligned}
    \hat W &= \sum_{i=0}^{n} A_i \otimes B_i \\
    \hat W \in \R^{k \times d}, \text{ } &A_i \in \R^{n \times n}, \text{ } B_i \in \R^{\frac{k}{n} \times \frac{d}{n}}.
\end{aligned}
\end{equation}
A linear layer $x \hat W + b$ with this parameterization is called a parametrized hypercomplex multiplication (PHM) layer \cite{pha}. Compacter takes this idea further, parametrizing $B_i$ similar to LoRA (Section \ref{sec:lora}) as $B_i = B_i^{\text{down}} B_i^{\text{up}}$, where all matrices are of rank at most $r$. Matrices $A_i$ are shared across all adapter layers for further parameter efficiency. The corresponding layer is referred to as Low-rank PHM (LPHM).
Note that all $A_i$ and $B_i$ tensors are 3D tensors with the first dimension equal to $n$, the number of Kronecker products in the PHM layer.
Compacter layer pseudocode:
\begin{lstlisting}
def compacter(x):
    x = LPHM(x)  # Essentially an FFN
    x = gelu(x)  # but
    x = LPHM(x)  # LPHM replaces linear
    return x

def lphm_forward(x):
    B = B_d @ B_u
    W = batched_kronecker_product(A, B)
    W = sum(W, dim=0)
    return x @ W + b
\end{lstlisting}

Compacter comes in two flavors: two adapters per transformer block or a single adapter after a feedforward layer (Compacter++). With only 0.05\% additional parameters, Compacter++ performs on par or better than adapters with 0.8\% additional parameters.
The model has been evaluated on T5 Base (less than 1B parameters) and T0-3B models.

\subsection{S4}
\label{sec:s4}
\citet{design_spaces} conducted a search of various \textbf{combinations of parameter-efficient fine-tuning techniques}. Their search space includes dividing consecutive layers into four uneven groups, allocating varying amounts of trainable parameters to each layer, determining which groups to fine-tune, and which PEFT methods to apply to each group.

Their proposed method, S4, divides layers into four groups ($G_{1,2,3,4}$) using a ``spindle'' pattern: more layers are allocated to the middle groups and fewer to the top and bottom groups. All groups are trainable, with trainable parameters uniformly allocated across the layers within each group. Different combinations of PEFT methods are applied to different groups. Specifically:
\begin{equation}
\begin{aligned}
G_1 &: A, L & G_3 &: A, P, B \\
G_2 &: A, P & G_4 &: P, B, L \\
\end{aligned}
\end{equation}

where A stands for Adapters (Section \ref{sec:adapters}), P for Prefix-Tuning (Section \ref{sec:prefix_tuning}), B for BitFit (Section \ref{sec:bitfit}), and L for LoRA (Section \ref{sec:lora}).

The search experiments were conducted on the T5-base model and the GLUE dataset with 0.5\% trainable parameters. The S4 method was then applied to T5-3B, RoBERTa, and XL-Net, consistently outperforming individual BitFit, Prefix Tuning, LoRA, and Adapters across different architectures, model sizes, and tasks.

\section{Comparison of PEFT Methods}
\label{sec:peft_comparison_theory}

Parameter efficiency involves multiple aspects: storage, memory, computation, and performance. However, achieving parameter efficiency alone does not necessarily lead to reduced RAM usage or faster training. When evaluating PEFT methods, it is important to consider various aspects of performance and efficiency, yet many publications focus solely on downstream performance and the number of parameters used by the PEFT method.

The ``number of parameters'' is a commonly reported metric in PEFT literature, but it can refer to different things: the number of \textbf{trainable parameters}, \textbf{changed parameters}, or the \textbf{rank of the update} (e.g., \citet{intrinsic_said}). In some cases, the number of trainable vs. changed parameters can differ by orders of magnitude. Examples include DiffPruning and LT-SFT \cite{diff_pruning,lottery_ticket_tuning}, which first fine-tune the full network and then prune the update afterward, or Prefix Tuning \cite{prefix_tuning}, which uses an FCN reparametrization of the prompt parameters.

We provide a detailed comparison of the trainable and updated parameters for 28 PEFT methods in Table \ref{tab:scale_table} and discuss this issue further in Section \ref{sec:best_practices}. Generally, sparse methods tend to have more trainable than changed parameters, while reparametrization methods often have fewer trainable parameters due to the nature of reparametrization.

In our study, we consider \textbf{five key dimensions} essential to benchmark the effectiveness of parameter-efficient fine-tuning methods. These dimensions include storage efficiency, memory efficiency, computational efficiency, inference overhead, and downstream performance metrics (e.g. accuracy). Our analysis of the published literature shows that while these dimensions are interconnected, improvements along one of the axes do not necessarily translate into improvements along the others. For example, optimizing for parameter efficiency alone does not guarantee reduced RAM usage. Table \ref{tab:efficiency} summarizes our findings.

Despite their significance, PEFT performance metrics such as memory efficiency, training speed, and inference overhead (e.g., throughput) are only occasionally quantified in the papers. However, presenting these metrics only helps to further analyze a particular PEFT method of interest, in isolation. Head-to-head comparison across different PEFT methods is still challenging, primarily because of the impact of experimental setup on performance metrics. Thus, to address this gap, we fix the experimental set-up and perform a \textbf{large-scale experimental comparison of PEFT methods} as a part of this survey. We discuss details of the experimental setup and our results in the following sections.


\subsection{Experimental Comparison: Setup}
\label{sec:peft_comparison_setup}

Our experimental comparison is designed to provide a comprehensive evaluation of PEFT methods, exceeding the scope and depth of existing studies such as \citet{delta_tuning}. We have carefully selected 14 PEFT methods representing diverse categories within our taxonomy – including Additive, Selective, Reparametrization-based, and Hybrid methods – to ensure a broad and inclusive analysis. Notably, we exclude sparse selective methods, acknowledging their limited practicality on modern hardware and their primary focus on storage efficiency. Furthermore, our study omits BitFit in the context of T5 networks, which do not utilize biases. 
In addition to these PEFT methods, we include a full fine-tuning baseline to provide a reference point for both downstream task performance and efficiency metrics. Unlike \citet{delta_tuning}, which limited their experimental comparison to four PEFT methods and focused primarily on downstream performance and memory consumption, our experimental design covers a wider range of methods and evaluates them in all the following: memory efficiency, training speed, inference speed, and downstream performance.

\paragraph{Datasets}

We use both natural language understanding (NLU) and natural language generation (NLG) tasks for the comparison of the methods.

While the GLUE benchmark \cite{wang2018glue} is commonly used to evaluate parameter-efficient fine-tuning methods in the existing literature, many models now match or exceed human performance on GLUE tasks, some even with no or minimal training\footnote{\href{https://gluebenchmark.com/leaderboard}{gluebenchmark.com/leaderboard}}. This makes GLUE less effective at evaluating fine-tuning procedure performance. More recently, proposed alternatives to GLUE include MMLU \cite{hendrycks2021measuring}, HELM \cite{liang2023holistic}, and BigBench \cite{srivastava2022imitation}. MMLU emphasizes zero- or few-shot evaluations, making it unsuitable for assessing fine-tuning. Both HELM and BigBench present computational challenges due to their task diversity, especially when comparing a broad range of methods and models of up to 11B parameters.

In contrast, SuperGLUE tasks remain both demanding (with only a few models surpassing the human baseline) and computationally manageable. Specifically, we selected BoolQ, RTE, and COPA for this study. BoolQ is a yes/no question-answering dataset mainly evaluating models' world knowledge. COPA focuses on commonsense causal reasoning, for example, ``Premise: The man broke his toe. What was the CAUSE of this? Alternative 1: He got a hole in his sock. Alternative 2: He dropped a hammer on his foot.'' RTE is a natural language inference dataset where, given a premise, the model needs to predict if the hypothesis can be inferred from it, contradicts it, or is not relevant.

For more diverse comparisons, we also include one natural language generation dataset: CNN-Dailymail \cite{cnn_dailymail}, a large (300K training examples) summarization dataset. From the surveyed literature, we found that summarization usually highlights differences between PEFT methods and full fine-tuning, making this dataset particularly useful.

\paragraph{Models}

To compare parameter-efficient fine-tuning methods, we apply them to three sizes of T5: Large (0.7B), 3B, and 11B \cite{t5}. The range from 0.7B to 11B models not only tests each method's effectiveness at different scales but also presents common challenges associated with large-scale model training. A key aspect of this comparison is to demonstrate how PEFT methods can address practical issues such as memory constraints. For instance, the 11B model allows us to compare PEFT methods' performance and efficiency in one of the most relevant practical cases when full fine-tuning does not fit even into 80GB of GPU memory.

\paragraph{PEFT Methods}

We use the following fine-tuning methods in our comparison:

\begin{itemize}
\setlength\itemsep{0em}
    \item Full tuning -- regular fine-tuning of all model parameters
    \item Houlsby -- Adapters inserted after the attention and FCN layers of the Transformer as described in Section \ref{sec:additive_section_adapters} \cite{adapters}
    \item Pfeiffer -- Adapters inserted only after the FCN layers \cite{modular_deep_learning}
    \item Parallel Adapter -- scaled parallel adapter as described in \cite{parallel_adapter}
    \item IA3 -- (IA)$^3$ learns re-scaling vectors for keys, values, and hidden FFN activations (Section \ref{sec:ia3}) \cite{}
    \item Prefix Tuning -- learns a prefix added to keys and values and uses FCN reparametrization for these parameters (Section \ref{sec:prefix_tuning}) \cite{prefix_tuning}
    \item Prompt Tuning -- learns a prefix added to keys directly (Section \ref{sec:prompt_tuning}) \cite{prompt_tuning}
    \item LN tuning -- fine-tune only layer-norm parameters \cite{layer_norm_tuning}
    \item LoRA (q and v) -- LoRA applied to the query and value networks only (Section \ref{sec:lora}) \cite{lora}
    \item LoRA (all linear) -- LoRA applied to all linear layers in the model \cite{lora}
    \item KronA -- LoRA-like Kronecker product-based reparametrization of the weight matrices (Section \ref{sec:krona}) \cite{krona}
    \item MAM -- Mix-and-Match Adapters (Section \ref{sec:mam_adapter}) \cite{}
    \item Compacter -- Kronecker product-based reparametrization of Adapter layers as described in Section \ref{sec:compacter} \cite{compacter}
    \item Compacter++ -- Compacter layers that are only applied after the FCN in the Transformer, similar to the idea of Pfeiffer vs Houlsby Adapters \cite{}
    \item UniPELT -- a hybrid method that combines LoRA, Prefix Tuning, and Adapters through gating (Section \ref{sec:unipelt}) \cite{unipelt}
\end{itemize}

\paragraph{Metrics}

In our evaluation, we focus on assessing PEFT method efficiency in terms of memory consumption, training speed, and inference speed, and then compare models on downstream metrics.

To quantify memory efficiency, we track the maximum RAM consumption during training using \texttt{torch.cuda.max\_memory\_allocated()}. Training speed is quantified by the number of input tokens processed per second during training and for inference -- during evaluation. We do not explicitly merge reparametrization-based methods into model weights during evaluation to present results for a typical use-case when methods like LoRA are used in the adapter fashion. When merged, there should be no difference between reparametrization-based methods and regular training in terms of the inference speed. We use accuracy for SuperGLUE datasets and ROUGE-L for summarization.

\paragraph{Implementation details and hyperparameters}

All models are fine-tuned in text-to-text fashion following \citet{t5}. We use Adapters and PEFT libraries \cite{adapters_library_v2,peft_libarry} for most of the methods and implement several methods in our repository\footnote{\href{http://github.com/guitaricet/peft_comparison}{github.com/guitaricet/peft\_comparison}} from scratch. When using existing implementations, we utilize default architecture hyperparameters for the method from the corresponding library, which are usually close to the hyperparameters reported in the method's original paper.

For all NLU datasets, we perform a learning rate sweep over values \{1e-3, 1e-4, and 1e-5\} and select the best, which we then train on two more random seeds to estimate the standard deviation. Our preliminary experiments indicate a negligible impact of weight decay in our setup (<0.01). Due to computational constraints, CNN/Dailymail experiments only use one seed and a learning rate of 1e-3, which we found to perform consistently well across PEFT methods. We estimate the standard deviation of CNN/Dailymail runs via several random seeds on randomly selected PEFT methods and find it to be lower than 0.5 ROUGE-L points, which we use for the standard deviation estimate for the rest of the methods.

We use a maximum input sequence length of 512 tokens in all our experiments. In NLU experiments, the maximum output sequence length is 8, and for summarization, it is 128 tokens.

Each NLU model undergoes training for either 3 epochs or a minimum of 100 update steps. For CNN/Dailymail, the training duration is set to one epoch (9 thousand update steps). We use a batch size of 32 in all our experiments, utilizing gradient accumulation to achieve this batch size when needed. While all SuperGLUE tasks converge by the end of training in most of our experiments, we observe that CNN/Dailymail continues to improve throughout the training and does not plateau. Our setup thereby favors methods exhibiting faster learning, which is especially relevant for low-resource scenarios commonly faced in PEFT applications.

In total, we train three models of size 0.7-11B parameters with 14 PEFT methods, five runs for each of the three NLU datasets, and one for the summarization dataset. Together with the full fine-tuning baseline, this brings the total experiment count to around 700. We report raw (non-aggregated) results in Appendix \ref{sec:peft_comparison_raw}.

\begin{table}[t]
    \centering
    \begin{tabular}{l|l|c}
    \toprule
        Model & Datasets & Microbatch size  \\
    \midrule
        T5$_{LARGE}$ & RTE, COPA & 32 \\
        T5$_{LARGE}$ & BoolQ & 16 \\
        T5$_{LARGE}$ & CNN & 4 \\
        T5$_{3B}$    & RTE, COPA & 4 \\
        T5$_{3B}$ & BoolQ & 2 \\
        T5$_{3B}$ & CNN & 1 \\
        T5$_{11B}$   & RTE, COPA, BoolQ, CNN & 1 \\
    \bottomrule
    \end{tabular}
    \caption{Microbatch sizes used in our experiments}
    \label{tab:peft_comparison_microbatch}
\end{table}

\paragraph{Hardware setup}
We estimate throughput using a single A100 40GB GPU for most of the experiments, with several exceptions due to out-of-memory issues. UniPELT, MAM, and Prefix Tuning for T5-11B were trained with a single A100 80GB GPU, which should give comparable throughput numbers to the A100 40GB. Full fine-tuning T5-11B experiments were performed with two A100 80GB GPUs using model-parallel distributed training. RAM estimates for this model training are the total memory consumption of both GPUs, which should give an estimate comparable to the rest of the experiments, as optimizer states are not shared between GPUs in model-parallel training.

Table \ref{tab:peft_comparison_microbatch} specifies the number of examples processed simultaneously (microbatch size) in our experiments. Microbatch sizes are kept consistent across different tuning methods to enable fair comparison of memory efficiency. This also allows us to isolate the throughput improvements from the PEFT method itself, rather than increased batch size. Methods with lower memory consumption could further benefit from increased batch sizes.

\subsection{Comparison Results: Downstream Performance}
\label{sec:peft_comparison_downstream}

Table \ref{tab:peft_comparison_dataset_average_short} shows downstream metrics averaged over the datasets. Scores are averaged, and standard deviations are aggregated using the Euclidean mean of per-dataset variances. This table compares the downstream performance of PEFT methods across model scales. Non-aggregated results for all our experiments are available in Appendix \ref{sec:peft_comparison_raw}.


We note a few key observations:

\paragraph{Houlsby Adapters and LoRA consistently perform the best}

Houlsby Adapters and LoRA are the only methods that consistently achieve full-tuning performance with little to no effort in hyperparameter tuning.

\paragraph{Hybrid methods are especially sensitive to hyperparameters}

MAM Adapters and UniPELT were consistently hard to train. While the results in Table \ref{tab:peft_comparison_dataset_average_short} include only the best model from our sweep over three learning rates, additional experiments to improve MAM and UniPELT only marginally improved their performance. We attribute this to the generally poor performance of Prompt Tuning when trained in a compute-limited scenario.

\begin{table}[t]
\centering
\begin{small}
\begin{tabular}{l|ccc}
\toprule
\textbf{Method} & T5$_{LARGE}$ & T5$_{3B}$ & T5$_{11B}$ \\
\midrule
\textbf{Additive methods}&&&\\
Adapters (Houlsby) &
$\textbf{67.34}_{\pm9.58}$ &
$\uline{74.66}_{\pm1.68}$ &
$\textbf{76.16}_{\pm1.47}$ \\
Adapters (Pfeiffer)& $62.93_{\pm3.52}$ & $\uline{69.92}_{\pm5.61}$ & $50.72_{\pm1.69}$ \\
Parallel Adapter   & $\uline{66.78}_{\pm3.85}$ & $\uline{74.15}_{\pm0.88}$ & $\uline{68.74}_{\pm12.73}$ \\
IA3                & $55.06_{\pm1.80}$ & $41.77_{\pm0.50}$ & $61.05_{\pm3.42}$ \\
Prefix Tuning      & $45.05_{\pm3.89}$ & $48.90_{\pm5.37}$ & $51.93_{\pm2.21}$ \\
Prompt Tuning      & $8.97_{\pm30.91}$ & $8.38_{\pm0.50}$ & - \\

\textbf{Selective methods}&&&\\
LN Tuning         & $\uline{64.68}_{\pm4.59}$ & $72.95_{\pm1.38}$ & $\textbf{73.77}_{\pm0.93}$ \\

\textbf{Reparametrization-based methods}&&&\\
LoRA (q and v)   &$\textbf{67.42}_{\pm2.32}$& $\textbf{75.49}_{\pm1.71}$ & $\textbf{76.20}_{\pm1.27}$ \\
LoRA (all linear)&$\textbf{68.76}_{\pm1.83}$& $\textbf{75.22}_{\pm1.28}$ & $\textbf{76.58}_{\pm2.16}$ \\
KronA             & $\uline{65.68}_{\pm3.27}$ & $71.98_{\pm0.57}$ & $\uline{72.13}_{\pm7.30}$ \\

\textbf{Hybrid methods}&&&\\
MAM               & $46.90_{\pm6.47}$ & $45.57_{\pm4.67}$ & $51.49_{\pm0.54}$ \\
Compacter         & $64.48_{\pm1.81}$ & $70.72_{\pm0.87}$ & $\textbf{74.33}_{\pm1.40}$ \\
Compacter++       & $64.78_{\pm2.23}$ & $71.00_{\pm1.62}$ & $\textbf{74.72}_{\pm0.82}$ \\
Unipelt           & $44.10_{\pm15.48}$ & $47.16_{\pm4.84}$ & $52.29_{\pm3.09}$ \\

\midrule
Full tuning       & $67.22$ & $74.83$ & $73.25$ \\
\bottomrule
\end{tabular}

\vspace{2pt}
\end{small}
\caption{Average model performance on our collection of datasets (Section \ref{sec:peft_comparison_setup}) with 95\% confidence intervals (two standard deviations). We \textbf{bold} values that outperform full-tuning by mean value. We \uline{underline} values that achieve full-tuning performance within the confidence interval.}
\label{tab:peft_comparison_dataset_average_short}
\end{table}

\paragraph{Prefix Tuning and Prompt Tuning significantly differ in performance}

Prefix Tuning \cite{prefix_tuning} and Prompt Tuning \cite{prompt_tuning} are two different PEFT methods that are easy to confuse in terms of naming and concept. Both methods use the idea of continuous prompt optimization, but Prefix Tuning reparametrizes the trainable prefix via a fully-connected network (Section~\ref{sec:prefix_tuning}). In contrast, Prompt Tuning directly optimizes the prefix, albeit at the cost of slower convergence and typically much larger prefix length.
We observe significant differences between these methods in our experiments. Both of them suffer from slow convergence, which substantially hurts performance in our setup. However, Prompt Tuning never outperformed the constant prediction baseline.\footnote{Since we use a text-to-text approach, results worse than the constant prediction are expected. A badly trained network will never generate a correct class name.} Additionally, Prompt Tuning was extremely sensitive to the random seed (especially for T5-large and 3B models), as observed by its high standard deviation from the mean.


\paragraph{Multiple methods underperform their reported values}

Multiple methods that had claimed to outperform Adapters or LoRA (virtually all other methods) do not perform well in our setup. This includes most of the methods with the exception of Parallel Adapter, Compacter, and KronA, which perform on par with the best methods in several cases, especially for 11B models.

\paragraph{Pfeiffer Adapters Perform Significantly Worse Than Houlsby}

\citet{adapter_fusion} observes that inserting adapters only after the FCN in the Transformer achieves similar performance as inserting adapters after both FCN and Attention (MHA) layers. However, in our experiments we find a significant and consistent difference of up to 15 points that increases with model scale. This highlights the importance of evaluating methods for both small and large models.

\paragraph{Layer Norm Tuning is unexpectedly competitive}

Layer Norm parameters are rarely used for parameter-efficient fine-tuning; we found only a few mentions of the method \cite{layer_norm_tuning,t_few}. However, it can be implemented in one line of code and shows performance competitive to full fine-tuning for T5-Large and T5-11B. We want to highlight this result and recommend using LN tuning as a baseline for future PEFT work.

\subsection{Comparison Results: Efficiency}
\label{sec:peft_comparison_efficiency}
{
\newcommand{\cb}{\cellcolor{bluegrad}}
\newcommand{\rb}{\cellcolor{redgrad}}
\newcommand{\gr}[1]{\textcolor{gray!90}{#1}}
\newcommand{\st}{$^*$}
\renewcommand{\arraystretch}{0.6} 
\begin{table}
\centering
\begin{tabular}{@{}ll|c|r|r|r|r|r@{}}
\toprule

\multirow{2}{*}{Model} & \multirow{2}{*}{PEFT Method} & \multirow{2}{*}{Score} & \multicolumn{2}{c|}{Trainable params} & RAM & \multicolumn{2}{c}{Throughput} \\
& & & \multicolumn{1}{c|}{Millions} & \multicolumn{1}{c|}{\%} & \multicolumn{1}{c|}{\footnotesize{(GB)}} & \multicolumn{1}{c|}{Train.} & \multicolumn{1}{c}{Infer.}\\

\midrule
T5$_{Large}$ & Full tuning    & $67.22\hspace{1.8em}$ & 737.67 & 
\rb100 &
\rb20.7 &
100   &
\cb100 \\
T5$_{Large}$ & Houlsby    & $\textbf{67.34}_{\pm9.58}$& 12.69  & 1.720  &\cb12.1 & 101.7 & 70.1 \\
T5$_{Large}$ & Pfeiffer           & $62.93_{\pm3.52}$ & 6.34   & 0.860  & 14.0 & 108.6 & 76.8 \\
T5$_{Large}$ & Parallel Adapter   & $66.78_{\pm3.85}$ & 50.41  & 6.833  & 14.7 & 123.1 & 73.2 \\
T5$_{Large}$ & IA3                & $55.06_{\pm1.80}$ & 0.34   & 0.047  & 16.2 & \cb124.7 & \gr{76.2} \\
T5$_{Large}$ & Prefix tuning      & $45.05_{\pm3.89}$ & 77.31  & 10.481 & 15.4 & 113.2 & 65.0 \\
T5$_{Large}$ & Prompt tuning      & $8.97_{\pm30.91}$ & 4.42   & 0.600 &\cb10.6& 108.8 & 64.5 \\
T5$_{Large}$ & LN tuning          & $64.68_{\pm4.59}$ & 0.12   & \cb0.017&14.6 & \cb143.2 &\cb103.1 \\
T5$_{Large}$&LoRa (q and v)&$\textbf{67.42}_{\pm2.32}$& 2.36   & 0.320  & 14.3 & \cb123.7 & \cb\gr{87.4} \\
T5$_{Large}$&LoRA (all)    &$\textbf{68.76}_{\pm1.83}$& 8.65   & 1.173  & 15.9 & \rb80.9  & \gr{66.9} \\
T5$_{Large}$ & KronA              & $65.68_{\pm3.27}$ & 0.22   &\cb0.030& 14.6 & 109.3 & \gr{82.3} \\
T5$_{Large}$ & Compacter          & $64.48_{\pm1.81}$ & 0.30   & 0.041  & 14.4 & \rb81.3  & \rb59.0 \\
T5$_{Large}$ & Compacter++        & $64.78_{\pm2.23}$ & 0.15   & \cb0.021 & \cb13.9 & 95.4 & 68.1 \\
T5$_{Large}$ & MAM                & $46.90_{\pm6.47}$ & 171.07 & \rb23.191& \rb16.6 & 106.1  & \rb59.1 \\
T5$_{Large}$ & Unipelt            & $44.10_{\pm15.48}$& 86.22  & \rb11.689& \rb16.7 & \rb61.1  & \rb45.2 \\
\addlinespace
T5$_{3B}$    & Full tuning     & $74.83\hspace{1.8em}$&2851.60& \rb100  &\rb32.9& 100   &\cb100.0 \\
T5$_{3B}$    & Houlsby            & $74.66_{\pm1.68}$ & 12.69 & 0.445   &\cb9.3 & 93.1  & 58.2 \\
T5$_{3B}$    & Pfeiffer           & $69.92_{\pm5.61}$ & 6.34  & 0.223   &\cb9.6 & 102.9 & 64.5 \\
T5$_{3B}$    & Parallel Adapter   & $74.15_{\pm0.88}$ & 50.41 & 1.768   & 10.0  & 102.7  & 61.2 \\
T5$_{3B}$    & IA3                & $41.77_{\pm0.50}$ & 1.38  & 0.048   & 10.8  &\cb103.9  &\gr{65.6} \\
T5$_{3B}$    & Prefix tuning      & $48.90_{\pm5.37}$ & 77.25 & 2.709   & 12.3  &\cb105.2  & 56.8 \\
T5$_{3B}$    & Prompt tuning      & $8.38_{\pm0.50}$  & 17.69 & 0.621   &\cb8.9 & 100.0 & 58.8 \\
T5$_{3B}$    & LN tuning          & $72.95_{\pm1.38}$ & 0.12  &\cb0.004 & 9.7   &\cb140.7&\cb101.8 \\
T5$_{3B}$   &LoRa (q and v)&$\textbf{75.49}_{\pm1.71}$& 3.93  & 0.138   &  9.6  & 103.0  &\cb\gr{79.8} \\
T5$_{3B}$   &LoRA (all)    &$\textbf{75.22}_{\pm1.28}$& 25.17 & 0.883   & 10.6  &\rb70.2   & \gr{56.7} \\
T5$_{3B}$    & KronA              & $71.98_{\pm0.57}$ & 0.41  & 0.014   & 10.8  & 81.2   & \gr{74.0} \\
T5$_{3B}$    & Compacter          & $70.72_{\pm0.87}$ & 0.30  &\cb0.011 &  9.7  &\rb71.5 &\rb46.0 \\
T5$_{3B}$    & Compacter++        & $71.00_{\pm1.62}$ & 0.15  &\cb0.005 &  9.5  & 86.9 & 54.3 \\
T5$_{3B}$    & MAM                & $45.57_{\pm4.67}$ & 171.07&\rb5.999 &\rb14.2& 84.5  & \rb52.1 \\
T5$_{3B}$    & Unipelt            & $47.16_{\pm4.84}$ & 316.70&\rb11.106&\rb12.8&\rb51.0  & \rb38.2 \\
\addlinespace
T5$_{11B}$   & Full tuning    & $73.25\hspace{1.8em}$&11307.32 & \rb100&\rb104.9&\rb100.0 & \cb100  \\
T5$_{11B}$   & Houlsby   & $\textbf{76.16}_{\pm1.47}$  & 12.69 & 0.112 & 28.8  & 214.9 & 70.3 \\
T5$_{11B}$   & Pfeiffer           & $50.72_{\pm1.69}$  &  6.34 & 0.056&\cb28.6 & \cb239.5 &\cb 76.3 \\
T5$_{11B}$   & Parallel Adapter   & $68.74_{\pm12.73}$ & 50.41 & 0.446 & 29.0  & \cb233.3 & 73.9 \\
T5$_{11B}$   & IA3                & $61.05_{\pm3.42}$  &  3.48 & 0.031 & 31.0  & 232.2 & \gr{75.5} \\
T5$_{11B}$   & Prefix tuning      & $51.93_{\pm2.21}$ &1211.99 & 10.719& 37.9  & 182.3 & 62.3 \\
T5$_{11B}$   & Prompt tuning      & -                  & 70.78 & 0.626 & 30.2  & 209.5 & 64.0 \\
T5$_{11B}$   & LN tuning     &$\textbf{73.77}_{\pm0.93}$&0.12 &\cb0.001&\cb28.4&\cb275.9 &\cb97.8 \\
T5$_{11B}$   & LoRa (q and v)&$\textbf{76.20}_{\pm1.27}$&20.05 & 0.177 & 28.9 & 216.9 & \gr{81.3} \\
T5$_{11B}$   & LoRA (all)    &$\textbf{76.58}_{\pm2.16}$&91.23 & 0.807 & 30.8 & 161.7 & \gr{62.4} \\
T5$_{11B}$   & KronA              & $72.13_{\pm7.30}$  & 0.77  & 0.007 & 33.2 & 174.0 & \gr{71.8} \\
T5$_{11B}$   & Compacter  & $\textbf{74.33}_{\pm1.40}$ & 0.30  & \cb0.003 & \cb28.6 & 162.7 &\rb56.8 \\
T5$_{11B}$   & Compacter++& $\textbf{74.72}_{\pm0.82}$ & 0.15  & \cb0.001 & \cb28.5 & 208.8 & 64.2 \\
T5$_{11B}$   & MAM                & $51.49_{\pm0.54}$  &1262.39&\rb11.164 & \rb43.3 & \rb152.8 & \rb57.3 \\
T5$_{11B}$   & Unipelt            & $52.29_{\pm3.09}$  &1224.78&\rb10.832 & \rb38.9 & \rb100.2 & \rb46.8 \\
\bottomrule
\end{tabular}
\caption{
Average model performance on our collection of datasets and efficiency metrics: number of trainable parameters, GPU memory consumption, training and inference throughput. We use \textbf{bold} scores to indicate that method outperforms full fine-tuning. \gr{Gray} values for inference throughput indicate that PEFT weights can be merged into the network (made 100\%) if needed. Throughput 95\%-confidence intervals are: $\pm6$ (T5$_{LARGE}$), $\pm7$ (T5$_{3B}$), and $\pm3$ (T5$_{11B}$).
}
\label{tab:peft_comparison_dataset_average_tall}
\end{table}
}

Table \ref{tab:peft_comparison_dataset_average_tall} presents a detailed comparison of efficiency and performance for the 14 PEFT methods compared in our study. We show the actual number of \textbf{trainable parameters} (as opposed to changed parameters), the maximum GPU memory consumption during training, and the throughput in ktok/s (thousands of tokens per second) both during training and inference.

\paragraph{All PEFT methods reduce memory consumption}

As expected, all methods from our study significantly reduce memory consumption. The smallest improvement we see is \textbf{4GB} in the UniPELT and T5-Large combination, which is quite considerable because it is 10\% of the GPU RAM. The biggest improvement is \textbf{71.5GB} in the Compacter++ and T5-11B combination. This allows fine-tuning T5-11B on a single 40GB GPU instead of two 80GB GPUs and dramatically improves training speed by a factor of more than two.

\paragraph{Smaller models (<1B) can train slower with PEFT}

Any PEFT (Parameter-Efficient Fine-Tuning) method that adds parameters to the network involves additional forward (and potentially backward) pass overhead. For sufficiently large models or when only a few parameters are added, this overhead can be negligible. However, if the method adds too many parameters, it can lead to \textbf{slower training compared to regular fine-tuning}. We observe this in T5-Large models, which are small only compared to billion-scale models, as they have 738M parameters\footnote{T5-Large is significantly bigger than BERT and RoBERTa models}. For instance, applying LoRA to all T5-Large parameters results in a 20\% training slowdown. Similar slowdowns are noted for MAM adapters, Compacter, and UniPELT, with 20\%, 5\%, and 40\% slower training, respectively, compared to full fine-tuning. Despite these slowdowns, they all offer memory improvements.

\paragraph{PEFT significantly affects inference speed}

In all PEFT methods that add trainable parameters to the network, we observe a significant slowdown in inference speed. The slowdown ranges from 33-55\% for T5-Large, 20-60\% for T5-3B, and 20-55\% for T5-11B (absolute points). Within the set of additive methods, we observe that Pfeiffer adapters and (IA)$^3$ offer the best inference speeds. 
It is important to note that in our throughput estimation for reparametrization-based methods, we did not merge the method parameters into the network. If merged, they would have the same inference speed as regular fine-tuning, as no additional parameters are present. However, methods like LoRA are increasingly used in modular approaches, such as referenced in \cite{lorahub}, without merging LoRA parameters. The results from Table \ref{tab:peft_comparison_dataset_average_tall} are relevant for these scenarios.

\paragraph{Kronecker-Based Reparametrizations Do Not Improve Memory Efficiency, But Improve Speed}

Across different model scales, we observe that extremely parameter-efficient methods like Compacter and KronA, which employ Kronecker products to enhance parameter efficiency, do not significantly reduce memory usage. Despite training with two orders of magnitude fewer parameters than LoRA, the memory consumption of Compacter and KronA is nearly identical to that of LoRA. For instance, LoRA optimizes 20 million parameters for T5-11B, while KronA and Compacter each optimize less than 0.5 million. Nevertheless, all methods consume approximately 28.6GB of GPU memory. This result becomes intuitive in hindsight: beyond a certain point, the memory used for optimizer states and gradients becomes negligible, overshadowed by other factors such as model weights and hidden states. Nevertheless, we observe significant training and inference speed improvements with KronA over LoRA. This likely occurs due to the efficient Kronecker-vector product implementation in KronA (Section \ref{sec:krona}).


\textbf{In conclusion}, our experimental comparison shows several expected results, such as significant improvements in memory consumption and speed. However, we also observed some surprising results. Notably, we observed that methods like Layer Norm Tuning, which are often overlooked, can be unexpectedly effective. Additionally, the effects of various PEFT methods on inference speed, especially in larger models, highlight the complex trade-offs between efficiency and performance. These insights emphasize the need for a comprehensive evaluation of PEFT methods, taking into account not only memory and speed but also their scalability across different model sizes.

Based on our experiments, we note a few key take-away points for the practitioners. 

\begin{tcolorbox}[
    colback=gray!05,  
    colframe=black!50, 
    boxrule=0.5pt,     
    arc=3mm,           
    width=\linewidth,  
    leftrule=0.1mm,      
    rightrule=0.1mm,     
    bottomrule=0.1mm,    
    toprule=0.1mm,       
]
\textbf{Key Findings: }
\begin{itemize}
    \item Houlsby Adapters \cite{adapters} and LoRA \cite{lora} perform at par or better than full-tuning, with little to no effort in hyperparameter tuning. 
    \item Layer Norm \cite{layer_norm_tuning} tuning provides a highly competitive and efficient method that is easy to implement. 
\end{itemize}

\end{tcolorbox}

\section{Challenges and guidelines}

Survey papers tend to discuss reporting issues, and this one is no exception. We identified several challenges and inconsistencies that make it difficult to evaluate PEFT methods and draw direct comparisons between different PEFT methods, which warrant discussion.

\paragraph{Reporting parameter count} 
One of the primary challenges stems from the difference in the way researchers report parameter counts. These inconsistencies arise from the inherent complexity of the problem. Parameter counts can be categorized into three types: the number of \textbf{trainable parameters}, the number of \textbf{changed parameters} between the original and fine-tuned models, and the \textbf{rank} of the difference between the original and fine-tuned models. These parameter counts are not equivalent. For example, IntrinsicSAID (Section \ref{sec:intrinsic_said}) learns a low-rank ($\sim$100-1000) transformation of model parameters. However, it changes all ($100\%$) of the model's parameters. DiffPruning (Section \ref{sec:diff_pruning}) learns an update of $0.5\%$ of the parameters, but it actually trains $200\%$ of the parameters: fine-tuning the model and learning the binary mask. For reparameterization-based methods (Sections \ref{sec:lora}, \ref{sec:krona}, \ref{sec:compacter}), memory requirements may vary depending on the implementation design choices. Of the three types, the number of trainable parameters is the most reliable predictor of memory efficiency. However, it is still imperfect: Ladder-side Tuning trains more parameters than LoRA or BitFit, but it uses less RAM by avoiding backpropagation to the main network. 

\begin{tcolorbox}[
    colback=gray!05,  
    colframe=black!50, 
    boxrule=0.5pt,     
    arc=3mm,           
    width=\linewidth,  
    leftrule=0.1mm,      
    rightrule=0.1mm,     
    bottomrule=0.1mm,    
    toprule=0.1mm,       
]
\textbf{Guideline - 1}: Explicit reporting of the number of parameters and the type i.e. trainable parameters or changed parameters or rank of the changes to the model being tuned.
\end{tcolorbox}

\paragraph{Reporting efficiency}
Evaluating the efficiency of PEFT methods solely based on parameter count is challenging due to the non-linear relationship between parameter count and efficiency. 
Efficiency in training time is better assessed through memory consumption and training speed. Most PEFT categories, except for Sparse-selective methods, significantly improve RAM usage. However, the Intrinsic SAID \cite{intrinsic_said} method, which is Reparametrization-based, can result in higher memory usage than full training due to the Fastfood transformation's demands.
Our experiments revealed that modularity in hybrid PEFT (e.g., MAM adapters \cite{}, UniPELT \cite{unipelt}) methods comes at the cost of notably higher memory consumption. This emphasizes the need for studies to report memory consumption to help practitioners make informed decisions. We also noticed considerable variability in training speed even with similar RAM usage, suggesting that RAM consumption should be considered alongside training speed.
After training, the storage space required for the changed parameters is crucial for evaluating PEFT methods. Unlike full fine-tuning, which alters all model parameters, PEFT only requires saving a subset, significantly improving storage efficiency. However, methods like IPT require saving different parameter sets at various training stages, making clear reporting of space requirements essential.
Inference latency is another critical factor in practice. Additive methods typically introduce overhead because they require computations on both the original network and the added parameters, whereas Selective methods do not, as they operate on existing model weights. Moreover, additive and reparametrization-based methods (e.g., LoRA, KronA) offer advantages in multi-task inference by reducing memory usage from $O(NM)$ to $O(M + NA)$, where $A$ is the number of added weights per task. Some additive methods, like LST, can also enhance inference speed by using the original network solely as a feature extractor. For further details on multi-task training and inference, we refer readers to Modular Deep Learning \cite{modular_deep_learning}.

\begin{tcolorbox}[
    colback=gray!05,  
    colframe=black!50, 
    boxrule=0.5pt,     
    arc=3mm,           
    width=\linewidth,  
    leftrule=0.1mm,      
    rightrule=0.1mm,     
    bottomrule=0.1mm,    
    toprule=0.1mm,       
]
\textbf{Guideline - 2}: Explicit reporting of memory (RAM) consumption during training, token throughput during training and inference, and storage requirements are necessary to evaluate the efficiency of the PEFT method. Furthermore, if the proposed methods consist of multiple stages, all metrics should be reported for each stage. 
\end{tcolorbox}

\paragraph{Model sizes}
Another challenge arises from the variation in model sizes used in the evaluation of PEFT methods. It is important to assess methods fine-tuning different model sizes, especially \textgreater1B and \textless20B parameters. With the increase in the backbone model size, the need and usefulness of PEFT methods increase rapidly. Several studies \cite{intrinsic_said,lora} have demonstrated that larger models require fewer parameters to be updated during fine-tuning, both in terms of percentage and when the model is large enough, sometimes even in absolute terms \cite{prefix_tuning}. We would like to particularly stress this, considering that even recent papers often focus solely on BERT. Furthermore, in our experiments, Layer Norm tuning \cite{layer_norm_tuning} was the only consistently efficient method at different scales, while maintaining a competitive performance now downstream tasks. For all other methods, efficiency, and performance considerably varies at different model sizes. Thus, model size must be considered when reporting PEFT methods.

\begin{tcolorbox}[
    colback=gray!05,  
    colframe=black!50, 
    boxrule=0.5pt,     
    arc=3mm,           
    width=\linewidth,  
    leftrule=0.1mm,      
    rightrule=0.1mm,     
    bottomrule=0.1mm,    
    toprule=0.1mm,       
]
\textbf{Guideline - 3}: Efficiency metrics and downstream performance should be reported across multiple model scales. Authors should also consider reporting fine-tuning results for the model sizes that are most commonly used by the research community at the time of experimentation.
\end{tcolorbox}

\paragraph{Method Implementation} Another issue encountered is the state of published implementations. Many codebases are simply copies of the Transformers library \cite{Wolf_Transformers_State-of-the-Art_Natural_2020} or other repositories with only minor modifications. These copies often do not use git forks, making it difficult to identify the differences unless they are highlighted in the README file. But even when differences are easy to find, the code is frequently not readable or reusable. Users are often required to install a modified version of the Transformers library, which conflicts with the most recent version and lacks documentation or examples of how to reuse the method outside of the existing codebase. Despite these challenges, there are some methods with reusable implementations worth highlighting, such as LoRA\footnote{\href{https://github.com/microsoft/LoRA}{github.com/microsoft/LoRA}} and Compacter\footnote{\href{https://github.com/rabeehk/compacter}{github.com/rabeehk/compacter}}. These implementations stand out for their user-friendliness and adaptability, providing a solid foundation for further research and development.

\begin{tcolorbox}[
    colback=gray!05,  
    colframe=black!50, 
    boxrule=0.5pt,     
    arc=3mm,           
    width=\linewidth,  
    leftrule=0.1mm,      
    rightrule=0.1mm,     
    bottomrule=0.1mm,    
    toprule=0.1mm,       
]
\textbf{Guideline - 4}: The study presenting a PEFT method should be accompanied by an easy-to-use implementation of the method.
\end{tcolorbox}

\paragraph{Comparison} 
Intuitively, the presented PEFT method should be compared against popular approaches (e.g., LoRA, BitFit, Adapters) and the methods that share conceptual and architectural similarities with the presented method. However, the absence of standard benchmarks and metrics complicates the comparison of PEFT methods. New methods are often evaluated on different model/dataset combinations, making it challenging to draw meaningful conclusions.
We would like to highlight the papers that report a variety of metrics on standard datasets, simplifying comparison to other methods. For example, KronA \cite{krona} evaluated T5-base on the GLUE benchmark and reported accuracy, training time, and inference time while maintaining the same number of trainable parameters. UniPELT \cite{unipelt} assessed BERT on the GLUE benchmark and reported accuracy, training time, and inference latency, although it used different parameter counts for various methods. LST \cite{ladder_side_tuning} evaluated different T5 sizes on the GLUE benchmark, reporting metrics such as accuracy, training time, the number of updated parameters, and memory usage. MAM \cite{parallel_adapter} applied multiple models to the XSUM benchmark and reported accuracy across a range of trainable parameters, although memory comparisons were not provided.
However, even these papers lack full comparability due to differences in their evaluation settings, such as varying parameter counts or the absence of certain metrics like memory comparisons. These inconsistencies highlight the need for a standardized benchmark and unified metrics to facilitate more accurate comparisons and evaluations of PEFT methods.
Based on our survey and experiments we identified the principal qualities of each of the categories and summarized them in this section and Table~\ref{tab:type_comparison}.

\begin{tcolorbox}[
    colback=gray!05,  
    colframe=black!50, 
    boxrule=0.5pt,     
    arc=3mm,           
    width=\linewidth,  
    leftrule=0.1mm,      
    rightrule=0.1mm,     
    bottomrule=0.1mm,    
    toprule=0.1mm,       
]
Call for community efforts in developing standardized benchmarks and competition for evaluation of PEFT methods.
\end{tcolorbox}

\newcommand{\hug}{\includegraphics[height=1.3em]{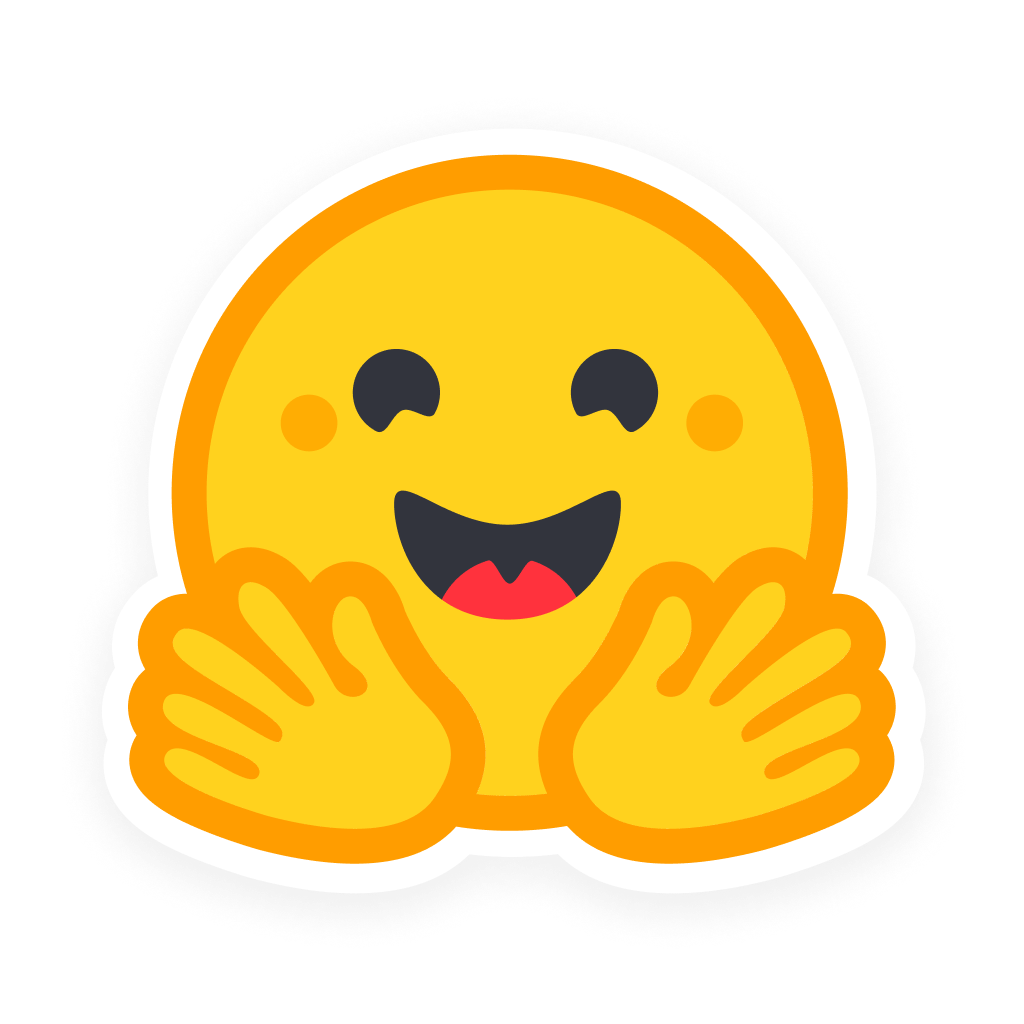}}
\newcommand{\adapterhub}{\includegraphics[height=1.3em]{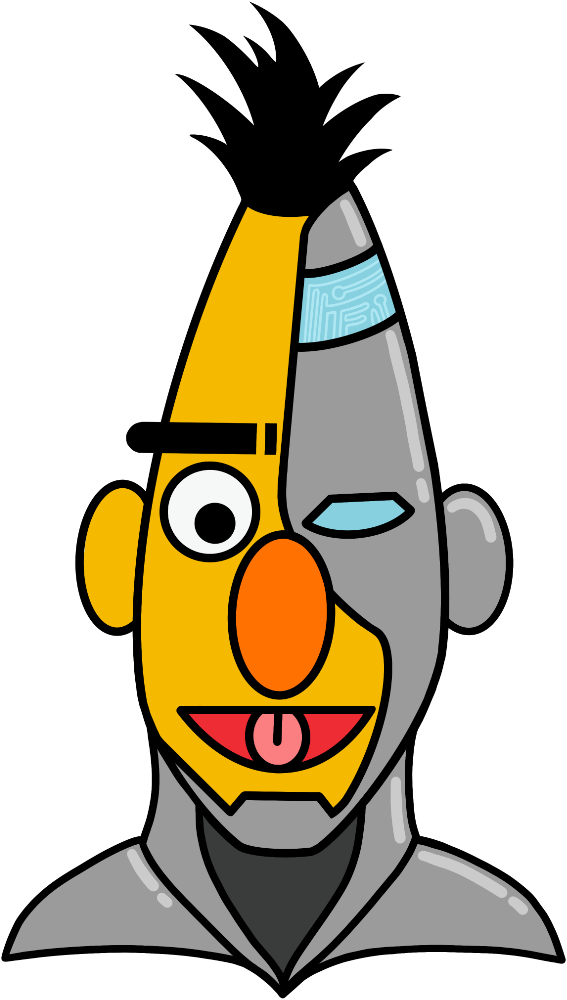}}

\begin{table}[t]
\begin{small}
    \centering
    \setlength{\tabcolsep}{4pt} 
    \begin{tabular}{lc|ccc|cc|c|c}
        \toprule
        & & \multicolumn{3}{c|}{Additive} & \multicolumn{2}{c|}{Selective} & \multirow{2}{*}{Reparam.} & \multirow{2}{*}{Hybrid} \\
        & & Adapters & Soft prompts & Other & Structured & Sparse & & \\
        \midrule
\multicolumn{2}{l|}{Storage efficiency} & \yes & \yes & \yes & \yes & \yes & \yes & \yes \\
\midrule
\multirow{2}{*}{Training}& RAM
& \yes & \yes & \yes & \yes & \no & \yes & \orange{maybe} \\
& Speed &
\no & \no & \orange{maybe} & \orange{maybe} & \no & \yes & \orange{maybe} \\
\midrule
\multirow{2}{*}{Inference}&\footnotesize{Single-task}
&\multicolumn{3}{c|}{\no\ Adds overhead} &
\multicolumn{2}{c|}{\yes\ No overhead} & \yes    & \orange{maybe} \\
&\footnotesize{Mutli-task}&
\multicolumn{3}{c|}{\yes Allows efficient inference} & \orange{maybe} & \no & \orange{maybe} & \orange{maybe} \\
\midrule
\multicolumn{2}{l|}{Implemented in} & \adapterhub & \hug \adapterhub & \hug \adapterhub & \no & \no & \hug \adapterhub & \adapterhub\\
        \bottomrule
    \end{tabular}
    \caption{
Comparison of PEFT categories in terms of the axes outlined in Section \ref{sec:peft_comparison}. For training, we evaluate methods by their memory (RAM) efficiency and speed relative to full training. \yes\ means more effective than regular training, \no\ means less effective, and \orange{maybe} suggests it might vary based on the specific method within the category. For inference, the table shows whether a method introduces single-task overhead and if it facilitates efficient multi-task inference. Lastly, the table highlights if these methods are implemented in popular open-source libraries: HuggingFace PEFT (\hug) and AdapterHub (\adapterhub).
    }
    \label{tab:type_comparison}
\end{small}
\end{table}

\section{Discussion}
\label{sec:discussion}
The growing accessibility of large language models \cite{zhang2022opt,zeng2022glm130b,yalm,touvron2023llama} and the democratization of their inference through low-bit quantization \cite{dettmers2022llm,Dettmers2022TheCF} have enabled the research community to study, experiment, and tackle new tasks with relatively modest compute budgets. Parameter-efficient fine-tuning is the next step that allows us not just to infer, but to modify these models.

Some methods, including Adapters, Prompt Tuning, LoRA, and (IA)$^3$, have shown their practicality at scale (Table~\ref{tab:scale_table}). However, in practice, matching the performance of full fine-tuning remains a challenge. One of the reasons is high \textbf{sensitivity to hyperparameters}, with optimal hyperparameters often significantly deviating from those used in full fine-tuning due to the varying number of trainable parameters. For instance, the optimal learning rate for parameter-efficient fine-tuning is generally much higher than that for full fine-tuning. The research community should promote in-depth investigations into the impact of hyperparameters on these methods and find reasonable defaults, as parameter-efficient fine-tuning of large models can be noticeably costly at the 20-100B scale. Additionally, efforts should be directed towards developing methods that minimize hyperparameter sensitivity, such as pre-training new parameters \cite{spot,prompt_mapping}.

Examining the taxonomy of methods and the progress made thus far, it is evident that low-rank reparameterization has been remarkably successful in enhancing parameter efficiency. LoRA-style (Section \ref{sec:lora}) and Kronecker-product (Sections \ref{sec:compacter} and \ref{sec:krona}) reparameterizations both decrease the number of trainable parameters while requiring minimal extra computation. A possible future direction for finding new PEFT models is exploring different \textbf{reparametrization techniques} with favorable trainable parameter count vs. rank ratio.

Another possible direction for improvement is utilizing what we know about \textbf{how transformer models process texts} \cite{rogers-etal-2020-primer}. Most of the PEFT methods work uniformly for the model, while we know that models process input differently at different layers. Utilizing this knowledge or building systems that have an adaptive number of parameters per layer could further improve parameter efficiency and accuracy.

In many respects, fine-tuning large language models faces the same challenges as those encountered in \textbf{edge machine learning} -- we consistently face constraints on memory, computation, and even energy consumption.
Techniques like quantization and pruning \cite{pmlr-v37-gupta15,optimal_brain_damage} that are widely used in edge machine learning now benefit large language models. As we move forward, it is not only plausible but also likely that more ideas could be exchanged between these two areas. Cross-disciplinary collaboration could facilitate the exchange of ideas, accelerating innovation and progress in parameter-efficient fine-tuning.


\acks{This work was funded in part by an Amazon Alexa AI research award to Anna Rumshisky. We would like to thank Vladimir Kluenkov and Victoria Maltseva for their help with Figure~2.}


\newpage


\vskip 0.2in
\bibliography{bibliography}

\begin{thebibliography}{99}
\providecommand{\natexlab}[1]{#1}
\providecommand{\url}[1]{\texttt{#1}}
\expandafter\ifx\csname urlstyle\endcsname\relax
  \providecommand{\doi}[1]{doi: #1}\else
  \providecommand{\doi}{doi: \begingroup \urlstyle{rm}\Url}\fi

\bibitem[Aghajanyan et~al.(2020)Aghajanyan, Zettlemoyer, and Gupta]{intrinsic_said}
Armen Aghajanyan, Luke Zettlemoyer, and Sonal Gupta.
\newblock Intrinsic dimensionality explains the effectiveness of language model fine-tuning.
\newblock In \emph{Annual Meeting of the Association for Computational Linguistics}, 2020.

\bibitem[AkbarTajari et~al.(2022)AkbarTajari, Rajaee, and Pilehvar]{layer_norm_tuning}
Mohammad AkbarTajari, Sara Rajaee, and Mohammad~Taher Pilehvar.
\newblock An empirical study on the transferability of transformer modules in parameter-efficient fine-tuning.
\newblock In Yoav Goldberg, Zornitsa Kozareva, and Yue Zhang, editors, \emph{Proceedings of the 2022 Conference on Empirical Methods in Natural Language Processing}, pages 10617--10625, Abu Dhabi, United Arab Emirates, December 2022. Association for Computational Linguistics.
\newblock \doi{10.18653/v1/2022.emnlp-main.726}.
\newblock URL \url{https://aclanthology.org/2022.emnlp-main.726}.

\bibitem[Ansell et~al.(2022)Ansell, Ponti, Korhonen, and Vuli{\'c}]{lottery_ticket_tuning}
Alan Ansell, Edoardo Ponti, Anna Korhonen, and Ivan Vuli{\'c}.
\newblock Composable sparse fine-tuning for cross-lingual transfer.
\newblock In \emph{Proceedings of the 60th Annual Meeting of the Association for Computational Linguistics (Volume 1: Long Papers)}, pages 1778--1796, Dublin, Ireland, May 2022. Association for Computational Linguistics.
\newblock \doi{10.18653/v1/2022.acl-long.125}.
\newblock URL \url{https://aclanthology.org/2022.acl-long.125}.

\bibitem[Arora et~al.(2018)Arora, Ge, Neyshabur, and Zhang]{Arora2018StrongerGB}
Sanjeev Arora, Rong Ge, Behnam Neyshabur, and Yi~Zhang.
\newblock Stronger generalization bounds for deep nets via a compression approach.
\newblock In \emph{International Conference on Machine Learning}, 2018.

\bibitem[Ba et~al.(2016)Ba, Kiros, and Hinton]{layer_norm}
Jimmy Ba, Jamie~Ryan Kiros, and Geoffrey~E. Hinton.
\newblock Layer normalization.
\newblock \emph{ArXiv}, abs/1607.06450, 2016.

\bibitem[Bahdanau et~al.(2014)Bahdanau, Cho, and Bengio]{bahdanau2014neural}
Dzmitry Bahdanau, Kyunghyun Cho, and Yoshua Bengio.
\newblock Neural machine translation by jointly learning to align and translate.
\newblock \emph{arXiv preprint arXiv:1409.0473}, 2014.

\bibitem[Ben-Zaken et~al.(2021)Ben-Zaken, Ravfogel, and Goldberg]{bitfit}
Elad Ben-Zaken, Shauli Ravfogel, and Yoav Goldberg.
\newblock Bitfit: Simple parameter-efficient fine-tuning for transformer-based masked language-models.
\newblock \emph{ArXiv}, abs/2106.10199, 2021.

\bibitem[Bertsch et~al.(2024)Bertsch, Ivgi, Alon, Berant, Gormley, and Neubig]{bertsch2024context}
Amanda Bertsch, Maor Ivgi, Uri Alon, Jonathan Berant, Matthew~R Gormley, and Graham Neubig.
\newblock In-context learning with long-context models: An in-depth exploration.
\newblock \emph{arXiv preprint arXiv:2405.00200}, 2024.

\bibitem[Brown et~al.(2020)Brown, Mann, Ryder, Subbiah, Kaplan, Dhariwal, Neelakantan, Shyam, Sastry, Askell, Agarwal, Herbert-Voss, Krueger, Henighan, Child, Ramesh, Ziegler, Wu, Winter, Hesse, Chen, Sigler, Litwin, Gray, Chess, Clark, Berner, McCandlish, Radford, Sutskever, and Amodei]{brown2020language_gpt3}
Tom Brown, Benjamin Mann, Nick Ryder, Melanie Subbiah, Jared~D Kaplan, Prafulla Dhariwal, Arvind Neelakantan, Pranav Shyam, Girish Sastry, Amanda Askell, Sandhini Agarwal, Ariel Herbert-Voss, Gretchen Krueger, Tom Henighan, Rewon Child, Aditya Ramesh, Daniel Ziegler, Jeffrey Wu, Clemens Winter, Chris Hesse, Mark Chen, Eric Sigler, Mateusz Litwin, Scott Gray, Benjamin Chess, Jack Clark, Christopher Berner, Sam McCandlish, Alec Radford, Ilya Sutskever, and Dario Amodei.
\newblock Language models are few-shot learners.
\newblock In H.~Larochelle, M.~Ranzato, R.~Hadsell, M.~F. Balcan, and H.~Lin, editors, \emph{Advances in Neural Information Processing Systems}, volume~33, pages 1877--1901. Curran Associates, Inc., 2020.
\newblock URL \url{https://proceedings.neurips.cc/paper/2020/file/1457c0d6bfcb4967418bfb8ac142f64a-Paper.pdf}.

\bibitem[Cao et~al.(2022)Cao, Prakash, and Hamza]{attention_fusion}
Jin Cao, Chandan Prakash, and Wael Hamza.
\newblock Attention fusion: a light yet efficient late fusion mechanism for task adaptation in nlu.
\newblock In \emph{NAACL-HLT}, 2022.

\bibitem[Chavan et~al.(2023)Chavan, Liu, Gupta, Xing, and Shen]{chavan2023one}
Arnav Chavan, Zhuang Liu, Deepak Gupta, Eric Xing, and Zhiqiang Shen.
\newblock One-for-all: Generalized lora for parameter-efficient fine-tuning.
\newblock \emph{arXiv preprint arXiv:2306.07967}, 2023.

\bibitem[Chen et~al.(2023)Chen, Zhang, Shi, Li, Smola, and Yang]{design_spaces}
Jiaao Chen, Aston Zhang, Xingjian Shi, Mu~Li, Alexander~J. Smola, and Diyi Yang.
\newblock Parameter-efficient fine-tuning design spaces.
\newblock \emph{ArXiv}, abs/2301.01821, 2023.

\bibitem[Chiang et~al.(2021)Chiang, Rush, and Barak]{named_tensor}
David Chiang, Alexander~M. Rush, and Boaz Barak.
\newblock Named tensor notation.
\newblock \emph{ArXiv}, abs/2102.13196, 2021.

\bibitem[Chowdhery et~al.(2022)Chowdhery, Narang, Devlin, Bosma, Mishra, Roberts, Barham, Chung, Sutton, Gehrmann, Schuh, Shi, Tsvyashchenko, Maynez, Rao, Barnes, Tay, Shazeer, Prabhakaran, Reif, Du, Hutchinson, Pope, Bradbury, Austin, Isard, Gur-Ari, Yin, Duke, Levskaya, Ghemawat, Dev, Michalewski, Garc{\'i}a, Misra, Robinson, Fedus, Zhou, Ippolito, Luan, Lim, Zoph, Spiridonov, Sepassi, Dohan, Agrawal, Omernick, Dai, Pillai, Pellat, Lewkowycz, Moreira, Child, Polozov, Lee, Zhou, Wang, Saeta, D{\'i}az, Firat, Catasta, Wei, Meier-Hellstern, Eck, Dean, Petrov, and Fiedel]{palm}
Aakanksha Chowdhery, Sharan Narang, Jacob Devlin, Maarten Bosma, Gaurav Mishra, Adam Roberts, Paul Barham, Hyung~Won Chung, Charles Sutton, Sebastian Gehrmann, Parker Schuh, Kensen Shi, Sasha Tsvyashchenko, Joshua Maynez, Abhishek Rao, Parker Barnes, Yi~Tay, Noam~M. Shazeer, Vinodkumar Prabhakaran, Emily Reif, Nan Du, Benton~C. Hutchinson, Reiner Pope, James Bradbury, Jacob Austin, Michael Isard, Guy Gur-Ari, Pengcheng Yin, Toju Duke, Anselm Levskaya, Sanjay Ghemawat, Sunipa Dev, Henryk Michalewski, Xavier Garc{\'i}a, Vedant Misra, Kevin Robinson, Liam Fedus, Denny Zhou, Daphne Ippolito, David Luan, Hyeontaek Lim, Barret Zoph, Alexander Spiridonov, Ryan Sepassi, David Dohan, Shivani Agrawal, Mark Omernick, Andrew~M. Dai, Thanumalayan~Sankaranarayana Pillai, Marie Pellat, Aitor Lewkowycz, Erica Moreira, Rewon Child, Oleksandr Polozov, Katherine Lee, Zongwei Zhou, Xuezhi Wang, Brennan Saeta, Mark D{\'i}az, Orhan Firat, Michele Catasta, Jason Wei, Kathleen~S. Meier-Hellstern, Douglas Eck, Jeff Dean, Slav Petrov,
  and Noah Fiedel.
\newblock Palm: Scaling language modeling with pathways.
\newblock \emph{ArXiv}, abs/2204.02311, 2022.

\bibitem[Dettmers and Zettlemoyer(2022)]{Dettmers2022TheCF}
Tim Dettmers and Luke Zettlemoyer.
\newblock The case for 4-bit precision: k-bit inference scaling laws.
\newblock \emph{ArXiv}, abs/2212.09720, 2022.

\bibitem[Dettmers et~al.(2022)Dettmers, Lewis, Belkada, and Zettlemoyer]{dettmers2022llm}
Tim Dettmers, Mike Lewis, Younes Belkada, and Luke Zettlemoyer.
\newblock Llm. int8 (): 8-bit matrix multiplication for transformers at scale.
\newblock \emph{arXiv preprint arXiv:2208.07339}, 2022.

\bibitem[Dettmers et~al.(2023)Dettmers, Pagnoni, Holtzman, and Zettlemoyer]{dettmers2023qlora}
Tim Dettmers, Artidoro Pagnoni, Ari Holtzman, and Luke Zettlemoyer.
\newblock Qlora: Efficient finetuning of quantized llms, 2023.

\bibitem[Devlin et~al.(2019)Devlin, Chang, Lee, and Toutanova]{devlin2018bert}
Jacob Devlin, Ming-Wei Chang, Kenton Lee, and Kristina Toutanova.
\newblock {BERT}: Pre-training of deep bidirectional transformers for language understanding.
\newblock In \emph{Proceedings of the 2019 Conference of the North {A}merican Chapter of the Association for Computational Linguistics: Human Language Technologies, Volume 1 (Long and Short Papers)}, pages 4171--4186, Minneapolis, Minnesota, June 2019. Association for Computational Linguistics.
\newblock \doi{10.18653/v1/N19-1423}.
\newblock URL \url{https://aclanthology.org/N19-1423}.

\bibitem[Ding et~al.(2022)Ding, Qin, Yang, Wei, Yang, Su, Hu, Chen, Chan, Chen, Yi, Zhao, Wang, Liu, Zheng, Chen, Liu, Tang, Li, and Sun]{delta_tuning}
Ning Ding, Yujia Qin, Guang Yang, Fu~Wei, Zonghan Yang, Yusheng Su, Shengding Hu, Yulin Chen, Chi-Min Chan, Weize Chen, Jing Yi, Weilin Zhao, Xiaozhi Wang, Zhiyuan Liu, Haitao Zheng, Jianfei Chen, Yang Liu, Jie Tang, Juan Li, and Maosong Sun.
\newblock Delta tuning: A comprehensive study of parameter efficient methods for pre-trained language models.
\newblock \emph{ArXiv}, abs/2203.06904, 2022.

\bibitem[Donahue et~al.(2014)Donahue, Jia, Vinyals, Hoffman, Zhang, Tzeng, and Darrell]{last_layer_tuning}
Jeff Donahue, Yangqing Jia, Oriol Vinyals, Judy Hoffman, Ning Zhang, Eric Tzeng, and Trevor Darrell.
\newblock Decaf: A deep convolutional activation feature for generic visual recognition.
\newblock In Eric~P. Xing and Tony Jebara, editors, \emph{Proceedings of the 31st International Conference on Machine Learning}, volume~32 of \emph{Proceedings of Machine Learning Research}, pages 647--655, Bejing, China, 22--24 Jun 2014. PMLR.
\newblock URL \url{https://proceedings.mlr.press/v32/donahue14.html}.

\bibitem[Edalati et~al.(2022)Edalati, Tahaei, Kobyzev, Nia, Clark, and Rezagholizadeh]{krona}
Ali Edalati, Marzieh~S. Tahaei, Ivan Kobyzev, V.~Nia, James~J. Clark, and Mehdi Rezagholizadeh.
\newblock Krona: Parameter efficient tuning with kronecker adapter.
\newblock \emph{ArXiv}, abs/2212.10650, 2022.

\bibitem[Fedus et~al.(2021)Fedus, Zoph, and Shazeer]{switch}
William Fedus, Barret Zoph, and Noam~M. Shazeer.
\newblock Switch transformers: Scaling to trillion parameter models with simple and efficient sparsity.
\newblock \emph{ArXiv}, abs/2101.03961, 2021.

\bibitem[Fu et~al.(2021)Fu, Huang, Chen, Tian, and Zhao]{lets}
Cheng Fu, Hanxian Huang, Xinyun Chen, Yuandong Tian, and Jishen Zhao.
\newblock Learn-to-share: A hardware-friendly transfer learning framework exploiting computation and parameter sharing.
\newblock In Marina Meila and Tong Zhang, editors, \emph{Proceedings of the 38th International Conference on Machine Learning}, volume 139 of \emph{Proceedings of Machine Learning Research}, pages 3469--3479. PMLR, 18--24 Jul 2021.
\newblock URL \url{https://proceedings.mlr.press/v139/fu21a.html}.

\bibitem[Gheini et~al.(2021)Gheini, Ren, and May]{cross_attention_tuning}
Mozhdeh Gheini, Xiang Ren, and Jonathan May.
\newblock Cross-attention is all you need: Adapting pretrained transformers for machine translation.
\newblock In \emph{Conference on Empirical Methods in Natural Language Processing}, 2021.

\bibitem[Guo et~al.(2020)Guo, Rush, and Kim]{diff_pruning}
Demi Guo, Alexander~M. Rush, and Yoon Kim.
\newblock Parameter-efficient transfer learning with diff pruning.
\newblock In \emph{Annual Meeting of the Association for Computational Linguistics}, 2020.

\bibitem[Guo et~al.(2023)Guo, Greengard, Xing, and Kim]{guo2023lq}
Han Guo, Philip Greengard, Eric~P Xing, and Yoon Kim.
\newblock Lq-lora: Low-rank plus quantized matrix decomposition for efficient language model finetuning.
\newblock \emph{arXiv preprint arXiv:2311.12023}, 2023.

\bibitem[Gupta et~al.(2015)Gupta, Agrawal, Gopalakrishnan, and Narayanan]{pmlr-v37-gupta15}
Suyog Gupta, Ankur Agrawal, Kailash Gopalakrishnan, and Pritish Narayanan.
\newblock Deep learning with limited numerical precision.
\newblock In Francis Bach and David Blei, editors, \emph{Proceedings of the 32nd International Conference on Machine Learning}, volume~37 of \emph{Proceedings of Machine Learning Research}, pages 1737--1746, Lille, France, 07--09 Jul 2015. PMLR.
\newblock URL \url{https://proceedings.mlr.press/v37/gupta15.html}.

\bibitem[Hambardzumyan et~al.(2021)Hambardzumyan, Khachatrian, and May]{Hambardzumyan2021WARPWA}
Karen Hambardzumyan, Hrant Khachatrian, and Jonathan May.
\newblock Warp: Word-level adversarial reprogramming.
\newblock In \emph{Annual Meeting of the Association for Computational Linguistics}, 2021.

\bibitem[He et~al.(2022{\natexlab{a}})He, Zhou, Ma, Berg-Kirkpatrick, and Neubig]{parallel_adapter}
Junxian He, Chunting Zhou, Xuezhe Ma, Taylor Berg-Kirkpatrick, and Graham Neubig.
\newblock Towards a unified view of parameter-efficient transfer learning.
\newblock In \emph{International Conference on Learning Representations}, 2022{\natexlab{a}}.
\newblock URL \url{https://openreview.net/forum?id=0RDcd5Axok}.

\bibitem[He et~al.(2016)He, Zhang, Ren, and Sun]{resnet}
Kaiming He, Xiangyu Zhang, Shaoqing Ren, and Jian Sun.
\newblock Deep residual learning for image recognition.
\newblock \emph{2016 IEEE Conference on Computer Vision and Pattern Recognition (CVPR)}, Jun 2016.
\newblock \doi{10.1109/cvpr.2016.90}.
\newblock URL \url{http://dx.doi.org/10.1109/cvpr.2016.90}.

\bibitem[He et~al.(2022{\natexlab{b}})He, Ding, Dong, Zhang, and Tao]{sparse_adapter}
Shwai He, Liang Ding, Daize Dong, Jeremy Zhang, and Dacheng Tao.
\newblock {S}parse{A}dapter: An easy approach for improving the parameter-efficiency of adapters.
\newblock In \emph{Findings of the Association for Computational Linguistics: EMNLP 2022}, pages 2184--2190, Abu Dhabi, United Arab Emirates, December 2022{\natexlab{b}}. Association for Computational Linguistics.
\newblock URL \url{https://aclanthology.org/2022.findings-emnlp.160}.

\bibitem[Hendrycks et~al.(2021)Hendrycks, Burns, Basart, Zou, Mazeika, Song, and Steinhardt]{hendrycks2021measuring}
Dan Hendrycks, Collin Burns, Steven Basart, Andy Zou, Mantas Mazeika, Dawn Song, and Jacob Steinhardt.
\newblock Measuring massive multitask language understanding.
\newblock In \emph{International Conference on Learning Representations}, 2021.
\newblock URL \url{https://openreview.net/forum?id=d7KBjmI3GmQ}.

\bibitem[Holmes et~al.(2021)Holmes, Zhang, He, and Wu]{nxm_transformer}
Connor Holmes, Minjia Zhang, Yuxiong He, and Bo~Wu.
\newblock Nxmtransformer: Semi-structured sparsification for natural language understanding via admm.
\newblock In M.~Ranzato, A.~Beygelzimer, Y.~Dauphin, P.S. Liang, and J.~Wortman Vaughan, editors, \emph{Advances in Neural Information Processing Systems}, volume~34, pages 1818--1830. Curran Associates, Inc., 2021.
\newblock URL \url{https://proceedings.neurips.cc/paper/2021/file/0e4f5cc9f4f3f7f1651a6b9f9214e5b1-Paper.pdf}.

\bibitem[Houlsby et~al.(2019)Houlsby, Giurgiu, Jastrzebski, Morrone, de~Laroussilhe, Gesmundo, Attariyan, and Gelly]{adapters}
Neil Houlsby, Andrei Giurgiu, Stanislaw Jastrzebski, Bruna Morrone, Quentin de~Laroussilhe, Andrea Gesmundo, Mona Attariyan, and Sylvain Gelly.
\newblock Parameter-efficient transfer learning for nlp.
\newblock In \emph{International Conference on Machine Learning}, 2019.

\bibitem[Hu et~al.(2022)Hu, yelong shen, Wallis, Allen-Zhu, Li, Wang, Wang, and Chen]{lora}
Edward~J Hu, yelong shen, Phillip Wallis, Zeyuan Allen-Zhu, Yuanzhi Li, Shean Wang, Lu~Wang, and Weizhu Chen.
\newblock Lo{RA}: Low-rank adaptation of large language models.
\newblock In \emph{International Conference on Learning Representations}, 2022.
\newblock URL \url{https://openreview.net/forum?id=nZeVKeeFYf9}.

\bibitem[Huang et~al.(2018)Huang, Vaswani, Uszkoreit, Shazeer, Simon, Hawthorne, Dai, Hoffman, Dinculescu, and Eck]{huang2018music}
Cheng-Zhi~Anna Huang, Ashish Vaswani, Jakob Uszkoreit, Noam Shazeer, Ian Simon, Curtis Hawthorne, Andrew~M Dai, Matthew~D Hoffman, Monica Dinculescu, and Douglas Eck.
\newblock Music transformer.
\newblock \emph{arXiv preprint arXiv:1809.04281}, 2018.

\bibitem[Huang et~al.(2023)Huang, Liu, Lin, Pang, Du, and Lin]{lorahub}
Chengsong Huang, Qian Liu, Bill~Yuchen Lin, Tianyu Pang, Chao Du, and Min Lin.
\newblock Lorahub: Efficient cross-task generalization via dynamic lora composition, 2023.

\bibitem[Jeon et~al.(2024)Jeon, Kim, and Kim]{jeon2024l4q}
Hyesung Jeon, Yulhwa Kim, and Jae-joon Kim.
\newblock L4q: Parameter efficient quantization-aware training on large language models via lora-wise lsq.
\newblock \emph{arXiv preprint arXiv:2402.04902}, 2024.

\bibitem[Karimi~Mahabadi et~al.(2021)Karimi~Mahabadi, Henderson, and Ruder]{compacter}
Rabeeh Karimi~Mahabadi, James Henderson, and Sebastian Ruder.
\newblock Compacter: Efficient low-rank hypercomplex adapter layers.
\newblock In M.~Ranzato, A.~Beygelzimer, Y.~Dauphin, P.S. Liang, and J.~Wortman Vaughan, editors, \emph{Advances in Neural Information Processing Systems}, volume~34, pages 1022--1035. Curran Associates, Inc., 2021.
\newblock URL \url{https://proceedings.neurips.cc/paper/2021/file/081be9fdff07f3bc808f935906ef70c0-Paper.pdf}.

\bibitem[Keles et~al.(2023)Keles, Wijewardena, and Hegde]{keles2023computational}
Feyza~Duman Keles, Pruthuvi~Mahesakya Wijewardena, and Chinmay Hegde.
\newblock On the computational complexity of self-attention.
\newblock In \emph{International Conference on Algorithmic Learning Theory}, pages 597--619. PMLR, 2023.

\bibitem[Khrushchev et~al.(2022)Khrushchev, Vasilev, Petrov, and Zinov]{yalm}
Mikhail Khrushchev, Ruslan Vasilev, Alexey Petrov, and Nikolay Zinov.
\newblock {YaLM 100B}, 6 2022.
\newblock URL \url{https://github.com/yandex/YaLM-100B}.

\bibitem[Kingma and Ba(2015)]{adam}
Diederik~P. Kingma and Jimmy Ba.
\newblock Adam: {A} method for stochastic optimization.
\newblock In Yoshua Bengio and Yann LeCun, editors, \emph{3rd International Conference on Learning Representations, {ICLR} 2015, San Diego, CA, USA, May 7-9, 2015, Conference Track Proceedings}, 2015.
\newblock URL \url{http://arxiv.org/abs/1412.6980}.

\bibitem[Le et~al.(2013)Le, Sarl{\'o}s, and Smola]{Le2013FastfoodAK}
Quoc~V. Le, Tam{\'a}s Sarl{\'o}s, and Alex Smola.
\newblock Fastfood: Approximate kernel expansions in loglinear time.
\newblock \emph{ArXiv}, abs/1408.3060, 2013.

\bibitem[LeCun et~al.(1989)LeCun, Denker, and Solla]{optimal_brain_damage}
Yann LeCun, John Denker, and Sara Solla.
\newblock Optimal brain damage.
\newblock In D.~Touretzky, editor, \emph{Advances in Neural Information Processing Systems}, volume~2. Morgan-Kaufmann, 1989.
\newblock URL \url{https://proceedings.neurips.cc/paper/1989/file/6c9882bbac1c7093bd25041881277658-Paper.pdf}.

\bibitem[Lester et~al.(2021)Lester, Al-Rfou, and Constant]{prompt_tuning}
Brian Lester, Rami Al-Rfou, and Noah Constant.
\newblock The power of scale for parameter-efficient prompt tuning.
\newblock \emph{ArXiv}, abs/2104.08691, 2021.

\bibitem[Lewis et~al.(2019)Lewis, Liu, Goyal, Ghazvininejad, Mohamed, Levy, Stoyanov, and Zettlemoyer]{lewis2019bart}
Mike Lewis, Yinhan Liu, Naman Goyal, Marjan Ghazvininejad, Abdelrahman Mohamed, Omer Levy, Ves Stoyanov, and Luke Zettlemoyer.
\newblock Bart: Denoising sequence-to-sequence pre-training for natural language generation, translation, and comprehension.
\newblock \emph{arXiv preprint arXiv:1910.13461}, 2019.

\bibitem[Li et~al.(2018)Li, Farkhoor, Liu, and Yosinski]{measuring_the_intrinsic_dimension}
Chunyuan Li, Heerad Farkhoor, Rosanne Liu, and Jason Yosinski.
\newblock Measuring the intrinsic dimension of objective landscapes.
\newblock In \emph{International Conference on Learning Representations}, 2018.

\bibitem[Li and Liang(2021)]{prefix_tuning}
Xiang~Lisa Li and Percy Liang.
\newblock Prefix-tuning: Optimizing continuous prompts for generation.
\newblock In \emph{Proceedings of the 59th Annual Meeting of the Association for Computational Linguistics and the 11th International Joint Conference on Natural Language Processing (Volume 1: Long Papers)}, pages 4582--4597, Online, August 2021. Association for Computational Linguistics.
\newblock \doi{10.18653/v1/2021.acl-long.353}.
\newblock URL \url{https://aclanthology.org/2021.acl-long.353}.

\bibitem[Li et~al.(2023)Li, Yu, Liang, He, Karampatziakis, Chen, and Zhao]{li2023loftq}
Yixiao Li, Yifan Yu, Chen Liang, Pengcheng He, Nikos Karampatziakis, Weizhu Chen, and Tuo Zhao.
\newblock Loftq: Lora-fine-tuning-aware quantization for large language models.
\newblock \emph{arXiv preprint arXiv:2310.08659}, 2023.

\bibitem[Lialin et~al.(2023)Lialin, Shivagunde, Muckatira, and Rumshisky]{relora}
Vladislav Lialin, Namrata Shivagunde, Sherin Muckatira, and Anna Rumshisky.
\newblock Stack more layers differently: High-rank training through low-rank updates.
\newblock \emph{arXiv preprint arXiv:2307.05695}, 2023.

\bibitem[Liang et~al.(2023)Liang, Bommasani, Lee, Tsipras, Soylu, Yasunaga, Zhang, Narayanan, Wu, Kumar, Newman, Yuan, Yan, Zhang, Cosgrove, Manning, Ré, Acosta-Navas, Hudson, Zelikman, Durmus, Ladhak, Rong, Ren, Yao, Wang, Santhanam, Orr, Zheng, Yuksekgonul, Suzgun, Kim, Guha, Chatterji, Khattab, Henderson, Huang, Chi, Xie, Santurkar, Ganguli, Hashimoto, Icard, Zhang, Chaudhary, Wang, Li, Mai, Zhang, and Koreeda]{liang2023holistic}
Percy Liang, Rishi Bommasani, Tony Lee, Dimitris Tsipras, Dilara Soylu, Michihiro Yasunaga, Yian Zhang, Deepak Narayanan, Yuhuai Wu, Ananya Kumar, Benjamin Newman, Binhang Yuan, Bobby Yan, Ce~Zhang, Christian Cosgrove, Christopher~D. Manning, Christopher Ré, Diana Acosta-Navas, Drew~A. Hudson, Eric Zelikman, Esin Durmus, Faisal Ladhak, Frieda Rong, Hongyu Ren, Huaxiu Yao, Jue Wang, Keshav Santhanam, Laurel Orr, Lucia Zheng, Mert Yuksekgonul, Mirac Suzgun, Nathan Kim, Neel Guha, Niladri Chatterji, Omar Khattab, Peter Henderson, Qian Huang, Ryan Chi, Sang~Michael Xie, Shibani Santurkar, Surya Ganguli, Tatsunori Hashimoto, Thomas Icard, Tianyi Zhang, Vishrav Chaudhary, William Wang, Xuechen Li, Yifan Mai, Yuhui Zhang, and Yuta Koreeda.
\newblock Holistic evaluation of language models, 2023.

\bibitem[Liu et~al.(2022)Liu, Tam, Muqeeth, Mohta, Huang, Bansal, and Raffel]{t_few}
Haokun Liu, Derek Tam, Mohammed Muqeeth, Jay Mohta, Tenghao Huang, Mohit Bansal, and Colin Raffel.
\newblock Few-shot parameter-efficient fine-tuning is better and cheaper than in-context learning.
\newblock \emph{ArXiv}, abs/2205.05638, 2022.

\bibitem[Liu et~al.(2024)Liu, Wang, Yin, Molchanov, Wang, Cheng, and Chen]{liu2024dora}
Shih-Yang Liu, Chien-Yi Wang, Hongxu Yin, Pavlo Molchanov, Yu-Chiang~Frank Wang, Kwang-Ting Cheng, and Min-Hung Chen.
\newblock Dora: Weight-decomposed low-rank adaptation.
\newblock \emph{arXiv preprint arXiv:2402.09353}, 2024.

\bibitem[Liu et~al.(2021)Liu, Zheng, Du, Ding, Qian, Yang, and Tang]{p_tuning}
Xiao Liu, Yanan Zheng, Zhengxiao Du, Ming Ding, Yujie Qian, Zhilin Yang, and Jie Tang.
\newblock Gpt understands, too.
\newblock \emph{ArXiv}, abs/2103.10385, 2021.

\bibitem[Lu et~al.(2023)Lu, Bigoulaeva, Sachdeva, Madabushi, and Gurevych]{lu2023emergent}
Sheng Lu, Irina Bigoulaeva, Rachneet Sachdeva, Harish~Tayyar Madabushi, and Iryna Gurevych.
\newblock Are emergent abilities in large language models just in-context learning?
\newblock \emph{arXiv preprint arXiv:2309.01809}, 2023.

\bibitem[Maddox et~al.(2020)Maddox, Benton, and Wilson]{parametercounting}
Wesley~J. Maddox, Gregory Benton, and Andrew~Gordon Wilson.
\newblock Rethinking parameter counting: Effective dimensionality revisted.
\newblock \emph{arXiv preprint arXiv:2003.02139}, 2020.

\bibitem[Malladi et~al.(2022)Malladi, Wettig, Yu, Chen, and Arora]{malladi2022kernel}
Sadhika Malladi, Alexander Wettig, Dingli Yu, Danqi Chen, and Sanjeev Arora.
\newblock A kernel-based view of language model fine-tuning.
\newblock \emph{arXiv preprint arXiv:2210.05643}, 2022.

\bibitem[Mangrulkar et~al.(2022)Mangrulkar, Gugger, Debut, Belkada, Paul, and Bossan]{peft_libarry}
Sourab Mangrulkar, Sylvain Gugger, Lysandre Debut, Younes Belkada, Sayak Paul, and Benjamin Bossan.
\newblock Peft: State-of-the-art parameter-efficient fine-tuning methods.
\newblock \url{https://github.com/huggingface/peft}, 2022.

\bibitem[Mao et~al.(2021)Mao, Mathias, Hou, Almahairi, Ma, Han, tau Yih, and Khabsa]{unipelt}
Yuning Mao, Lambert Mathias, Rui Hou, Amjad Almahairi, Hao Ma, Jiawei Han, Wen tau Yih, and Madian Khabsa.
\newblock Unipelt: A unified framework for parameter-efficient language model tuning.
\newblock \emph{ArXiv}, abs/2110.07577, 2021.

\bibitem[Pfeiffer et~al.(2020{\natexlab{a}})Pfeiffer, Kamath, R{\"u}ckl{\'e}, Cho, and Gurevych]{adapter_fusion}
Jonas Pfeiffer, Aishwarya Kamath, Andreas R{\"u}ckl{\'e}, Kyunghyun Cho, and Iryna Gurevych.
\newblock Adapterfusion: Non-destructive task composition for transfer learning.
\newblock \emph{ArXiv}, abs/2005.00247, 2020{\natexlab{a}}.

\bibitem[Pfeiffer et~al.(2020{\natexlab{b}})Pfeiffer, Rücklé, Poth, Kamath, Vulić, Ruder, Cho, and Gurevych]{adapterhub}
Jonas Pfeiffer, Andreas Rücklé, Clifton Poth, Aishwarya Kamath, Ivan Vulić, Sebastian Ruder, Kyunghyun Cho, and Iryna Gurevych.
\newblock Adapterhub: A framework for adapting transformers.
\newblock \emph{Proceedings of the 2020 Conference on Empirical Methods in Natural Language Processing: System Demonstrations}, 2020{\natexlab{b}}.
\newblock \doi{10.18653/v1/2020.emnlp-demos.7}.
\newblock URL \url{http://dx.doi.org/10.18653/v1/2020.emnlp-demos.7}.

\bibitem[Pfeiffer et~al.(2023)Pfeiffer, Ruder, Vulic, and Ponti]{modular_deep_learning}
Jonas Pfeiffer, Sebastian Ruder, Ivan Vulic, and E.~Ponti.
\newblock Modular deep learning.
\newblock \emph{ArXiv}, abs/2302.11529, 2023.

\bibitem[Poth et~al.(2023)Poth, Sterz, Paul, Purkayastha, Engl{\"a}nder, Imhof, Vuli, Ruder, Gurevych, and Pfeiffer]{adapters_library_v2}
Clifton Poth, Hannah Sterz, Indraneil Paul, Sukannya Purkayastha, Leon Engl{\"a}nder, Timo Imhof, Ivan Vuli, Sebastian Ruder, Iryna Gurevych, and Jonas Pfeiffer.
\newblock Adapters: A unified library for parameter-efficient and modular transfer learning.
\newblock In Yansong Feng and Els Lefever, editors, \emph{Proceedings of the 2023 Conference on Empirical Methods in Natural Language Processing: System Demonstrations}, pages 149--160, Singapore, December 2023. Association for Computational Linguistics.
\newblock \doi{10.18653/v1/2023.emnlp-demo.13}.
\newblock URL \url{https://aclanthology.org/2023.emnlp-demo.13}.

\bibitem[Qin et~al.(2024)Qin, Ma, Zheng, Li, Zhang, Liu, Luo, Liu, and Magno]{qin2024accurate}
Haotong Qin, Xudong Ma, Xingyu Zheng, Xiaoyang Li, Yang Zhang, Shouda Liu, Jie Luo, Xianglong Liu, and Michele Magno.
\newblock Accurate lora-finetuning quantization of llms via information retention.
\newblock \emph{arXiv preprint arXiv:2402.05445}, 2024.

\bibitem[Qin et~al.(2021)Qin, Wang, Su, Lin, Ding, Yi, Chen, Liu, Li, Hou, Li, Sun, and Zhou]{ipt}
Yujia Qin, Xiaozhi Wang, Yusheng Su, Yankai Lin, Ning Ding, Jing Yi, Weize Chen, Zhiyuan Liu, Juanzi Li, Lei Hou, Peng Li, Maosong Sun, and Jie Zhou.
\newblock Exploring universal intrinsic task subspace via prompt tuning.
\newblock 2021.

\bibitem[Radford et~al.(2019)Radford, Wu, Child, Luan, Amodei, and Sutskever]{radford2019language}
Alec Radford, Jeff Wu, Rewon Child, David Luan, Dario Amodei, and Ilya Sutskever.
\newblock Language models are unsupervised multitask learners.
\newblock 2019.

\bibitem[Raffel et~al.(2019)Raffel, Shazeer, Roberts, Lee, Narang, Matena, Zhou, Li, and Liu]{t5}
Colin Raffel, Noam Shazeer, Adam Roberts, Katherine Lee, Sharan Narang, Michael Matena, Yanqi Zhou, Wei Li, and Peter~J Liu.
\newblock Exploring the limits of transfer learning with a unified text-to-text transformer.
\newblock \emph{arXiv preprint arXiv:1910.10683}, 2019.

\bibitem[Rajpurkar et~al.(2018)Rajpurkar, Jia, and Liang]{squad2}
Pranav Rajpurkar, Robin Jia, and Percy Liang.
\newblock Know what you don’t know: Unanswerable questions for squad.
\newblock \emph{Proceedings of the 56th Annual Meeting of the Association for Computational Linguistics (Volume 2: Short Papers)}, 2018.
\newblock \doi{10.18653/v1/p18-2124}.
\newblock URL \url{http://dx.doi.org/10.18653/v1/P18-2124}.

\bibitem[Rebuffi et~al.(2017)Rebuffi, Bilen, and Vedaldi]{Rebuffi2017Adapters}
Sylvestre-Alvise Rebuffi, Hakan Bilen, and Andrea Vedaldi.
\newblock Learning multiple visual domains with residual adapters.
\newblock In I.~Guyon, U.~Von Luxburg, S.~Bengio, H.~Wallach, R.~Fergus, S.~Vishwanathan, and R.~Garnett, editors, \emph{Advances in Neural Information Processing Systems}, volume~30. Curran Associates, Inc., 2017.
\newblock URL \url{https://proceedings.neurips.cc/paper/2017/file/e7b24b112a44fdd9ee93bdf998c6ca0e-Paper.pdf}.

\bibitem[Rebuffi et~al.(2018)Rebuffi, Bilen, and Vedaldi]{Rebuffi2018EfficientPO}
Sylvestre-Alvise Rebuffi, Hakan Bilen, and Andrea Vedaldi.
\newblock Efficient parametrization of multi-domain deep neural networks.
\newblock \emph{2018 IEEE/CVF Conference on Computer Vision and Pattern Recognition}, pages 8119--8127, 2018.

\bibitem[Rogers et~al.(2020)Rogers, Kovaleva, and Rumshisky]{rogers-etal-2020-primer}
Anna Rogers, Olga Kovaleva, and Anna Rumshisky.
\newblock A primer in {BERT}ology: What we know about how {BERT} works.
\newblock \emph{Transactions of the Association for Computational Linguistics}, 8:\penalty0 842--866, 2020.
\newblock \doi{10.1162/tacl_a_00349}.
\newblock URL \url{https://aclanthology.org/2020.tacl-1.54}.

\bibitem[Salimans and Kingma(2016)]{weight_normalization}
Tim Salimans and Diederik~P. Kingma.
\newblock Weight normalization: a simple reparameterization to accelerate training of deep neural networks.
\newblock In \emph{Proceedings of the 30th International Conference on Neural Information Processing Systems}, NIPS'16, page 901–909, Red Hook, NY, USA, 2016. Curran Associates Inc.
\newblock ISBN 9781510838819.

\bibitem[Sanh et~al.(2022)Sanh, Webson, Raffel, Bach, Sutawika, Alyafeai, Chaffin, Stiegler, Raja, Dey, Bari, Xu, Thakker, Sharma, Szczechla, Kim, Chhablani, Nayak, Datta, Chang, Jiang, Wang, Manica, Shen, Yong, Pandey, Bawden, Wang, Neeraj, Rozen, Sharma, Santilli, Fevry, Fries, Teehan, Scao, Biderman, Gao, Wolf, and Rush]{t_zero}
Victor Sanh, Albert Webson, Colin Raffel, Stephen Bach, Lintang Sutawika, Zaid Alyafeai, Antoine Chaffin, Arnaud Stiegler, Arun Raja, Manan Dey, M~Saiful Bari, Canwen Xu, Urmish Thakker, Shanya~Sharma Sharma, Eliza Szczechla, Taewoon Kim, Gunjan Chhablani, Nihal Nayak, Debajyoti Datta, Jonathan Chang, Mike Tian-Jian Jiang, Han Wang, Matteo Manica, Sheng Shen, Zheng~Xin Yong, Harshit Pandey, Rachel Bawden, Thomas Wang, Trishala Neeraj, Jos Rozen, Abheesht Sharma, Andrea Santilli, Thibault Fevry, Jason~Alan Fries, Ryan Teehan, Teven~Le Scao, Stella Biderman, Leo Gao, Thomas Wolf, and Alexander~M Rush.
\newblock Multitask prompted training enables zero-shot task generalization.
\newblock In \emph{International Conference on Learning Representations}, 2022.
\newblock URL \url{https://openreview.net/forum?id=9Vrb9D0WI4}.

\bibitem[Scao et~al.(2022)Scao, Fan, Akiki, Pavlick, Ili'c, Hesslow, Castagn'e, Luccioni, Yvon, Gall{\'e}, Tow, Rush, Biderman, Webson, Ammanamanchi, Wang, Sagot, Muennighoff, del Moral, Ruwase, Bawden, Bekman, McMillan-Major, Beltagy, Nguyen, Saulnier, Tan, Suarez, Sanh, Laurenccon, Jernite, Launay, Mitchell, Raffel, Gokaslan, Simhi, Etxabe, Aji, Alfassy, Rogers, Nitzav, Xu, Mou, Emezue, Klamm, Leong, van Strien, Adelani, Radev, Ponferrada, Levkovizh, Kim, Natan, Toni, Dupont, Kruszewski, Pistilli, ElSahar, Benyamina, Tran, Yu, Abdulmumin, Johnson, Gonzalez-Dios, de~la Rosa, Chim, Dodge, Zhu, Chang, Frohberg, Tobing, Bhattacharjee, Almubarak, Chen, Lo, von Werra, Weber, Phan, Allal, Tanguy, Dey, Mu{\~n}oz, Masoud, Grandury, vSavsko, Huang, Coavoux, Singh, Jiang, Vu, Jauhar, Ghaleb, Subramani, Kassner, Khamis, Nguyen, Espejel, de~Gibert, Villegas, Henderson, Colombo, Amuok, Lhoest, Harliman, Bommasani, L'opez, Ribeiro, Osei, Pyysalo, Nagel, Bose, Muhammad, Sharma, Longpre, Nikpoor, Silberberg, Pai, Zink,
  Torrent, Schick, Thrush, Danchev, Nikoulina, Laippala, Lepercq, Prabhu, Alyafeai, Talat, Raja, Heinzerling, Si, Salesky, Mielke, Lee, Sharma, Santilli, Chaffin, Stiegler, Datta, Szczechla, Chhablani, Wang, Pandey, Strobelt, Fries, Rozen, Gao, Sutawika, Bari, Al-shaibani, Manica, Nayak, Teehan, Albanie, Shen, Ben-David, Bach, Kim, Bers, F{\'e}vry, Neeraj, Thakker, Raunak, Tang, Yong, Sun, Brody, Uri, Tojarieh, Roberts, Chung, Tae, Phang, Press, Li, Narayanan, Bourfoune, Casper, Rasley, Ryabinin, Mishra, Zhang, Shoeybi, Peyrounette, Patry, Tazi, Sanseviero, von Platen, Cornette, Lavall'ee, Lacroix, Rajbhandari, Gandhi, Smith, Requena, Patil, Dettmers, Baruwa, Singh, Cheveleva, Ligozat, Subramonian, N'ev'eol, Lovering, Garrette, Tunuguntla, Reiter, Taktasheva, Voloshina, Bogdanov, Winata, Schoelkopf, Kalo, Novikova, Forde, Clive, Kasai, Kawamura, Hazan, Carpuat, Clinciu, Kim, Cheng, Serikov, Antverg, van~der Wal, Zhang, Zhang, Gehrmann, Pais, Shavrina, Scialom, Yun, Limisiewicz, Rieser, Protasov, Mikhailov,
  Pruksachatkun, Belinkov, Bamberger, Kasner, Rueda, Pestana, Feizpour, Khan, Faranak, Santos, Hevia, Unldreaj, Aghagol, Abdollahi, Tammour, HajiHosseini, Behroozi, Ajibade, Saxena, Ferrandis, Contractor, Lansky, David, Kiela, Nguyen, Tan, Baylor, Ozoani, Mirza, Ononiwu, Rezanejad, Jones, Bhattacharya, Solaiman, Sedenko, Nejadgholi, Passmore, Seltzer, Sanz, Fort, Dutra, Samagaio, Elbadri, Mieskes, Gerchick, Akinlolu, McKenna, Qiu, Ghauri, Burynok, Abrar, Rajani, Elkott, Fahmy, Samuel, An, Kromann, Hao, Alizadeh, Shubber, Wang, Roy, Viguier, Le, Oyebade, Le, Yang, Nguyen, Kashyap, Palasciano, Callahan, Shukla, Miranda-Escalada, Singh, Beilharz, Wang, de~Brito, Zhou, Jain, Xu, Fourrier, Perin'an, Molano, Yu, Manjavacas, Barth, Fuhrimann, Altay, Bayrak, Burns, Vrabec, Bello, Dash, Kang, Giorgi, Golde, Posada, Sivaraman, Bulchandani, Liu, Shinzato, de~Bykhovetz, Takeuchi, P{\`a}mies, Castillo, Nezhurina, Sanger, Samwald, Cullan, Weinberg, Wolf, Mihaljcic, Liu, Freidank, Kang, Seelam, Dahlberg, Broad, Muellner,
  Fung, Haller, Chandrasekhar, Eisenberg, Martin, Canalli, Su, Su, Cahyawijaya, Garda, Deshmukh, Mishra, Kiblawi, Ott, Sang-aroonsiri, Kumar, Schweter, Bharati, Laud, Gigant, Kainuma, Kusa, Labrak, Bajaj, Venkatraman, Xu, Xu, chao Xu, Tan, Xie, Ye, Bras, Belkada, and Wolf]{bloom}
Teven~Le Scao, Angela Fan, Christopher Akiki, Elizabeth-Jane Pavlick, Suzana Ili'c, Daniel Hesslow, Roman Castagn'e, Alexandra~Sasha Luccioni, Franccois Yvon, Matthias Gall{\'e}, Jonathan Tow, Alexander~M. Rush, Stella~Rose Biderman, Albert Webson, Pawan~Sasanka Ammanamanchi, Thomas Wang, Beno{\^i}t Sagot, Niklas Muennighoff, Albert~Villanova del Moral, Olatunji Ruwase, Rachel Bawden, Stas Bekman, Angelina McMillan-Major, Iz~Beltagy, Huu Nguyen, Lucile Saulnier, Samson Tan, Pedro~Ortiz Suarez, Victor Sanh, Hugo Laurenccon, Yacine Jernite, Julien Launay, Margaret Mitchell, Colin Raffel, Aaron Gokaslan, Adi Simhi, Aitor~Soroa Etxabe, Alham~Fikri Aji, Amit Alfassy, Anna Rogers, Ariel~Kreisberg Nitzav, Canwen Xu, Chenghao Mou, Chris~C. Emezue, Christopher Klamm, Colin Leong, Daniel~Alexander van Strien, David~Ifeoluwa Adelani, Dragomir~R. Radev, Eduardo~G. Ponferrada, Efrat Levkovizh, Ethan Kim, Eyal~Bar Natan, Francesco~De Toni, G{\'e}rard Dupont, Germ{\'a}n Kruszewski, Giada Pistilli, Hady ElSahar, Hamza
  Benyamina, Hieu Tran, Ian Yu, Idris Abdulmumin, Isaac Johnson, Itziar Gonzalez-Dios, Javier de~la Rosa, Jenny Chim, Jesse Dodge, Jian Zhu, Jonathan Chang, Jorg Frohberg, Josephine~L. Tobing, Joydeep Bhattacharjee, Khalid Almubarak, Kimbo Chen, Kyle Lo, Leandro von Werra, Leon Weber, Long Phan, Loubna~Ben Allal, Ludovic Tanguy, Manan Dey, Manuel~Romero Mu{\~n}oz, Maraim Masoud, Mar'ia Grandury, Mario vSavsko, Max Huang, Maximin Coavoux, Mayank Singh, Mike Tian-Jian Jiang, Minh~Chien Vu, Mohammad~Ali Jauhar, Mustafa Ghaleb, Nishant Subramani, Nora Kassner, Nurulaqilla Khamis, Olivier Nguyen, Omar Espejel, Ona de~Gibert, Paulo Villegas, Peter Henderson, Pierre Colombo, Priscilla Amuok, Quentin Lhoest, Rheza Harliman, Rishi Bommasani, Roberto L'opez, Rui Ribeiro, Salomey Osei, Sampo Pyysalo, Sebastian Nagel, Shamik Bose, Shamsuddeen~Hassan Muhammad, Shanya Sharma, S.~Longpre, Somaieh Nikpoor, Stanislav Silberberg, Suhas Pai, Sydney Zink, Tiago~Timponi Torrent, Timo Schick, Tristan Thrush, Valentin Danchev,
  Vassilina Nikoulina, Veronika Laippala, Violette Lepercq, Vrinda Prabhu, Zaid Alyafeai, Zeerak Talat, Arun Raja, Benjamin Heinzerling, Chenglei Si, Elizabeth Salesky, Sabrina~J. Mielke, Wilson~Y. Lee, Abheesht Sharma, Andrea Santilli, Antoine Chaffin, Arnaud Stiegler, Debajyoti Datta, Eliza Szczechla, Gunjan Chhablani, Han Wang, Harshit Pandey, Hendrik Strobelt, Jason~Alan Fries, Jos Rozen, Leo Gao, Lintang Sutawika, M~Saiful Bari, Maged~S. Al-shaibani, Matteo Manica, Nihal~V. Nayak, Ryan Teehan, Samuel Albanie, Sheng Shen, Srulik Ben-David, Stephen~H. Bach, Taewoon Kim, Tali Bers, Thibault F{\'e}vry, Trishala Neeraj, Urmish Thakker, Vikas Raunak, Xiang Tang, Zheng~Xin Yong, Zhiqing Sun, Shaked Brody, Y~Uri, Hadar Tojarieh, Adam Roberts, Hyung~Won Chung, Jaesung Tae, Jason Phang, Ofir Press, Conglong Li, Deepak Narayanan, Hatim Bourfoune, Jared Casper, Jeff Rasley, Max Ryabinin, Mayank Mishra, Minjia Zhang, Mohammad Shoeybi, Myriam Peyrounette, Nicolas Patry, Nouamane Tazi, Omar Sanseviero, Patrick von
  Platen, Pierre Cornette, Pierre~Franccois Lavall'ee, R{\'e}mi Lacroix, Samyam Rajbhandari, Sanchit Gandhi, Shaden Smith, St{\'e}phane Requena, Suraj Patil, Tim Dettmers, Ahmed Baruwa, Amanpreet Singh, Anastasia Cheveleva, Anne-Laure Ligozat, Arjun Subramonian, Aur'elie N'ev'eol, Charles Lovering, Daniel~H Garrette, Deepak~R. Tunuguntla, Ehud Reiter, Ekaterina Taktasheva, Ekaterina Voloshina, Eli Bogdanov, Genta~Indra Winata, Hailey Schoelkopf, Jan-Christoph Kalo, Jekaterina Novikova, Jessica~Zosa Forde, Jordan Clive, Jungo Kasai, Ken Kawamura, Liam Hazan, Marine Carpuat, Miruna Clinciu, Najoung Kim, Newton Cheng, Oleg Serikov, Omer Antverg, Oskar van~der Wal, Rui Zhang, Ruochen Zhang, Sebastian Gehrmann, S.~Osher Pais, Tatiana Shavrina, Thomas Scialom, Tian Yun, Tomasz Limisiewicz, Verena Rieser, Vitaly Protasov, Vladislav Mikhailov, Yada Pruksachatkun, Yonatan Belinkov, Zachary Bamberger, Zdenvek Kasner, Alice Rueda, Amanda Pestana, Amir Feizpour, Ammar Khan, Amy Faranak, Ananda Santa~Rosa Santos, Anthony
  Hevia, Antigona Unldreaj, Arash Aghagol, Arezoo Abdollahi, Aycha Tammour, Azadeh HajiHosseini, Bahareh Behroozi, Benjamin~Olusola Ajibade, Bharat~Kumar Saxena, Carlos~Mu{\~n}oz Ferrandis, Danish Contractor, David~M. Lansky, Davis David, Douwe Kiela, Duong~Anh Nguyen, Edward Tan, Emily Baylor, Ezinwanne Ozoani, Fatim~T Mirza, Frankline Ononiwu, Habib Rezanejad, H.A. Jones, Indrani Bhattacharya, Irene Solaiman, Irina Sedenko, Isar Nejadgholi, Jan Passmore, Joshua Seltzer, Julio~Bonis Sanz, Karen Fort, L{\'i}via~Macedo Dutra, Mairon Samagaio, Maraim Elbadri, Margot Mieskes, Marissa Gerchick, Martha Akinlolu, Michael McKenna, Mike Qiu, M.~K.~K. Ghauri, Mykola Burynok, Nafis Abrar, Nazneen Rajani, Nour Elkott, Nourhan Fahmy, Olanrewaju~Modupe Samuel, Ran An, R.~P. Kromann, Ryan Hao, Samira Alizadeh, Sarmad Shubber, Silas~L. Wang, Sourav Roy, Sylvain Viguier, Thanh-Cong Le, Tobi Oyebade, Trieu Nguyen~Hai Le, Yoyo Yang, Zachary~Kyle Nguyen, Abhinav~Ramesh Kashyap, Alfredo Palasciano, Alison Callahan, Anima Shukla,
  Antonio Miranda-Escalada, Ayush~Kumar Singh, Benjamin Beilharz, Bo~Wang, Caio Matheus~Fonseca de~Brito, Chenxi Zhou, Chirag Jain, Chuxin Xu, Cl{\'e}mentine Fourrier, Daniel~Le'on Perin'an, Daniel Molano, Dian Yu, Enrique Manjavacas, Fabio Barth, Florian Fuhrimann, Gabriel Altay, Giyaseddin Bayrak, Gully~A. Burns, Helena~U. Vrabec, Iman~I.B. Bello, Isha Dash, Ji~Soo Kang, John Giorgi, Jonas Golde, Jose~David Posada, Karthi Sivaraman, Lokesh Bulchandani, Lu~Liu, Luisa Shinzato, Madeleine~Hahn de~Bykhovetz, Maiko Takeuchi, Marc P{\`a}mies, Mar{\'i}a~Andrea Castillo, Marianna Nezhurina, Mario Sanger, Matthias Samwald, Michael Cullan, Michael Weinberg, M~Wolf, Mina Mihaljcic, Minna Liu, Moritz Freidank, Myungsun Kang, Natasha Seelam, Nathan Dahlberg, Nicholas~Michio Broad, Nikolaus Muellner, Pascale Fung, Patricia Haller, R.~Chandrasekhar, R.~Eisenberg, Robert Martin, Rodrigo~L. Canalli, Rosaline Su, Ruisi Su, Samuel Cahyawijaya, Samuele Garda, Shlok~S Deshmukh, Shubhanshu Mishra, Sid Kiblawi, Simon Ott, Sinee
  Sang-aroonsiri, Srishti Kumar, Stefan Schweter, Sushil~Pratap Bharati, T.~A. Laud, Th'eo Gigant, Tomoya Kainuma, Wojciech Kusa, Yanis Labrak, Yashasvi Bajaj, Y.~Venkatraman, Yifan Xu, Ying Xu, Yun chao Xu, Zhee~Xao Tan, Zhongli Xie, Zifan Ye, Mathilde Bras, Younes Belkada, and Thomas Wolf.
\newblock Bloom: A 176b-parameter open-access multilingual language model.
\newblock \emph{ArXiv}, abs/2211.05100v2, 2022.

\bibitem[Schick and Sch{\"u}tze(2021)]{pet}
Timo Schick and Hinrich Sch{\"u}tze.
\newblock It{'}s not just size that matters: Small language models are also few-shot learners.
\newblock In \emph{Proceedings of the 2021 Conference of the North American Chapter of the Association for Computational Linguistics: Human Language Technologies}, pages 2339--2352, Online, June 2021. Association for Computational Linguistics.
\newblock \doi{10.18653/v1/2021.naacl-main.185}.
\newblock URL \url{https://aclanthology.org/2021.naacl-main.185}.

\bibitem[See et~al.(2017)See, Liu, and Manning]{cnn_dailymail}
Abigail See, Peter~J. Liu, and Christopher~D. Manning.
\newblock Get to the point: Summarization with pointer-generator networks.
\newblock In \emph{Proceedings of the 55th Annual Meeting of the Association for Computational Linguistics (Volume 1: Long Papers)}, pages 1073--1083, Vancouver, Canada, July 2017. Association for Computational Linguistics.
\newblock \doi{10.18653/v1/P17-1099}.
\newblock URL \url{https://www.aclweb.org/anthology/P17-1099}.

\bibitem[Shazeer et~al.(2017)Shazeer, Mirhoseini, Maziarz, Davis, Le, Hinton, and Dean]{moe}
Noam Shazeer, *Azalia Mirhoseini, *Krzysztof Maziarz, Andy Davis, Quoc Le, Geoffrey Hinton, and Jeff Dean.
\newblock Outrageously large neural networks: The sparsely-gated mixture-of-experts layer.
\newblock In \emph{International Conference on Learning Representations}, 2017.
\newblock URL \url{https://openreview.net/forum?id=B1ckMDqlg}.

\bibitem[Shoeybi et~al.(2019)Shoeybi, Patwary, Puri, LeGresley, Casper, and Catanzaro]{megatron}
Mohammad Shoeybi, Mostofa Patwary, Raul Puri, Patrick LeGresley, Jared Casper, and Bryan Catanzaro.
\newblock Megatron-lm: Training multi-billion parameter language models using model parallelism.
\newblock \emph{ArXiv}, abs/1909.08053, 2019.

\bibitem[Srivastava et~al.(2022)Srivastava, Rastogi, Rao, Shoeb, Abid, Fisch, Brown, Santoro, Gupta, Garriga-Alonso, Kluska, Lewkowycz, Agarwal, Power, Ray, Warstadt, Kocurek, Safaya, Tazarv, Xiang, Parrish, Nie, Hussain, Askell, Dsouza, Slone, Rahane, Iyer, Andreassen, Madotto, Santilli, Stuhlmüller, Dai, La, Lampinen, Zou, Jiang, Chen, Vuong, Gupta, Gottardi, Norelli, Venkatesh, Gholamidavoodi, Tabassum, Menezes, Kirubarajan, Mullokandov, Sabharwal, Herrick, Efrat, Erdem, Karakaş, Roberts, Loe, Zoph, Bojanowski, Özyurt, Hedayatnia, Neyshabur, Inden, Stein, Ekmekci, Lin, Howald, Diao, Dour, Stinson, Argueta, Ramírez, Singh, Rathkopf, Meng, Baral, Wu, Callison-Burch, Waites, Voigt, Manning, Potts, Ramirez, Rivera, Siro, Raffel, Ashcraft, Garbacea, Sileo, Garrette, Hendrycks, Kilman, Roth, Freeman, Khashabi, Levy, González, Perszyk, Hernandez, Chen, Ippolito, Gilboa, Dohan, Drakard, Jurgens, Datta, Ganguli, Emelin, Kleyko, Yuret, Chen, Tam, Hupkes, Misra, Buzan, Mollo, Yang, Lee, Shutova, Cubuk, Segal,
  Hagerman, Barnes, Donoway, Pavlick, Rodola, Lam, Chu, Tang, Erdem, Chang, Chi, Dyer, Jerzak, Kim, Manyasi, Zheltonozhskii, Xia, Siar, Martínez-Plumed, Happé, Chollet, Rong, Mishra, Winata, de~Melo, Kruszewski, Parascandolo, Mariani, Wang, Jaimovitch-López, Betz, Gur-Ari, Galijasevic, Kim, Rashkin, Hajishirzi, Mehta, Bogar, Shevlin, Schütze, Yakura, Zhang, Wong, Ng, Noble, Jumelet, Geissinger, Kernion, Hilton, Lee, Fisac, Simon, Koppel, Zheng, Zou, Kocoń, Thompson, Kaplan, Radom, Sohl-Dickstein, Phang, Wei, Yosinski, Novikova, Bosscher, Marsh, Kim, Taal, Engel, Alabi, Xu, Song, Tang, Waweru, Burden, Miller, Balis, Berant, Frohberg, Rozen, Hernandez-Orallo, Boudeman, Jones, Tenenbaum, Rule, Chua, Kanclerz, Livescu, Krauth, Gopalakrishnan, Ignatyeva, Markert, Dhole, Gimpel, Omondi, Mathewson, Chiafullo, Shkaruta, Shridhar, McDonell, Richardson, Reynolds, Gao, Zhang, Dugan, Qin, Contreras-Ochando, Morency, Moschella, Lam, Noble, Schmidt, He, Colón, Metz, Şenel, Bosma, Sap, ter Hoeve, Farooqi, Faruqui,
  Mazeika, Baturan, Marelli, Maru, Quintana, Tolkiehn, Giulianelli, Lewis, Potthast, Leavitt, Hagen, Schubert, Baitemirova, Arnaud, McElrath, Yee, Cohen, Gu, Ivanitskiy, Starritt, Strube, Swędrowski, Bevilacqua, Yasunaga, Kale, Cain, Xu, Suzgun, Tiwari, Bansal, Aminnaseri, Geva, Gheini, T, Peng, Chi, Lee, Krakover, Cameron, Roberts, Doiron, Nangia, Deckers, Muennighoff, Keskar, Iyer, Constant, Fiedel, Wen, Zhang, Agha, Elbaghdadi, Levy, Evans, Casares, Doshi, Fung, Liang, Vicol, Alipoormolabashi, Liao, Liang, Chang, Eckersley, Htut, Hwang, Miłkowski, Patil, Pezeshkpour, Oli, Mei, Lyu, Chen, Banjade, Rudolph, Gabriel, Habacker, Delgado, Millière, Garg, Barnes, Saurous, Arakawa, Raymaekers, Frank, Sikand, Novak, Sitelew, LeBras, Liu, Jacobs, Zhang, Salakhutdinov, Chi, Lee, Stovall, Teehan, Yang, Singh, Mohammad, Anand, Dillavou, Shleifer, Wiseman, Gruetter, Bowman, Schoenholz, Han, Kwatra, Rous, Ghazarian, Ghosh, Casey, Bischoff, Gehrmann, Schuster, Sadeghi, Hamdan, Zhou, Srivastava, Shi, Singh, Asaadi, Gu,
  Pachchigar, Toshniwal, Upadhyay, Shyamolima, Debnath, Shakeri, Thormeyer, Melzi, Reddy, Makini, Lee, Torene, Hatwar, Dehaene, Divic, Ermon, Biderman, Lin, Prasad, Piantadosi, Shieber, Misherghi, Kiritchenko, Mishra, Linzen, Schuster, Li, Yu, Ali, Hashimoto, Wu, Desbordes, Rothschild, Phan, Wang, Nkinyili, Schick, Kornev, Telleen-Lawton, Tunduny, Gerstenberg, Chang, Neeraj, Khot, Shultz, Shaham, Misra, Demberg, Nyamai, Raunak, Ramasesh, Prabhu, Padmakumar, Srikumar, Fedus, Saunders, Zhang, Vossen, Ren, Tong, Zhao, Wu, Shen, Yaghoobzadeh, Lakretz, Song, Bahri, Choi, Yang, Hao, Chen, Belinkov, Hou, Hou, Bai, Seid, Zhao, Wang, Wang, Wang, and Wu]{srivastava2022imitation}
Aarohi Srivastava, Abhinav Rastogi, Abhishek Rao, Abu Awal~Md Shoeb, Abubakar Abid, Adam Fisch, Adam~R. Brown, Adam Santoro, Aditya Gupta, Adrià Garriga-Alonso, Agnieszka Kluska, Aitor Lewkowycz, Akshat Agarwal, Alethea Power, Alex Ray, Alex Warstadt, Alexander~W. Kocurek, Ali Safaya, Ali Tazarv, Alice Xiang, Alicia Parrish, Allen Nie, Aman Hussain, Amanda Askell, Amanda Dsouza, Ambrose Slone, Ameet Rahane, Anantharaman~S. Iyer, Anders Andreassen, Andrea Madotto, Andrea Santilli, Andreas Stuhlmüller, Andrew Dai, Andrew La, Andrew Lampinen, Andy Zou, Angela Jiang, Angelica Chen, Anh Vuong, Animesh Gupta, Anna Gottardi, Antonio Norelli, Anu Venkatesh, Arash Gholamidavoodi, Arfa Tabassum, Arul Menezes, Arun Kirubarajan, Asher Mullokandov, Ashish Sabharwal, Austin Herrick, Avia Efrat, Aykut Erdem, Ayla Karakaş, B.~Ryan Roberts, Bao~Sheng Loe, Barret Zoph, Bartłomiej Bojanowski, Batuhan Özyurt, Behnam Hedayatnia, Behnam Neyshabur, Benjamin Inden, Benno Stein, Berk Ekmekci, Bill~Yuchen Lin, Blake Howald,
  Cameron Diao, Cameron Dour, Catherine Stinson, Cedrick Argueta, César~Ferri Ramírez, Chandan Singh, Charles Rathkopf, Chenlin Meng, Chitta Baral, Chiyu Wu, Chris Callison-Burch, Chris Waites, Christian Voigt, Christopher~D. Manning, Christopher Potts, Cindy Ramirez, Clara~E. Rivera, Clemencia Siro, Colin Raffel, Courtney Ashcraft, Cristina Garbacea, Damien Sileo, Dan Garrette, Dan Hendrycks, Dan Kilman, Dan Roth, Daniel Freeman, Daniel Khashabi, Daniel Levy, Daniel~Moseguí González, Danielle Perszyk, Danny Hernandez, Danqi Chen, Daphne Ippolito, Dar Gilboa, David Dohan, David Drakard, David Jurgens, Debajyoti Datta, Deep Ganguli, Denis Emelin, Denis Kleyko, Deniz Yuret, Derek Chen, Derek Tam, Dieuwke Hupkes, Diganta Misra, Dilyar Buzan, Dimitri~Coelho Mollo, Diyi Yang, Dong-Ho Lee, Ekaterina Shutova, Ekin~Dogus Cubuk, Elad Segal, Eleanor Hagerman, Elizabeth Barnes, Elizabeth Donoway, Ellie Pavlick, Emanuele Rodola, Emma Lam, Eric Chu, Eric Tang, Erkut Erdem, Ernie Chang, Ethan~A. Chi, Ethan Dyer, Ethan
  Jerzak, Ethan Kim, Eunice~Engefu Manyasi, Evgenii Zheltonozhskii, Fanyue Xia, Fatemeh Siar, Fernando Martínez-Plumed, Francesca Happé, Francois Chollet, Frieda Rong, Gaurav Mishra, Genta~Indra Winata, Gerard de~Melo, Germán Kruszewski, Giambattista Parascandolo, Giorgio Mariani, Gloria Wang, Gonzalo Jaimovitch-López, Gregor Betz, Guy Gur-Ari, Hana Galijasevic, Hannah Kim, Hannah Rashkin, Hannaneh Hajishirzi, Harsh Mehta, Hayden Bogar, Henry Shevlin, Hinrich Schütze, Hiromu Yakura, Hongming Zhang, Hugh~Mee Wong, Ian Ng, Isaac Noble, Jaap Jumelet, Jack Geissinger, Jackson Kernion, Jacob Hilton, Jaehoon Lee, Jaime~Fernández Fisac, James~B. Simon, James Koppel, James Zheng, James Zou, Jan Kocoń, Jana Thompson, Jared Kaplan, Jarema Radom, Jascha Sohl-Dickstein, Jason Phang, Jason Wei, Jason Yosinski, Jekaterina Novikova, Jelle Bosscher, Jennifer Marsh, Jeremy Kim, Jeroen Taal, Jesse Engel, Jesujoba Alabi, Jiacheng Xu, Jiaming Song, Jillian Tang, Joan Waweru, John Burden, John Miller, John~U. Balis,
  Jonathan Berant, Jörg Frohberg, Jos Rozen, Jose Hernandez-Orallo, Joseph Boudeman, Joseph Jones, Joshua~B. Tenenbaum, Joshua~S. Rule, Joyce Chua, Kamil Kanclerz, Karen Livescu, Karl Krauth, Karthik Gopalakrishnan, Katerina Ignatyeva, Katja Markert, Kaustubh~D. Dhole, Kevin Gimpel, Kevin Omondi, Kory Mathewson, Kristen Chiafullo, Ksenia Shkaruta, Kumar Shridhar, Kyle McDonell, Kyle Richardson, Laria Reynolds, Leo Gao, Li~Zhang, Liam Dugan, Lianhui Qin, Lidia Contreras-Ochando, Louis-Philippe Morency, Luca Moschella, Lucas Lam, Lucy Noble, Ludwig Schmidt, Luheng He, Luis~Oliveros Colón, Luke Metz, Lütfi~Kerem Şenel, Maarten Bosma, Maarten Sap, Maartje ter Hoeve, Maheen Farooqi, Manaal Faruqui, Mantas Mazeika, Marco Baturan, Marco Marelli, Marco Maru, Maria Jose~Ramírez Quintana, Marie Tolkiehn, Mario Giulianelli, Martha Lewis, Martin Potthast, Matthew~L. Leavitt, Matthias Hagen, Mátyás Schubert, Medina~Orduna Baitemirova, Melody Arnaud, Melvin McElrath, Michael~A. Yee, Michael Cohen, Michael Gu,
  Michael Ivanitskiy, Michael Starritt, Michael Strube, Michał Swędrowski, Michele Bevilacqua, Michihiro Yasunaga, Mihir Kale, Mike Cain, Mimee Xu, Mirac Suzgun, Mo~Tiwari, Mohit Bansal, Moin Aminnaseri, Mor Geva, Mozhdeh Gheini, Mukund~Varma T, Nanyun Peng, Nathan Chi, Nayeon Lee, Neta Gur-Ari Krakover, Nicholas Cameron, Nicholas Roberts, Nick Doiron, Nikita Nangia, Niklas Deckers, Niklas Muennighoff, Nitish~Shirish Keskar, Niveditha~S. Iyer, Noah Constant, Noah Fiedel, Nuan Wen, Oliver Zhang, Omar Agha, Omar Elbaghdadi, Omer Levy, Owain Evans, Pablo Antonio~Moreno Casares, Parth Doshi, Pascale Fung, Paul~Pu Liang, Paul Vicol, Pegah Alipoormolabashi, Peiyuan Liao, Percy Liang, Peter Chang, Peter Eckersley, Phu~Mon Htut, Pinyu Hwang, Piotr Miłkowski, Piyush Patil, Pouya Pezeshkpour, Priti Oli, Qiaozhu Mei, Qing Lyu, Qinlang Chen, Rabin Banjade, Rachel~Etta Rudolph, Raefer Gabriel, Rahel Habacker, Ramón~Risco Delgado, Raphaël Millière, Rhythm Garg, Richard Barnes, Rif~A. Saurous, Riku Arakawa, Robbe
  Raymaekers, Robert Frank, Rohan Sikand, Roman Novak, Roman Sitelew, Ronan LeBras, Rosanne Liu, Rowan Jacobs, Rui Zhang, Ruslan Salakhutdinov, Ryan Chi, Ryan Lee, Ryan Stovall, Ryan Teehan, Rylan Yang, Sahib Singh, Saif~M. Mohammad, Sajant Anand, Sam Dillavou, Sam Shleifer, Sam Wiseman, Samuel Gruetter, Samuel~R. Bowman, Samuel~S. Schoenholz, Sanghyun Han, Sanjeev Kwatra, Sarah~A. Rous, Sarik Ghazarian, Sayan Ghosh, Sean Casey, Sebastian Bischoff, Sebastian Gehrmann, Sebastian Schuster, Sepideh Sadeghi, Shadi Hamdan, Sharon Zhou, Shashank Srivastava, Sherry Shi, Shikhar Singh, Shima Asaadi, Shixiang~Shane Gu, Shubh Pachchigar, Shubham Toshniwal, Shyam Upadhyay, Shyamolima, Debnath, Siamak Shakeri, Simon Thormeyer, Simone Melzi, Siva Reddy, Sneha~Priscilla Makini, Soo-Hwan Lee, Spencer Torene, Sriharsha Hatwar, Stanislas Dehaene, Stefan Divic, Stefano Ermon, Stella Biderman, Stephanie Lin, Stephen Prasad, Steven~T. Piantadosi, Stuart~M. Shieber, Summer Misherghi, Svetlana Kiritchenko, Swaroop Mishra, Tal
  Linzen, Tal Schuster, Tao Li, Tao Yu, Tariq Ali, Tatsu Hashimoto, Te-Lin Wu, Théo Desbordes, Theodore Rothschild, Thomas Phan, Tianle Wang, Tiberius Nkinyili, Timo Schick, Timofei Kornev, Timothy Telleen-Lawton, Titus Tunduny, Tobias Gerstenberg, Trenton Chang, Trishala Neeraj, Tushar Khot, Tyler Shultz, Uri Shaham, Vedant Misra, Vera Demberg, Victoria Nyamai, Vikas Raunak, Vinay Ramasesh, Vinay~Uday Prabhu, Vishakh Padmakumar, Vivek Srikumar, William Fedus, William Saunders, William Zhang, Wout Vossen, Xiang Ren, Xiaoyu Tong, Xinran Zhao, Xinyi Wu, Xudong Shen, Yadollah Yaghoobzadeh, Yair Lakretz, Yangqiu Song, Yasaman Bahri, Yejin Choi, Yichi Yang, Yiding Hao, Yifu Chen, Yonatan Belinkov, Yu~Hou, Yufang Hou, Yuntao Bai, Zachary Seid, Zhuoye Zhao, Zijian Wang, Zijie~J. Wang, Zirui Wang, and Ziyi Wu.
\newblock Beyond the imitation game: Quantifying and extrapolating the capabilities of language models, 2022.

\bibitem[Su et~al.(2021)Su, Wang, Qin, Chan, Lin, Liu, Li, Li, Hou, Sun, and Zhou]{prompt_mapping}
Yusheng Su, Xiaozhi Wang, Yujia Qin, Chi-Min Chan, Yankai Lin, Zhiyuan Liu, Peng Li, Juan-Zi Li, Lei Hou, Maosong Sun, and Jie Zhou.
\newblock On transferability of prompt tuning for natural language understanding.
\newblock \emph{ArXiv}, abs/2111.06719, 2021.

\bibitem[Sung et~al.(2021)Sung, Nair, and Raffel]{fish_mask}
Yi-Lin Sung, Varun Nair, and Colin~A Raffel.
\newblock Training neural networks with fixed sparse masks.
\newblock In M.~Ranzato, A.~Beygelzimer, Y.~Dauphin, P.S. Liang, and J.~Wortman Vaughan, editors, \emph{Advances in Neural Information Processing Systems}, volume~34, pages 24193--24205. Curran Associates, Inc., 2021.
\newblock URL \url{https://proceedings.neurips.cc/paper/2021/file/cb2653f548f8709598e8b5156738cc51-Paper.pdf}.

\bibitem[Sung et~al.(2022)Sung, Cho, and Bansal]{ladder_side_tuning}
Yi-Lin Sung, Jaemin Cho, and Mohit Bansal.
\newblock Lst: Ladder side-tuning for parameter and memory efficient transfer learning.
\newblock \emph{ArXiv}, abs/2206.06522, 2022.

\bibitem[Touvron et~al.(2023)Touvron, Lavril, Izacard, Martinet, Lachaux, Lacroix, Rozi{\`e}re, Goyal, Hambro, Azhar, et~al.]{touvron2023llama}
Hugo Touvron, Thibaut Lavril, Gautier Izacard, Xavier Martinet, Marie-Anne Lachaux, Timoth{\'e}e Lacroix, Baptiste Rozi{\`e}re, Naman Goyal, Eric Hambro, Faisal Azhar, et~al.
\newblock Llama: Open and efficient foundation language models.
\newblock \emph{arXiv preprint arXiv:2302.13971}, 2023.

\bibitem[Vaswani et~al.(2017)Vaswani, Shazeer, Parmar, Uszkoreit, Jones, Gomez, Kaiser, and Polosukhin]{vaswani2017attention}
Ashish Vaswani, Noam Shazeer, Niki Parmar, Jakob Uszkoreit, Llion Jones, Aidan~N Gomez, {\L}ukasz Kaiser, and Illia Polosukhin.
\newblock Attention is all you need.
\newblock In \emph{Advances in neural information processing systems}, pages 5998--6008, 2017.

\bibitem[Vu et~al.(2021)Vu, Lester, Constant, Al-Rfou, and Cer]{spot}
Tu~Vu, Brian Lester, Noah Constant, Rami Al-Rfou, and Daniel~Matthew Cer.
\newblock Spot: Better frozen model adaptation through soft prompt transfer.
\newblock In \emph{Annual Meeting of the Association for Computational Linguistics}, 2021.

\bibitem[Vucetic et~al.(2022)Vucetic, Tayaranian, Ziaeefard, Clark, Meyer, and Gross]{far_edge}
Danilo Vucetic, Mohammadreza Tayaranian, Maryam Ziaeefard, James~J. Clark, Brett~H. Meyer, and Warren~J. Gross.
\newblock Efficient fine-tuning of bert models on the edge.
\newblock \emph{2022 IEEE International Symposium on Circuits and Systems (ISCAS)}, pages 1838--1842, 2022.

\bibitem[Wang et~al.(2018)Wang, Singh, Michael, Hill, Levy, and Bowman]{wang2018glue}
Alex Wang, Amanpreet Singh, Julian Michael, Felix Hill, Omer Levy, and Samuel~R Bowman.
\newblock Glue: A multi-task benchmark and analysis platform for natural language understanding.
\newblock \emph{arXiv preprint arXiv:1804.07461}, 2018.

\bibitem[Wang et~al.(2019)Wang, Pruksachatkun, Nangia, Singh, Michael, Hill, Levy, and Bowman]{wang2019superglue}
Alex Wang, Yada Pruksachatkun, Nikita Nangia, Amanpreet Singh, Julian Michael, Felix Hill, Omer Levy, and Samuel~R Bowman.
\newblock Superglue: A stickier benchmark for general-purpose language understanding systems.
\newblock \emph{arXiv preprint arXiv:1905.00537}, 2019.

\bibitem[Wang et~al.(2022)Wang, Mukherjee, Liu, Gao, Awadallah, and Gao]{adamix}
Yaqing Wang, Subhabrata Mukherjee, Xiaodong Liu, Jing Gao, Ahmed~Hassan Awadallah, and Jianfeng Gao.
\newblock Adamix: Mixture-of-adapter for parameter-efficient tuning of large language models.
\newblock \emph{ArXiv}, abs/2205.12410, 2022.

\bibitem[Wolf et~al.(2020)Wolf, Debut, Sanh, Chaumond, Delangue, Moi, Cistac, Ma, Jernite, Plu, Xu, Le~Scao, Gugger, Drame, Lhoest, and Rush]{Wolf_Transformers_State-of-the-Art_Natural_2020}
Thomas Wolf, Lysandre Debut, Victor Sanh, Julien Chaumond, Clement Delangue, Anthony Moi, Perric Cistac, Clara Ma, Yacine Jernite, Julien Plu, Canwen Xu, Teven Le~Scao, Sylvain Gugger, Mariama Drame, Quentin Lhoest, and Alexander~M. Rush.
\newblock {Transformers: State-of-the-Art Natural Language Processing}.
\newblock pages 38--45. Association for Computational Linguistics, 10 2020.
\newblock URL \url{https://www.aclweb.org/anthology/2020.emnlp-demos.6}.

\bibitem[Xie et~al.(2023)Xie, Yao, Shi, Liu, Zhou, Liu, Li, and Li]{xie2023difffit}
Enze Xie, Lewei Yao, Han Shi, Zhili Liu, Daquan Zhou, Zhaoqiang Liu, Jiawei Li, and Zhenguo Li.
\newblock Difffit: Unlocking transferability of large diffusion models via simple parameter-efficient fine-tuning.
\newblock \emph{arXiv preprint arXiv:2304.06648}, 2023.

\bibitem[Xu et~al.(2023)Xu, Xie, Gu, Chen, Chang, Zhang, Chen, Zhang, and Tian]{xu2023qa}
Yuhui Xu, Lingxi Xie, Xiaotao Gu, Xin Chen, Heng Chang, Hengheng Zhang, Zhensu Chen, Xiaopeng Zhang, and Qi~Tian.
\newblock Qa-lora: Quantization-aware low-rank adaptation of large language models.
\newblock \emph{arXiv preprint arXiv:2309.14717}, 2023.

\bibitem[Zeng et~al.(2022)Zeng, Liu, Du, Wang, Lai, Ding, Yang, Xu, Zheng, Xia, Tam, Ma, Xue, Zhai, Chen, Zhang, Dong, and Tang]{zeng2022glm130b}
Aohan Zeng, Xiao Liu, Zhengxiao Du, Zihan Wang, Hanyu Lai, Ming Ding, Zhuoyi Yang, Yifan Xu, Wendi Zheng, Xiao Xia, Weng~Lam Tam, Zixuan Ma, Yufei Xue, Jidong Zhai, Wenguang Chen, Peng Zhang, Yuxiao Dong, and Jie Tang.
\newblock Glm-130b: An open bilingual pre-trained model, 2022.

\bibitem[Zhang et~al.(2021)Zhang, Tay, Zhang, Chan, Luu, Hui, and Fu]{pha}
Aston Zhang, Yi~Tay, Shuai Zhang, Alvin Chan, Anh~Tuan Luu, Siu~Cheung Hui, and Jie Fu.
\newblock Beyond fully-connected layers with quaternions: Parameterization of hypercomplex multiplications with $1/n$ parameters.
\newblock In \emph{International Conference on Learning Representations}, 2021.

\bibitem[Zhang et~al.(2023{\natexlab{a}})Zhang, Chen, Bukharin, He, Cheng, Chen, and Zhao]{zhang2023adaptive}
Qingru Zhang, Minshuo Chen, Alexander Bukharin, Pengcheng He, Yu~Cheng, Weizhu Chen, and Tuo Zhao.
\newblock Adaptive budget allocation for parameter-efficient fine-tuning.
\newblock \emph{arXiv preprint arXiv:2303.10512}, 2023{\natexlab{a}}.

\bibitem[Zhang et~al.(2022)Zhang, Roller, Goyal, Artetxe, Chen, Chen, Dewan, Diab, Li, Lin, Mihaylov, Ott, Shleifer, Shuster, Simig, Koura, Sridhar, Wang, and Zettlemoyer]{zhang2022opt}
Susan Zhang, Stephen Roller, Naman Goyal, Mikel Artetxe, Moya Chen, Shuohui Chen, Christopher Dewan, Mona Diab, Xian Li, Xi~Victoria Lin, Todor Mihaylov, Myle Ott, Sam Shleifer, Kurt Shuster, Daniel Simig, Punit~Singh Koura, Anjali Sridhar, Tianlu Wang, and Luke Zettlemoyer.
\newblock Opt: Open pre-trained transformer language models, 2022.

\bibitem[Zhang et~al.(2023{\natexlab{b}})Zhang, Tan, Xu, Wang, Huang, and Huang]{zhang2023towards}
Zhen-Ru Zhang, Chuanqi Tan, Haiyang Xu, Chengyu Wang, Jun Huang, and Songfang Huang.
\newblock Towards adaptive prefix tuning for parameter-efficient language model fine-tuning.
\newblock \emph{arXiv preprint arXiv:2305.15212}, 2023{\natexlab{b}}.

\bibitem[Zhao et~al.(2024)Zhao, Zhang, Chen, Wang, Anandkumar, and Tian]{zhao2024galore}
Jiawei Zhao, Zhenyu Zhang, Beidi Chen, Zhangyang Wang, Anima Anandkumar, and Yuandong Tian.
\newblock Galore: Memory-efficient llm training by gradient low-rank projection.
\newblock \emph{arXiv preprint arXiv:2403.03507}, 2024.

\bibitem[Zhu et~al.(2021)Zhu, Feng, Zhao, Wang, and Li]{parallel_adapter2}
Yaoming Zhu, Jiangtao Feng, Chengqi Zhao, Mingxuan Wang, and Lei Li.
\newblock Counter-interference adapter for multilingual machine translation.
\newblock In \emph{Findings of the Association for Computational Linguistics: EMNLP 2021}, pages 2812--2823, Punta Cana, Dominican Republic, November 2021. Association for Computational Linguistics.
\newblock \doi{10.18653/v1/2021.findings-emnlp.240}.
\newblock URL \url{https://aclanthology.org/2021.findings-emnlp.240}.

\end{thebibliography}

\appendix

\section{Prompts}
\label{Appendix:prompts}

Prompts used in our experimental comparison
\begin{itemize}
    \item BoolQ: "Given a passage and a yes/no question, identify if the answer is \"yes\" or \"no\"."
    \item CB: "Given a premise, identify if the hypothesis entails, contradicts or is neutral to the premise."
    \item COPA: "Given a premise, a question (cause/effect) and two alternative choices, identify plausible answer from the alternative choices."
    \item RTE: "Given a premise, identify if the hypothesis entails premise or not."
\end{itemize}

\section{PEFT Comparison: full experimental results}
\label{sec:peft_comparison_raw}

The following table contains raw results used to produce Tables \ref{tab:peft_comparison_dataset_average_short} and \ref{tab:peft_comparison_dataset_average_tall}. Before these experiments, each model-method-dataset combination was swept over learning rates 1e-3, 1e-4, 5e-5. In our initial experiments we also included a sweep over weight decay (0 and 0.1), but we found it to minimally affect the results of these models on these datasets, always less than 0.01.


{
\fontsize{9pt}{9pt}\selectfont 
\renewcommand{\arraystretch}{0.6} 
\begin{longtable}{l|lll|l}
        \toprule
        Model & Method & Dataset & Seed & Score \\
        \midrule
        T5 3B & Full tuning & RTE & ~ & 90.70 \\ 
        T5 3B & Full tuning & COPA & ~ & 92.00 \\ 
        T5 3B & Full tuning & BoolQ & ~ & 89.30 \\ 
        T5 3B & Full tuning & CNN & ~ & 27.32 \\ 
        
        \midrule
        T5 3B & Adapters (Houlsby) & RTE & 0 & 90.61 \\ 
        T5 3B & Adapters (Houlsby) & RTE & 1 & 90.25 \\ 
        T5 3B & Adapters (Houlsby) & RTE & 42 & 90.61 \\ 
        T5 3B & Adapters (Houlsby) & COPA & 0 & 93.00 \\ 
        T5 3B & Adapters (Houlsby) & COPA & 1 & 92.00 \\ 
        T5 3B & Adapters (Houlsby) & COPA & 42 & 90.00 \\ 
        T5 3B & Adapters (Houlsby) & BoolQ & 0 & 89.17 \\ 
        T5 3B & Adapters (Houlsby) & BoolQ & 1 & 88.59 \\ 
        T5 3B & Adapters (Houlsby) & BoolQ & 42 & 88.35 \\ 
        T5 3B & Adapters (Houlsby) & CNN & 42 & 27.77 \\ 
        
        \midrule
        T5 3B & Adapters (Pfeiffer) & RTE & 0 & 86.63 \\ 
        T5 3B & Adapters (Pfeiffer) & RTE & 1 & 84.84 \\ 
        T5 3B & Adapters (Pfeiffer) & RTE & 42 & 85.92 \\ 
        T5 3B & Adapters (Pfeiffer) & COPA & 0 & 76.00 \\ 
        T5 3B & Adapters (Pfeiffer) & COPA & 1 & 70.00 \\ 
        T5 3B & Adapters (Pfeiffer) & COPA & 42 & 81.00 \\ 
        T5 3B & Adapters (Pfeiffer) & BoolQ & 0 & 88.53 \\ 
        T5 3B & Adapters (Pfeiffer) & BoolQ & 1 & 89.11 \\ 
        T5 3B & Adapters (Pfeiffer) & BoolQ & 42 & 88.81 \\ 
        T5 3B & Adapters (Pfeiffer) & CNN & 0 & 29.40 \\ 
        
        \midrule
        T5 3B & Scaled Parallel & RTE & 0 & 90.61 \\ 
        T5 3B & Scaled Parallel & RTE & 1 & 91.34 \\ 
        T5 3B & Scaled Parallel & RTE & 42 & 89.89 \\ 
        T5 3B & Scaled Parallel & COPA & 0 & 87.00 \\ 
        T5 3B & Scaled Parallel & COPA & 1 & 87.00 \\ 
        T5 3B & Scaled Parallel & COPA & 42 & 87.00 \\ 
        T5 3B & Scaled Parallel & BoolQ & 0 & 88.90 \\ 
        T5 3B & Scaled Parallel & BoolQ & 1 & 88.93 \\ 
        T5 3B & Scaled Parallel & BoolQ & 42 & 89.02 \\ 
        T5 3B & Scaled Parallel & CNN & 0 & 30.03 \\ 
        
        \midrule
        T5 3B & Ln tuning & RTE & 0 & 87.37 \\ 
        T5 3B & Ln tuning & RTE & 1 & 87.00 \\ 
        T5 3B & Ln tuning & RTE & 42 & 88.45 \\ 
        T5 3B & Ln tuning & COPA & 0 & 89.00 \\ 
        T5 3B & Ln tuning & COPA & 1 & 89.00 \\ 
        T5 3B & Ln tuning & COPA & 42 & 87.00 \\ 
        T5 3B & Ln tuning & BoolQ & 0 & 87.25 \\ 
        T5 3B & Ln tuning & BoolQ & 1 & 87.49 \\ 
        T5 3B & Ln tuning & BoolQ & 42 & 87.49 \\ 
        T5 3B & Ln tuning & CNN & 0 & 28.43 \\ 
        
        \midrule
        T5 3B & LoRa (q and v) & RTE & 0 & 91.70 \\ 
        T5 3B & LoRa (q and v) & RTE & 1 & 90.25 \\ 
        T5 3B & LoRa (q and v) & RTE & 42 & 89.17 \\ 
        T5 3B & LoRa (q and v) & COPA & 0 & 94.00 \\ 
        T5 3B & LoRa (q and v) & COPA & 1 & 92.00 \\ 
        T5 3B & LoRa (q and v) & COPA & 42 & 93.00 \\ 
        T5 3B & LoRa (q and v) & BoolQ & 0 & 89.08 \\ 
        T5 3B & LoRa (q and v) & BoolQ & 1 & 88.69 \\ 
        T5 3B & LoRa (q and v) & BoolQ & 42 & 88.53 \\ 
        T5 3B & LoRa (q and v) & CNN & 0 & 29.80 \\ 
        \midrule
        T5 3B & LoRa (all) & RTE & 0 & 89.89 \\ 
        T5 3B & LoRa (all) & RTE & 1 & 89.89 \\ 
        T5 3B & LoRa (all) & RTE & 42 & 89.53 \\ 
        T5 3B & LoRa (all) & COPA & 0 & 94.00 \\ 
        T5 3B & LoRa (all) & COPA & 1 & 94.00 \\ 
        T5 3B & LoRa (all) & COPA & 42 & 92.00 \\ 
        T5 3B & LoRa (all) & BoolQ & 0 & 88.65 \\ 
        T5 3B & LoRa (all) & BoolQ & 1 & 88.78 \\ 
        T5 3B & LoRa (all) & BoolQ & 42 & 88.81 \\ 
        T5 3B & LoRa (all) & CNN & 0 & 29.02 \\ 
        \midrule
        T5 3B & Krona & RTE & 0 & 88.09 \\ 
        T5 3B & Krona & RTE & 1 & 88.09 \\ 
        T5 3B & Krona & RTE & 42 & 88.45 \\ 
        T5 3B & Krona & COPA & 0 & 84.00 \\ 
        T5 3B & Krona & COPA & 1 & 84.00 \\ 
        T5 3B & Krona & COPA & 42 & 84.00 \\ 
        T5 3B & Krona & BoolQ & 0 & 87.68 \\ 
        T5 3B & Krona & BoolQ & 1 & 87.98 \\ 
        T5 3B & Krona & BoolQ & 42 & 87.68 \\ 
        T5 3B & Krona & CNN & 0 & 27.94 \\ 
        \midrule
        T5 3B & Compacter & RTE & 0 & 88.09 \\ 
        T5 3B & Compacter & RTE & 1 & 88.09 \\ 
        T5 3B & Compacter & RTE & 42 & 88.81 \\ 
        T5 3B & Compacter & COPA & 0 & 79.00 \\ 
        T5 3B & Compacter & COPA & 1 & 79.00 \\ 
        T5 3B & Compacter & COPA & 42 & 80.00 \\ 
        T5 3B & Compacter & BoolQ & 0 & 87.50 \\ 
        T5 3B & Compacter & BoolQ & 1 & 87.65 \\ 
        T5 3B & Compacter & BoolQ & 42 & 87.61 \\ 
        T5 3B & Compacter & CNN & 0 & 27.64 \\ 
        \midrule
        T5 3B & Compacter++ & RTE & 0 & 88.45 \\ 
        T5 3B & Compacter++ & RTE & 1 & 88.09 \\ 
        T5 3B & Compacter++ & RTE & 42 & 88.09 \\ 
        T5 3B & Compacter++ & COPA & 0 & 79.00 \\ 
        T5 3B & Compacter++ & COPA & 1 & 82.00 \\ 
        T5 3B & Compacter++ & COPA & 42 & 80.00 \\ 
        T5 3B & Compacter++ & BoolQ & 0 & 87.75 \\ 
        T5 3B & Compacter++ & BoolQ & 1 & 87.62 \\ 
        T5 3B & Compacter++ & BoolQ & 42 & 87.75 \\ 
        T5 3B & Compacter++ & CNN & 0 & 27.74 \\ 
        \midrule
        T5 3B & IA3 & RTE & 0 & 51.26 \\ 
        T5 3B & IA3 & RTE & 1 & 51.26 \\ 
        T5 3B & IA3 & RTE & 42 & 51.26 \\ 
        T5 3B & IA3 & COPA & 0 & 1.00 \\ 
        T5 3B & IA3 & COPA & 1 & 1.00 \\ 
        T5 3B & IA3 & COPA & 42 & 1.00 \\ 
        T5 3B & IA3 & BoolQ & 0 & 87.28 \\ 
        T5 3B & IA3 & BoolQ & 1 & 87.28 \\ 
        T5 3B & IA3 & BoolQ & 42 & 87.28 \\ 
        T5 3B & IA3 & CNN & 0 & 27.55 \\ 
        \midrule
        T5 3B & IA3 (min steps 500) & RTE & 0 & 51.26 \\ 
        T5 3B & IA3 (min steps 500) & RTE & 1 & 51.26 \\ 
        T5 3B & IA3 (min steps 500) & RTE & 42 & 51.26 \\ 
        T5 3B & IA3 (min steps 500) & COPA & 0 & 1.00 \\ 
        T5 3B & IA3 (min steps 500) & COPA & 1 & 1.00 \\ 
        T5 3B & IA3 (min steps 500) & COPA & 42 & 1.00 \\ 
        T5 3B & IA3 (min steps 500) & BoolQ & 0 & 88.12 \\ 
        T5 3B & IA3 (min steps 500) & BoolQ & 1 & 88.12 \\ 
        T5 3B & IA3 (min steps 500) & BoolQ & 42 & 88.12 \\ 
        T5 3B & IA3 (min steps 500) & CNN & 0 & ~ \\ 
        \midrule
        T5 3B & MAM & RTE & 0 & 52.71 \\ 
        T5 3B & MAM & RTE & 1 & 53.07 \\ 
        T5 3B & MAM & RTE & 42 & 53.43 \\ 
        T5 3B & MAM & COPA & 0 & 55.00 \\ 
        T5 3B & MAM & COPA & 1 & 55.00 \\ 
        T5 3B & MAM & COPA & 42 & 63.00 \\ 
        T5 3B & MAM & BoolQ & 0 & 64.36 \\ 
        T5 3B & MAM & BoolQ & 1 & 63.86 \\ 
        T5 3B & MAM & BoolQ & 42 & 64.36 \\ 
        T5 3B & MAM & CNN & 0 & 7.36 \\ 
        \midrule
        T5 3B & Prefix\_tuning & RTE & 0 & 55.23 \\ 
        T5 3B & Prefix\_tuning & RTE & 1 & 51.99 \\ 
        T5 3B & Prefix\_tuning & RTE & 42 & 53.79 \\ 
        T5 3B & Prefix\_tuning & COPA & 0 & 61.00 \\ 
        T5 3B & Prefix\_tuning & COPA & 1 & 58.00 \\ 
        T5 3B & Prefix\_tuning & COPA & 42 & 51.00 \\ 
        T5 3B & Prefix\_tuning & BoolQ & 0 & 64.85 \\ 
        T5 3B & Prefix\_tuning & BoolQ & 1 & 65.35 \\ 
        T5 3B & Prefix\_tuning & BoolQ & 42 & 63.86 \\ 
        T5 3B & Prefix\_tuning & CNN & 0 & 21.34 \\ 
        \midrule
        T5 3B & Prefix\_tuning\_flat & RTE & 0 & ~ \\ 
        T5 3B & Prefix\_tuning\_flat & RTE & 1 & ~ \\ 
        T5 3B & Prefix\_tuning\_flat & RTE & 42 & ~ \\ 
        T5 3B & Prefix\_tuning\_flat & COPA & 0 & ~ \\ 
        T5 3B & Prefix\_tuning\_flat & COPA & 1 & ~ \\ 
        T5 3B & Prefix\_tuning\_flat & COPA & 42 & ~ \\ 
        T5 3B & Prefix\_tuning\_flat & BoolQ & 0 & ~ \\ 
        T5 3B & Prefix\_tuning\_flat & BoolQ & 1 & ~ \\ 
        T5 3B & Prefix\_tuning\_flat & BoolQ & 42 & ~ \\ 
        T5 3B & Prefix\_tuning\_flat & CNN & 0 & 8.38 \\ 
        \midrule
        T5 3B & Unipelt & RTE & 0 & 53.07 \\ 
        T5 3B & Unipelt & RTE & 1 & 52.71 \\ 
        T5 3B & Unipelt & RTE & 42 & 54.51 \\ 
        T5 3B & Unipelt & COPA & 0 & 57.00 \\ 
        T5 3B & Unipelt & COPA & 1 & 57.00 \\ 
        T5 3B & Unipelt & COPA & 42 & 57.00 \\ 
        T5 3B & Unipelt & BoolQ & 0 & 64.36 \\ 
        T5 3B & Unipelt & BoolQ & 1 & 63.86 \\ 
        T5 3B & Unipelt & BoolQ & 42 & 55.94 \\ 
        T5 3B & Unipelt & CNN & 0 & 16.84 \\ 
        \midrule
        T5 11B & Full tuning & RTE & ~ & 91.09 \\ 
        T5 11B & Full tuning & COPA & ~ & 85.71 \\ 
        T5 11B & Full tuning & BoolQ & ~ & 88.89 \\ 
        T5 11B & Full tuning & CNN & ~ & 27.32 \\ 
        \midrule
        T5 11B & Adapters (Houlsby) & RTE & 0 & 92.06 \\ 
        T5 11B & Adapters (Houlsby) & RTE & 1 & 90.25 \\ 
        T5 11B & Adapters (Houlsby) & RTE & 42 & 90.97 \\ 
        T5 11B & Adapters (Houlsby) & COPA & 0 & 94.00 \\ 
        T5 11B & Adapters (Houlsby) & COPA & 1 & 93.00 \\ 
        T5 11B & Adapters (Houlsby) & COPA & 42 & 92.00 \\ 
        T5 11B & Adapters (Houlsby) & BoolQ & 0 & 92.57 \\ 
        T5 11B & Adapters (Houlsby) & BoolQ & 1 & 92.08 \\ 
        T5 11B & Adapters (Houlsby) & BoolQ & 42 & 92.57 \\ 
        T5 11B & Adapters (Houlsby) & CNN & 0 & 28.12 \\ 
        \midrule
        T5 11B & Adapters (Pfeiffer) & RTE & 0 & 52.71 \\ 
        T5 11B & Adapters (Pfeiffer) & RTE & 1 & 52.71 \\ 
        T5 11B & Adapters (Pfeiffer) & RTE & 42 & 54.51 \\ 
        T5 11B & Adapters (Pfeiffer) & COPA & 0 & 55.00 \\ 
        T5 11B & Adapters (Pfeiffer) & COPA & 1 & 55.00 \\ 
        T5 11B & Adapters (Pfeiffer) & COPA & 42 & 55.00 \\ 
        T5 11B & Adapters (Pfeiffer) & BoolQ & 0 & 63.82 \\ 
        T5 11B & Adapters (Pfeiffer) & BoolQ & 1 & 65.84 \\ 
        T5 11B & Adapters (Pfeiffer) & BoolQ & 42 & 63.61 \\ 
        T5 11B & Adapters (Pfeiffer) & CNN & 0 & 30.14 \\ 
        \midrule
        T5 11B & Scaled Parallel & RTE & 0 & 90.61 \\ 
        T5 11B & Scaled Parallel & RTE & 1 & 90.97 \\ 
        T5 11B & Scaled Parallel & RTE & 42 & 90.25 \\ 
        T5 11B & Scaled Parallel & COPA & 0 & 69.00 \\ 
        T5 11B & Scaled Parallel & COPA & 1 & 79.00 \\ 
        T5 11B & Scaled Parallel & COPA & 42 & 55.00 \\ 
        T5 11B & Scaled Parallel & BoolQ & 0 & 85.64 \\ 
        T5 11B & Scaled Parallel & BoolQ & 1 & 92.08 \\ 
        T5 11B & Scaled Parallel & BoolQ & 42 & 84.65 \\ 
        T5 11B & Scaled Parallel & CNN & 0 & 29.21 \\ 
        \midrule
        T5 11B & Ln tuning & RTE & 0 & 88.09 \\ 
        T5 11B & Ln tuning & RTE & 1 & 88.45 \\ 
        T5 11B & Ln tuning & RTE & 42 & 88.09 \\ 
        T5 11B & Ln tuning & COPA & 0 & 90.00 \\ 
        T5 11B & Ln tuning & COPA & 1 & 90.00 \\ 
        T5 11B & Ln tuning & COPA & 42 & 91.00 \\ 
        T5 11B & Ln tuning & BoolQ & 0 & 90.59 \\ 
        T5 11B & Ln tuning & BoolQ & 1 & 90.10 \\ 
        T5 11B & Ln tuning & BoolQ & 42 & 91.09 \\ 
        T5 11B & Ln tuning & CNN & 0 & 25.96 \\ 
        T5 11B & Ln tuning & CNN & 1 & ~ \\ 
        T5 11B & Ln tuning & CNN & 42 & ~ \\ 
        \midrule
        T5 11B & LoRa (q and v) & RTE & 0 & 89.89 \\ 
        T5 11B & LoRa (q and v) & RTE & 1 & 91.34 \\ 
        T5 11B & LoRa (q and v) & RTE & 42 & 91.34 \\ 
        T5 11B & LoRa (q and v) & COPA & 0 & 93.00 \\ 
        T5 11B & LoRa (q and v) & COPA & 1 & 92.00 \\ 
        T5 11B & LoRa (q and v) & COPA & 42 & 92.00 \\ 
        T5 11B & LoRa (q and v) & BoolQ & 0 & 92.08 \\ 
        T5 11B & LoRa (q and v) & BoolQ & 1 & 93.07 \\ 
        T5 11B & LoRa (q and v) & BoolQ & 42 & 92.08 \\ 
        T5 11B & LoRa (q and v) & CNN & 0 & 29.19 \\ 
        T5 11B & LoRa (q and v) & CNN & 1 & ~ \\ 
        T5 11B & LoRa (q and v) & CNN & 42 & ~ \\ 
        \midrule
        T5 11B & LoRa (all) & RTE & 0 & 92.78 \\ 
        T5 11B & LoRa (all) & RTE & 1 & 91.70 \\ 
        T5 11B & LoRa (all) & RTE & 42 & 90.97 \\ 
        T5 11B & LoRa (all) & COPA & 0 & 94.00 \\ 
        T5 11B & LoRa (all) & COPA & 1 & 94.00 \\ 
        T5 11B & LoRa (all) & COPA & 42 & 91.00 \\ 
        T5 11B & LoRa (all) & BoolQ & 0 & 92.57 \\ 
        T5 11B & LoRa (all) & BoolQ & 1 & 93.56 \\ 
        T5 11B & LoRa (all) & BoolQ & 42 & 92.08 \\ 
        T5 11B & LoRa (all) & CNN & 0 & 28.75 \\ 
        T5 11B & LoRa (all) & CNN & 1 & ~ \\ 
        T5 11B & LoRa (all) & CNN & 42 & ~ \\ 
        \midrule
        T5 11B & Krona & RTE & 0 & 86.11 \\ 
        T5 11B & Krona & RTE & 1 & 89.53 \\ 
        T5 11B & Krona & RTE & 42 & 85.71 \\ 
        T5 11B & Krona & COPA & 0 & 89.00 \\ 
        T5 11B & Krona & COPA & 1 & 89.00 \\ 
        T5 11B & Krona & COPA & 42 & 76.92 \\ 
        T5 11B & Krona & BoolQ & 0 & 88.61 \\ 
        T5 11B & Krona & BoolQ & 1 & ~ \\ 
        T5 11B & Krona & BoolQ & 42 & ~ \\ 
        T5 11B & Krona & CNN & 0 & 27.83 \\ 
        T5 11B & Krona & CNN & 1 & ~ \\ 
        T5 11B & Krona & CNN & 42 & ~ \\ 
        \midrule
        T5 11B & Compacter & RTE & 0 & 87.73 \\ 
        T5 11B & Compacter & RTE & 1 & 88.81 \\ 
        T5 11B & Compacter & RTE & 42 & 88.09 \\ 
        T5 11B & Compacter & COPA & 0 & 88.00 \\ 
        T5 11B & Compacter & COPA & 1 & 90.00 \\ 
        T5 11B & Compacter & COPA & 42 & 88.00 \\ 
        T5 11B & Compacter & BoolQ & 0 & 90.59 \\ 
        T5 11B & Compacter & BoolQ & 1 & 91.09 \\ 
        T5 11B & Compacter & BoolQ & 42 & 91.09 \\ 
        T5 11B & Compacter & CNN & 0 & 29.52 \\ 
        T5 11B & Compacter & CNN & 1 & ~ \\ 
        T5 11B & Compacter & CNN & 24 & ~ \\ 
        \midrule
        T5 11B & Compacter++ & RTE & 0 & 88.45 \\ 
        T5 11B & Compacter++ & RTE & 1 & 88.45 \\ 
        T5 11B & Compacter++ & RTE & 42 & 88.45 \\ 
        T5 11B & Compacter++ & COPA & 0 & 90.00 \\ 
        T5 11B & Compacter++ & COPA & 1 & 89.00 \\ 
        T5 11B & Compacter++ & COPA & 42 & 90.00 \\ 
        T5 11B & Compacter++ & BoolQ & 0 & 91.09 \\ 
        T5 11B & Compacter++ & BoolQ & 1 & 91.58 \\ 
        T5 11B & Compacter++ & BoolQ & 42 & 91.09 \\ 
        T5 11B & Compacter++ & CNN & 0 & 29.50 \\ 
        T5 11B & Compacter++ & CNN & 1 & ~ \\ 
        T5 11B & Compacter++ & CNN & 42 & ~ \\ 
        \midrule
        T5 11B & IA3 & RTE & 0 & 87.36 \\ 
        T5 11B & IA3 & RTE & 1 & ~ \\ 
        T5 11B & IA3 & RTE & 42 & ~ \\ 
        T5 11B & IA3 & COPA & 0 & 37.00 \\ 
        T5 11B & IA3 & COPA & 1 & ~ \\ 
        T5 11B & IA3 & COPA & 42 & ~ \\ 
        T5 11B & IA3 & BoolQ & 0 & 88.12 \\ 
        T5 11B & IA3 & BoolQ & 1 & ~ \\ 
        T5 11B & IA3 & BoolQ & 42 & ~ \\ 
        T5 11B & IA3 & CNN & 0 & 27.83 \\ 
        T5 11B & IA3 & CNN & 1 & ~ \\ 
        T5 11B & IA3 & CNN & 42 & ~ \\ 
        \midrule
        T5 11B & Prefix\_tuning & RTE & 0 & 54.15 \\ 
        T5 11B & Prefix\_tuning & RTE & 1 & 54.87 \\ 
        T5 11B & Prefix\_tuning & RTE & 42 & 53.79 \\ 
        T5 11B & Prefix\_tuning & COPA & 0 & 53.00 \\ 
        T5 11B & Prefix\_tuning & COPA & 1 & 55.00 \\ 
        T5 11B & Prefix\_tuning & COPA & 42 & 57.00 \\ 
        T5 11B & Prefix\_tuning & BoolQ & 0 & 70.30 \\ 
        T5 11B & Prefix\_tuning & BoolQ & 1 & 70.30 \\ 
        T5 11B & Prefix\_tuning & BoolQ & 42 & 71.29 \\ 
        T5 11B & Prefix\_tuning & CNN & 0 & 27.83 \\ 
        T5 11B & Prefix\_tuning & CNN & 1 & ~ \\ 
        T5 11B & Prefix\_tuning & CNN & 42 & ~ \\ 
        \midrule
        T5 11B & MAM & RTE & 0 & 53.07 \\ 
        T5 11B & MAM & RTE & 1 & 52.71 \\ 
        T5 11B & MAM & RTE & 42 & 52.71 \\ 
        T5 11B & MAM & COPA & 0 & 55.00 \\ 
        T5 11B & MAM & COPA & 1 & 55.00 \\ 
        T5 11B & MAM & COPA & 42 & 55.00 \\ 
        T5 11B & MAM & BoolQ & 0 & 70.30 \\ 
        T5 11B & MAM & BoolQ & 1 & 70.30 \\ 
        T5 11B & MAM & BoolQ & 42 & 70.30 \\ 
        T5 11B & MAM & CNN & 0 & 27.83 \\ 
        T5 11B & MAM & CNN & 1 & ~ \\ 
        T5 11B & MAM & CNN & 42 & ~ \\ 
        \midrule
        T5 11B & Unipelt & RTE & 0 & 52.71 \\ 
        T5 11B & Unipelt & RTE & 1 & 52.71 \\ 
        T5 11B & Unipelt & RTE & 42 & 52.71 \\ 
        T5 11B & Unipelt & COPA & 0 & 55.00 \\ 
        T5 11B & Unipelt & COPA & 1 & 61.00 \\ 
        T5 11B & Unipelt & COPA & 42 & 58.00 \\ 
        T5 11B & Unipelt & BoolQ & 0 & 70.30 \\ 
        T5 11B & Unipelt & BoolQ & 1 & 70.30 \\ 
        T5 11B & Unipelt & BoolQ & 42 & ~ \\ 
        T5 11B & Unipelt & CNN & 0 & 27.83 \\ 
        T5 11B & Unipelt & CNN & 1 & ~ \\ 
        T5 11B & Unipelt & CNN & 42 & ~ \\ 
        \midrule
        T5 Large & Full tuning & RTE & 42 & 87.00 \\ 
        T5 Large & Full tuning & RTE & 1 & 87.00 \\ 
        T5 Large & Full tuning & RTE & 0 & 86.28 \\ 
        T5 Large & Full tuning & COPA & 42 & 73.00 \\ 
        T5 Large & Full tuning & COPA & 1 & 70.00 \\ 
        T5 Large & Full tuning & COPA & 0 & 68.00 \\ 
        T5 Large & Full tuning & BoolQ & 42 & 85.38 \\ 
        T5 Large & Full tuning & BoolQ & 1 & 84.83 \\ 
        T5 Large & Full tuning & BoolQ & 0 & 84.43 \\ 
        T5 Large & Full tuning & CNN & 42 & 26.79 \\ 
        T5 Large & Full tuning & CNN & 1 & 26.76 \\ 
        T5 Large & Full tuning & CNN & 0 & 27.20 \\ 
        \midrule
        T5 Large & Adapters (Houlsby) & RTE & 42 & 87.73 \\ 
        T5 Large & Adapters (Houlsby) & RTE & 1 & 89.17 \\ 
        T5 Large & Adapters (Houlsby) & RTE & 0 & 89.53 \\ 
        T5 Large & Adapters (Houlsby) & COPA & 42 & 72.00 \\ 
        T5 Large & Adapters (Houlsby) & COPA & 1 & 75.00 \\ 
        T5 Large & Adapters (Houlsby) & COPA & 0 & 55.00 \\ 
        T5 Large & Adapters (Houlsby) & COPA & 42 & 60.00 \\ 
        T5 Large & Adapters (Houlsby) & COPA & 1 & 72.00 \\ 
        T5 Large & Adapters (Houlsby) & COPA & 0 & 80.00 \\ 
        T5 Large & Adapters (Houlsby) & BoolQ & 42 & 84.98 \\ 
        T5 Large & Adapters (Houlsby) & BoolQ & 1 & 84.53 \\ 
        T5 Large & Adapters (Houlsby) & BoolQ & 0 & 85.44 \\ 
        T5 Large & Adapters (Houlsby) & CNN & 42 & 26.77 \\ 
        T5 Large & Adapters (Houlsby) & CNN & 1 & 25.97 \\ 
        T5 Large & Adapters (Houlsby) & CNN & 0 & 26.96 \\ 
        \midrule
        T5 Large & Adapters (Pfeiffer) & RTE & 42 & 86.64 \\ 
        T5 Large & Adapters (Pfeiffer) & RTE & 1 & 87.00 \\ 
        T5 Large & Adapters (Pfeiffer) & RTE & 0 & 87.36 \\ 
        T5 Large & Adapters (Pfeiffer) & COPA & 42 & 56.00 \\ 
        T5 Large & Adapters (Pfeiffer) & COPA & 1 & 50.00 \\ 
        T5 Large & Adapters (Pfeiffer) & COPA & 0 & 50.00 \\ 
        T5 Large & Adapters (Pfeiffer) & BoolQ & 42 & 85.63 \\ 
        T5 Large & Adapters (Pfeiffer) & BoolQ & 1 & 85.87 \\ 
        T5 Large & Adapters (Pfeiffer) & BoolQ & 0 & 85.54 \\ 
        T5 Large & Adapters (Pfeiffer) & CNN & 42 & 26.51 \\ 
        T5 Large & Adapters (Pfeiffer) & CNN & 1 & 27.29 \\ 
        T5 Large & Adapters (Pfeiffer) & CNN & 0 & 27.29 \\ 
        \midrule
        T5 Large & Scaled Parallel & BoolQ & 42 & 84.74 \\ 
        T5 Large & Scaled Parallel & BoolQ & 1 & 85.23 \\ 
        T5 Large & Scaled Parallel & BoolQ & 0 & 84.98 \\ 
        T5 Large & Scaled Parallel & COPA & 42 & 66.00 \\ 
        T5 Large & Scaled Parallel & COPA & 1 & 72.00 \\ 
        T5 Large & Scaled Parallel & COPA & 0 & 65.00 \\ 
        T5 Large & Scaled Parallel & RTE & 42 & 87.36 \\ 
        T5 Large & Scaled Parallel & RTE & 1 & 88.45 \\ 
        T5 Large & Scaled Parallel & RTE & 0 & 87.36 \\ 
        T5 Large & Scaled Parallel & CNN & 42 & 26.84 \\ 
        T5 Large & Scaled Parallel & CNN & 1 & 26.59 \\ 
        T5 Large & Scaled Parallel & CNN & 0 & 26.75 \\ 
        \midrule
        T5 Large & Ln tuning & BoolQ & 42 & 83.46 \\ 
        T5 Large & Ln tuning & BoolQ & 1 & 83.55 \\ 
        T5 Large & Ln tuning & BoolQ & 0 & 83.24 \\ 
        T5 Large & Ln tuning & COPA & 42 & 60.00 \\ 
        T5 Large & Ln tuning & COPA & 1 & 69.00 \\ 
        T5 Large & Ln tuning & COPA & 0 & 63.00 \\ 
        T5 Large & Ln tuning & RTE & 42 & 85.92 \\ 
        T5 Large & Ln tuning & RTE & 1 & 85.92 \\ 
        T5 Large & Ln tuning & RTE & 0 & 85.56 \\ 
        T5 Large & Ln tuning & CNN & 42 & ~ \\ 
        T5 Large & Ln tuning & CNN & 1 & 25.50 \\ 
        T5 Large & Ln tuning & CNN & 0 & 25.50 \\ 
        \midrule
        T5 Large & LoRa (q and v) & RTE & 42 & 88.45 \\ 
        T5 Large & LoRa (q and v) & RTE & 1 & 87.00 \\ 
        T5 Large & LoRa (q and v) & RTE & 0 & 87.00 \\ 
        T5 Large & LoRa (q and v) & COPA & 42 & 71.00 \\ 
        T5 Large & LoRa (q and v) & COPA & 1 & 73.00 \\ 
        T5 Large & LoRa (q and v) & COPA & 0 & 70.00 \\ 
        T5 Large & LoRa (q and v) & BoolQ & 9803 & 81.75 \\ 
        T5 Large & LoRa (q and v) & BoolQ & 9486 & 84.65 \\ 
        T5 Large & LoRa (q and v) & BoolQ & 9247 & 83.42 \\ 
        T5 Large & LoRa (q and v) & CNN & 42 & 27.61 \\ 
        T5 Large & LoRa (q and v) & CNN & 1 & ~ \\ 
        T5 Large & LoRa (q and v) & CNN & 0 & ~ \\ 
        \midrule
        T5 Large & LoRa (all) & RTE & ~ & 87.73 \\ 
        T5 Large & LoRa (all) & RTE & ~ & 87.73 \\ 
        T5 Large & LoRa (all) & RTE & ~ & 87.73 \\ 
        T5 Large & LoRa (all) & COPA & ~ & 76.00 \\ 
        T5 Large & LoRa (all) & COPA & ~ & 73.00 \\ 
        T5 Large & LoRa (all) & COPA & ~ & 76.00 \\ 
        T5 Large & LoRa (all) & BoolQ & 42 & 85.35 \\ 
        T5 Large & LoRa (all) & BoolQ & 1 & 85.02 \\ 
        T5 Large & LoRa (all) & BoolQ & 0 & 85.66 \\ 
        T5 Large & LoRa (all) & CNN & 42 & 27.68 \\ 
        T5 Large & LoRa (all) & CNN & 1 & 26.25 \\ 
        T5 Large & LoRa (all) & CNN & 0 & ~ \\ 
        \midrule
        T5 Large & Krona & RTE & 42 & 86.28 \\ 
        T5 Large & Krona & RTE & 1 & 86.64 \\ 
        T5 Large & Krona & RTE & 0 & 86.28 \\ 
        T5 Large & Krona & COPA & 42 & 65.00 \\ 
        T5 Large & Krona & COPA & 1 & 70.00 \\ 
        T5 Large & Krona & COPA & 0 & 64.00 \\ 
        T5 Large & Krona & BoolQ & 42 & 83.06 \\ 
        T5 Large & Krona & BoolQ & 1 & 83.55 \\ 
        T5 Large & Krona & BoolQ & 0 & 83.64 \\ 
        T5 Large & Krona & CNN & 42 & 26.56 \\ 
        T5 Large & Krona & CNN & 1 & ~ \\ 
        T5 Large & Krona & CNN & 0 & ~ \\ 
        \midrule
        T5 Large & Compacter & BoolQ & 42 & 83.73 \\ 
        T5 Large & Compacter & BoolQ & 1 & 83.76 \\ 
        T5 Large & Compacter & BoolQ & 0 & 83.61 \\ 
        T5 Large & Compacter & COPA & 42 & 61.00 \\ 
        T5 Large & Compacter & COPA & 1 & 64.00 \\ 
        T5 Large & Compacter & COPA & 0 & 61.00 \\ 
        T5 Large & Compacter & RTE & 42 & 85.20 \\ 
        T5 Large & Compacter & RTE & 1 & 85.20 \\ 
        T5 Large & Compacter & RTE & 0 & 85.56 \\ 
        T5 Large & Compacter & CNN & 42 & ~ \\ 
        T5 Large & Compacter & CNN & 1 & 26.57 \\ 
        T5 Large & Compacter & CNN & 0 & 27.25 \\ 
        \midrule
        T5 Large & Compacter++ & BoolQ & 42 & 84.07 \\ 
        T5 Large & Compacter++ & BoolQ & 1 & 83.76 \\ 
        T5 Large & Compacter++ & BoolQ & 0 & 83.46 \\ 
        T5 Large & Compacter++ & COPA & 42 & 61.00 \\ 
        T5 Large & Compacter++ & COPA & 1 & 65.00 \\ 
        T5 Large & Compacter++ & COPA & 0 & 64.00 \\ 
        T5 Large & Compacter++ & RTE & 42 & 85.56 \\ 
        T5 Large & Compacter++ & RTE & 1 & 84.84 \\ 
        T5 Large & Compacter++ & RTE & 0 & 84.48 \\ 
        T5 Large & Compacter++ & CNN & 42 & 27.08 \\ 
        T5 Large & Compacter++ & CNN & 1 & ~ \\ 
        T5 Large & Compacter++ & CNN & 0 & ~ \\ 
        \midrule
        T5 Large & IA3 & RTE & 42 & 51.26 \\ 
        T5 Large & IA3 & RTE & 1 & 51.26 \\ 
        T5 Large & IA3 & RTE & 0 & 51.26 \\ 
        T5 Large & IA3 & COPA & 42 & 63.00 \\ 
        T5 Large & IA3 & COPA & 1 & 63.00 \\ 
        T5 Large & IA3 & COPA & 0 & 60.00 \\ 
        T5 Large & IA3 & BoolQ & 42 & 82.84 \\ 
        T5 Large & IA3 & BoolQ & 1 & 82.84 \\ 
        T5 Large & IA3 & BoolQ & 0 & 82.84 \\ 
        T5 Large & IA3 & CNN & 42 & 24.13 \\ 
        T5 Large & IA3 & CNN & 1 & ~ \\ 
        T5 Large & IA3 & CNN & 0 & ~ \\ 
        \midrule
        T5 Large & MAM & BoolQ & 42 & 64.65 \\ 
        T5 Large & MAM & BoolQ & 1 & 63.88 \\ 
        T5 Large & MAM & BoolQ & 0 & 64.46 \\ 
        T5 Large & MAM & COPA & 42 & 54.00 \\ 
        T5 Large & MAM & COPA & 1 & 55.00 \\ 
        T5 Large & MAM & COPA & 0 & 45.00 \\ 
        T5 Large & MAM & RTE & 42 & 46.93 \\ 
        T5 Large & MAM & RTE & 1 & 52.71 \\ 
        T5 Large & MAM & RTE & 0 & 52.71 \\ 
        T5 Large & MAM & CNN & 42 & 21.15 \\ 
        T5 Large & MAM & CNN & 1 & ~ \\ 
        T5 Large & MAM & CNN & 0 & ~ \\ 
        \midrule
        T5 Large & Prefix\_tuning & BoolQ & 42 & 62.17 \\ 
        T5 Large & Prefix\_tuning & BoolQ & 1 & 62.17 \\ 
        T5 Large & Prefix\_tuning & BoolQ & 0 & 62.17 \\ 
        T5 Large & Prefix\_tuning & COPA & 42 & 50.00 \\ 
        T5 Large & Prefix\_tuning & COPA & 1 & 51.00 \\ 
        T5 Large & Prefix\_tuning & COPA & 0 & 55.00 \\ 
        T5 Large & Prefix\_tuning & RTE & 42 & 50.90 \\ 
        T5 Large & Prefix\_tuning & RTE & 1 & 46.57 \\ 
        T5 Large & Prefix\_tuning & RTE & 0 & 50.54 \\ 
        T5 Large & Prefix\_tuning & CNN & 42 & 14.92 \\ 
        T5 Large & Prefix\_tuning & CNN & 1 & 17.37 \\ 
        T5 Large & Prefix\_tuning & CNN & 0 & 17.73 \\ 
        \midrule
        T5 Large & Prefix\_tuning\_flat & BoolQ & 42 & 0.24 \\ 
        T5 Large & Prefix\_tuning\_flat & BoolQ & 1 & 61.35 \\ 
        T5 Large & Prefix\_tuning\_flat & BoolQ & 0 & 22.81 \\ 
        T5 Large & Prefix\_tuning\_flat & COPA & 42 & 0.00 \\ 
        T5 Large & Prefix\_tuning\_flat & COPA & 1 & 0.00 \\ 
        T5 Large & Prefix\_tuning\_flat & COPA & 0 & 0.00 \\ 
        T5 Large & Prefix\_tuning\_flat & RTE & 42 & 0.00 \\ 
        T5 Large & Prefix\_tuning\_flat & RTE & 1 & 0.00 \\ 
        T5 Large & Prefix\_tuning\_flat & RTE & 0 & 0.00 \\ 
        T5 Large & Prefix\_tuning\_flat & CNN & 42 & 7.23 \\ 
        T5 Large & Prefix\_tuning\_flat & CNN & 1 & 7.26 \\ 
        T5 Large & Prefix\_tuning\_flat & CNN & 0 & 8.73 \\ 
        \midrule
        T5 Large & Unipelt & RTE & 42 & 52.71 \\ 
        T5 Large & Unipelt & RTE & 1 & 52.71 \\ 
        T5 Large & Unipelt & RTE & 0 & 52.71 \\ 
        T5 Large & Unipelt & COPA & 42 & 45.00 \\ 
        T5 Large & Unipelt & COPA & 1 & 55.00 \\ 
        T5 Large & Unipelt & COPA & 0 & 57.00 \\ 
        T5 Large & Unipelt & BoolQ & 42 & 62.17 \\ 
        T5 Large & Unipelt & BoolQ & 1 & 62.17 \\ 
        T5 Large & Unipelt & BoolQ & 0 & 37.83 \\ 
        T5 Large & Unipelt & CNN & 42 & 16.54 \\ 
        T5 Large & Unipelt & CNN & 1 & 18.23 \\ 
        T5 Large & Unipelt & CNN & 0 & 17.18 \\ \hline
\end{longtable}
}

\end{document}